\documentclass[letterpaper, 10 pt, conference]{ieeeconf}  

\IEEEoverridecommandlockouts                              

\usepackage[disable]{todonotes}

\usepackage{graphicx}
\usepackage{cite}
\usepackage{xcolor}
\usepackage{hyperref}
\usepackage[per-mode=symbol]{siunitx}
\usepackage{proof}
\usepackage{bussproofs}
\usepackage{mathpartir}
\usepackage{rotating}
\usepackage{booktabs}
\usepackage{subcaption}

\usepackage{algorithm2e}
\SetKwComment{Comment}{/* }{ */}
\RestyleAlgo{ruled}

\usepackage[all]{xy} \SelectTips{cm}{}  
\newdir{ >}{{}*!/-8pt/@{>}}
\newdir{|>}{%
	!/1.6pt/@{|}*:(1,-.2)@^{>}*:(1,+.2)@_{>}}

\usepackage{todonotes}
\newcommand{\todoil}[1]{\todo[inline,caption={}]{#1}}

\usepackage{mathtools}
\usepackage{hyperref}
\usepackage[capitalize, nameinlink]{cleveref}

\usepackage{aliascnt}
\usepackage[appendix=strip]{apxproof}

\theoremstyle{plain}
\newtheoremrep{theorem}{Theorem}[section]
\Crefname{theorem}{Thm.}{Thm.}
\newaliascnt{propositioncnt}{theorem}
\newtheoremrep{proposition}[propositioncnt]{Proposition}
\Crefname{propositioncnt}{Prop.}{Prop.}
\newaliascnt{lemmacnt}{theorem}
\newtheoremrep{lemma}[lemmacnt]{Lemma}
\Crefname{lemmacnt}{Lem.}{Lem.}
\newaliascnt{corollarycnt}{theorem}
\newtheoremrep{corollary}[corollarycnt]{Corollary}
\Crefname{corollarycnt}{Cor.}{Cor.}
\newaliascnt{factcnt}{theorem}

\Crefname{factcnt}{Fact}{Fact}
\newaliascnt{assumptioncnt}{theorem}
\newtheorem{assumption}[assumptioncnt]{Assumption}
\Crefname{assumptioncnt}{Asm.}{Asm.}
\theoremstyle{definition}
\newaliascnt{remarkcnt}{theorem}
\newtheorem{remark}[remarkcnt]{Remark}
\Crefname{remarkcnt}{Rem.}{Rem.}
\newaliascnt{notationcnt}{theorem}

\Crefname{notationcnt}{Notation}{Notation}

\newaliascnt{requirementcnt}{theorem}

\Crefname{requirementcnt}{Requirement}{Requirement}
\newaliascnt{requirementscnt}{theorem}

\Crefname{requirementscnt}{Requirements}{Requirements}

\Crefname{researchquestioncnt}{RQ}{RQ}

\theoremstyle{definition}
\newaliascnt{definitioncnt}{theorem}
\newtheoremrep{definition}[definitioncnt]{Definition}
\Crefname{definitioncnt}{Def.}{Def.}
\newaliascnt{examplecnt}{theorem}
\newtheoremrep{example}[examplecnt]{Example}
\Crefname{examplecnt}{Example}{Examples}

\crefformat{section}{\S{}#2#1#3}
\Crefname{table}{Table}{Table}
\Crefname{figure}{Fig.}{Fig.}
\Crefname{equation}{}{}
\Crefname{line}{Line}{Line}

\usepackage{stmaryrd,amssymb,amsmath}

\usepackage{mathrsfs}

\usepackage{wrapfig}
\usepackage{comment}
\specialcomment{auxproof}
{\mbox{}\newline\textbf{BEGIN: AUX-PROOF}\dotfill\newline}
{\mbox{}\newline\textbf{END: AUX-PROOF}\dotfill\newline}
\excludecomment{auxproof}

\newcommand{\dHL}{\ensuremath{\mathrm{dFHL}}}
\newcommand{\dL}{\ensuremath{\mathrm{dL}}}    

\newcommand{\bmax}{b_{\mathrm{max}}}
\newcommand{\bmin}{b_{\mathrm{min}}}
\newcommand{\amax}{a_{\mathrm{max}}}

\newcommand{\DM}{DM}
\newcommand{\AC}{AC}

\newcommand{\POV}[1]{\ensuremath{\mathsf{POV}{#1}}}
\newcommand{\SV}{\ensuremath{\mathsf{SV}}}
\newcommand{\CZ}{\ensuremath{\mathsf{CZ}}}

\newcommand{\xSV}{x_{\SV}}
\newcommand{\vSV}{v_{\SV}}
\newcommand{\tSV}{t_{\SV}}
\newcommand{\aPOV}{a_{\POV{}}}
\newcommand{\xPOV}{x_{\POV{}}}
\newcommand{\vPOV}{v_{\POV{}}}
\newcommand{\xPOVmax}{x_{\POV{}}^{\mathsf{max}}}
\newcommand{\vPOVmax}{v_{\POV{}}^{\mathsf{max}}}
\newcommand{\xPOVmin}{x_{\POV{}}^{\mathsf{min}}}
\newcommand{\vPOVmin}{v_{\POV{}}^{\mathsf{min}}}
\newcommand{\tPOV}{t_{\POV{}}}

\newcommand{\CZStartSV}{x_{\SV}^{\mathsf{CZStart}}}
\newcommand{\CZEndSV}{x_{\SV}^{\mathsf{CZEnd}}}
\newcommand{\CZStartPOV}{x_{\POV{}}^{\mathsf{CZStart}}}
\newcommand{\CZEndPOV}{x_{\POV{}}^{\mathsf{CZEnd}}}

\newcommand{\true}{\ensuremath{\mathsf{true}}}
\newcommand{\false}{\ensuremath{\mathsf{false}}}

\newcommand{\BC}{BC}

\newcommand{\hquad}[4]{\left\{#2\right\}~#3~\left\{#4\right\}\colon{}#1}

\newcommand{\dRSS}{\ensuremath{\mathsf{dRSS}}}

\newcommand{\Variables}{\ensuremath{V}}

\newcommand{\var}{\ensuremath{x}}
\newcommand{\term}{\ensuremath{e}}
\newcommand{\vars}{\ensuremath{\mathbf{x}}}
\newcommand{\funs}{\ensuremath{\mathbf{f}}}

\newcommand{\asserta}{\ensuremath{A}}
\newcommand{\assertb}{\ensuremath{B}}

\newcommand{\safetya}{\ensuremath{S}}

\newcommand{\coma}{\alpha}
\newcommand{\comb}{\beta}

\newcommand{\skipClause}{\mathsf{skip}}

\newcommand{\assignClause}[2]{#1 \mathop{{:}{=}} #2}

\newcommand{\dwhileKeyword}{\mathsf{dwhile}}
\newcommand{\dwhileHeader}[1]{\dwhileKeyword\,(#1)}
\newcommand{\dwhileClause}[2]{\dwhileHeader{#1}\,\{#2\}}

\newcommand{\odeClause}[2]{\dot{#1} = #2}

\newcommand{\whileKeyword}{\mathsf{while}}
\newcommand{\whileHeader}[1]{\whileKeyword\,(#1)}
\newcommand{\whileClause}[2]{\whileHeader{#1}\,\{#2\}}

\newcommand{\ifHeader}[1]{\mathsf{if}\,(#1)}
\newcommand{\elseKeyword}{\mathsf{else}}
\newcommand{\ifThenElse}[3]{\ifHeader{#1}\,\{#2\}\,\elseKeyword{}\,\{#3\}}

\newcommand{\limply}{\Rightarrow}
\newcommand{\bigland}{\bigwedge}
\newcommand{\biglor}{\bigvee}

\newcommand{\store}{\ensuremath{\rho}}
\newcommand{\update}[3]{\ensuremath{{#1}[{#2} \to {#3}]}}

\newcommand{\sem}[2]{\ensuremath{\left\llbracket {#1} \right\rrbracket_{#2}}}
\newcommand{\statea}{\ensuremath{s}}
\newcommand{\state}[2]{\ensuremath{\langle {#1}, {#2} \rangle}}
\newcommand{\red}[1]{\to}

\newcommand{\convergeSymbol}{\Downarrow}
\newcommand{\converge}[2]{\ensuremath{{#1} \mathrel{\convergeSymbol} {#2}}}

\newcommand{\set}[1]{\left\{ {#1} \right\}}

\newcommand{\R}{\ensuremath{\mathbb{R}}}

\newcommand{\KeYmaeraX}{\textsc{KeYmaera~X}}  

\usepackage{booktabs}
\usepackage{balance}

\usepackage{tikz}
\usepackage{tikz-qtree}
\usetikzlibrary{positioning}
\usetikzlibrary{automata}
\usetikzlibrary{calc}
\usetikzlibrary{arrows.meta}

\tikzset{FAstyle/.style={
    shorten >=1pt,
    node distance=5cm,
    on grid,
    auto,
    every state/.style={
      draw=blue!50,
      very thick,
      top color=white,
      bottom color=blue!20,
      minimum size=0pt,
      shape=rectangle
    },
    >=Stealth[round],
    thick,
    draw=black!50
  }
}

\title{\LARGE \bf
Formal Verification of Intersection Safety for Automated Driving*
}

\author{James Haydon$^{1}$, Martin Bondu$^{1}$, Clovis Eberhart$^{1,3}$, J\'er\'emy Dubut$^{2,1}$, and
Ichiro Hasuo$^{1,4}$%
\thanks{*The authors are supported by JST ERATO HASUO Metamathematics for Systems Design Project (No. JPMJER1603) and JST START Grant (No. JPMJST2213).
 IH is supported by JST CREST Grant (No. JPMJCR2012).}%
\thanks{$^{1}$National Institute of Informatics, Tokyo, Japan}
\thanks{$^{2}$AIST, Tokyo, Japan
}
\thanks{$^{3}$Japanese-French Laboratory for Informatics, IRL3527, Tokyo, Japan}
\thanks{$^{4}$SOKENDAI, Hayama, Japan
}}

\newcommand{\safe}{\mathsf{Safe}}
\newcommand{\ter}{\mathsf{Terminal}}
\newcommand{\init}{\mathsf{Init}}
\newcommand{\tend}{t_\mathsf{end}}
\newcommand{\bbR}{\mathbb{R}}
\newcommand{\Rplus}{\mathbb{R}_{+}}
\newcommand{\dyn}{\mathsf{dyn}}
\newcommand{\vari}{\mathsf{var}}
\newcommand{\jump}{\mathsf{jump}}
\newcommand{\Open}{\mathsf{Open}}
\newcommand{\Loc}{L}
\newcommand{\Edge}{E}
\newcommand{\Init}{\mathsf{Init}}
\newcommand{\Final}{\mathsf{Final}}
\newcommand{\Unsafe}{\mathsf{Unsafe}}
\newcommand{\Flow}{\mathsf{Flow}}
\newcommand{\Guard}{\mathsf{Guard}}
\newcommand{\Assign}{\mathsf{Assign}}
\newcommand{\Int}{\mathsf{int}}
\newcommand{\toProgr}[1]{[#1]}

\begin{document}

\maketitle
\thispagestyle{empty}
\pagestyle{empty}

\begin{abstract}
We build on our recent work on formalization of \emph{responsibility-sensitive safety} (RSS) and present the first formal framework that enables mathematical proofs of the safety of control strategies in intersection scenarios. Intersection scenarios are challenging due to the complex interaction between vehicles; to cope with it, we extend the program logic $\dHL$ in the previous work and introduce a novel formalism of \emph{hybrid control flow graphs} on which our algorithm can automatically discover an \emph{RSS condition} that ensures safety. An RSS condition thus discovered is experimentally evaluated; we observe that it is safe (as  our safety proof says) and  is not overly conservative.
\end{abstract}

\section{INTRODUCTION}
Safety of automated driving vehicles (ADVs) is a problem of growing industrial and social interest. New technologies
\begin{auxproof}
 , 
 especially in sensing and perception (such as lidars and deep neural networks), 
\end{auxproof}
are making ADV technologically feasible; but for the social acceptance, their safety should be guaranteed and explained.

\begin{auxproof}
 Many existing approaches to the safety assurance of ADVs are \emph{statistical}, such as accident statistics and testing (typically by computer simulation). Another family of approaches---namely \emph{logical} ones---are attracting growing attention, too.
 In logical approaches, safety is stated as a mathematical theorem and it is given a logical proof. Such approaches via proofs are called \emph{formal verification} and have been actively pursued for  software and computing hardware, with a number of notable successes. 

Principal advantages of formal verification are \emph{strong guarantees} (logical proofs never go wrong) and \emph{explainability} (a proof is a comprehensive record of step-by-step arguments towards a safety claim). Therefore, formal verification are widely seen as an important component in  society's effort towards ADV safety, complementing statistical approaches. 
\end{auxproof}

In this paper, we pursue \emph{formal verification} of ADV safety, that is, to provide its mathematical proofs. This \emph{logical} approach,  compared to \emph{statistical} approaches such as accident statistics and scenario-based testing, has major advantages of \emph{strong guarantees} (logical proofs never go wrong) and \emph{explainability} (a proof records step-by-step safety arguments).

Specifically, we introduce a theoretical framework for formal verification of ADV safety in \emph{intersection scenarios}.  We derive an algorithm for computing  assumptions (called \emph{RSS conditions}) needed for  safety proofs. We present its   implementation, and we experimentally evaluate its output.  We build our contributions
on top of  the conceptual methodology called RSS~\cite{ShalevShwartzSS17RSS} and its mathematical formalization~\cite{HasuoEHDBKPZPYSIKSS23}. 

Below we  describe our motivation and  contributions. We  introduce, at the 
same time, the technical context of RSS that we rely on. The high-level overview is summarized in~\cref{subsec:introContrib}. 

\subsection{Responsibility-Sensitive Safety (RSS) }\label{subsec:introRSS}
RSS is a rich concept with many aspects. The introduction below is focused on its implications on formal verification. See~\cite{Hasuo22RSSarXiv} for a more extensive introduction.

Formal verification of ADV safety is a natural idea with many important advantages. A big problem, however, is  its feasibility. In order to write mathematically rigorous proofs in formal verification,
 one needs rigorous \emph{definitions} of all the concepts involved. Such definitions amount to mathematical \emph{modeling} of target systems, which is hard for ADVs due to 1) the complexity of vehicles themselves and 2)  complex interaction between traffic participants.

\emph{Responsibility-sensitive safety (RSS)}~\cite{ShalevShwartzSS17RSS} is a methodology for this challenge of modeling. Its central idea is to logically split the safety theorem into the following two, and to focus the proof effort to the conditional safety lemma only.



\vspace{.3em}
\noindent
\begin{minipage}{.49\textwidth}
 \begin{lemma}[conditional safety] \label{lem:condSafe}
 If all vehicles comply with so-called RSS rules, then there is no collision.
 
 \end{lemma}
\end{minipage}

\vspace{.3em}
\noindent
\begin{minipage}{.49\textwidth}
\begin{assumption}[rule compliance]\label{asmp:ruleCompliance}
 All vehicles do comply with RSS rules.
\end{assumption}
\end{minipage}

\vspace{.3em}
This way, most of the modeling difficulties can be left out of the scope of mathematical arguments---for example, the internal working of vehicles is confined to the rule compliance assumption. This  makes rigorous proofs feasible.

One may wonder if \cref{asmp:ruleCompliance} may be too strong---it may be false and thus threaten the value of a safety proof. It turns out that RSS rules are \emph{effective} rules, not only technically but also socially and industrially. Their granularity is suited for imposing them as contracts and standards---as shown below in \cref{ex:RSSSafetyDistance}---making \cref{asmp:ruleCompliance} a realistic assumption. It is reasonable that vehicles comply with RSS rules (the rules can be even \emph{enforced} by safety architectures, see e.g.~\cite{HasuoEHDBKPZPYSIKSS23,EberhartDHH23}). \cref{asmp:ruleCompliance} imposes rules also on other traffic participants such as pedestrians. In this case, those rules are chosen to be very mild commonsense rules (e.g.\ ``no jumping to highways''), and they are enforced by social and legal means. See e.g.~\cite{Hasuo22RSSarXiv}.


An RSS rule $(A,\alpha)$ is a pair of an \emph{RSS condition} $A$ and a \emph{proper response} $\alpha$. A proper response is a specific control strategy; it can be thought of as a \emph{minimum risk maneuver (MRM)}.  Then \cref{lem:condSafe} states, to be precise, that the execution of $\alpha$, starting in a state where $A$ is true, is collision-free. 

Here is an example of an RSS rule from~\cite{ShalevShwartzSS17RSS}.

\begin{wrapfigure}[4]{r}{0pt}
 \centering
 \includegraphics[bb=66 217 305 271,clip,width=9em]{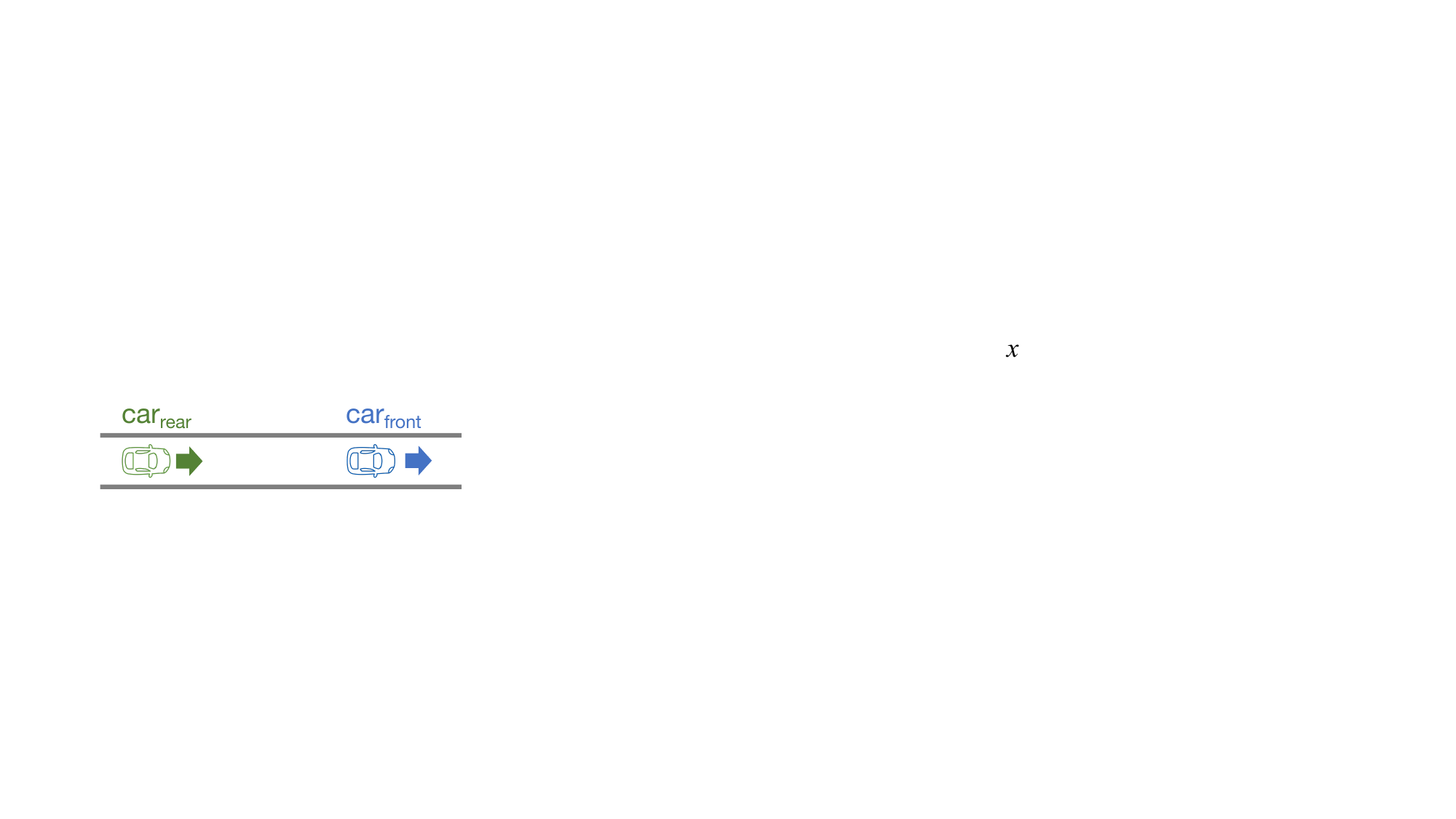}
 \caption{one-way traffic}
 \label{fig:onewayTraffic}
\end{wrapfigure}

\begin{example}[one-way traffic~\cite{ShalevShwartzSS17RSS}]\label{ex:RSSSafetyDistance}
Consider~\cref{fig:onewayTraffic}, where the subject vehicle
  (\SV{}, $\mathrm{car}_\mathrm{rear}$) drives behind another car
  ($\mathrm{car}_\mathrm{front}$). 
The RSS condition $A$ for this scenario is
  \begin{equation}\label{eq:RSSOnewayTraffic}
    A \;=\;\bigl(y_{f} - y_{r} > \dRSS(v_{f}, v_{r})\bigr),
  \end{equation}
  where $\dRSS(v_{f}, v_{r})$ is the \emph{RSS safety distance} defined by
  \begin{equation}\label{eq:RSSMinDist}\small
    \begin{aligned}
    &  \max\Bigl(\,0,\,
          v_{r}\rho + \frac{1}{2}\amax  \rho^2 + \frac{(v_{r} + \amax  \rho)^2}{2\bmin} -\frac{v_{f}^2}{2\bmax}\,\Bigr).
    \end{aligned}
  \end{equation}
  Here $y_{f}, y_{r}$ are positions of the cars,
  $v_{f}, v_{r}$ are velocities,
$\rho$
  is the  \emph{response time} for $\mathrm{car}_\mathrm{rear}$,
$a_{\max}$ is the maximum
acceleration rate,
  $\bmin$ is the maximum comfortable braking rate, and $\bmax$ is the maximum emergency
  braking rate.

 The proper response $\alpha$  dictates  \SV{}
        ($\mathrm{car}_\mathrm{rear}$) to engage the maximum comfortable braking (at rate $\bmin$)
        when condition~\cref{eq:RSSOnewayTraffic} is violated.

 For the RSS rule $(A,\alpha)$, proving the conditional safety lemma is not hard. See~\cite{ShalevShwartzSS17RSS,Hasuo22RSSarXiv}  (informal) and~\cite{HasuoEHDBKPZPYSIKSS23} (formal).
\end{example}

We note that the above RSS rule is an a priori, rigorous yet simple rule, the compliance with which is externally checkable.  It is  independent of specific car models or manufacturers. It applies to all vehicles by changing the values of parameters such as $\amax$.

 RSS rules can be used for various purposes, such as attribution of liability, safety metrics, regulations and standards, and runtime safeguard mechanism. See e.g.~\cite{ShalevShwartzSS17RSS,OborilS21,HasuoEHDBKPZPYSIKSS23}.
\begin{auxproof}
  One is attribution of liability~\cite{ShashuaSS18NHTSA}: by the conditional safety lemma, when there is a collision, then there is a vehicle that did not comply with RSS rules, which is held liable. Another is as safety regulations and standards:  the claim that a vehicle complies with RSS rules will convince society to accept it on public roads. RSS rules can be implemented on ADVs in a \emph{safety architecture}, as a additional safeguard  to an existing controller (see e.g.~\cite{OborilS20IV,HasuoEHDBKPZPYSIKSS23}).
\end{auxproof}

\subsection{Formalization of RSS by the Program Logic $\dHL$}
 An RSS rule must be derived, and proved  safe (\cref{lem:condSafe}), for each individual driving scenario. Towards broad application of RSS, we have to do such proofs for many scenarios. 
Doing so informally (in a pen-and-paper manner) is not desirable for scalability, maintainability, and accountability. 

This is why we pursued the formalization of RSS in~\cite{HasuoEHDBKPZPYSIKSS23}. We introduced a logic $\dHL$---a symbolic framework to write proofs in---extending classic \emph{Floyd--Hoare logic}~\cite{Hoare69} with differential equations. The logic $\dHL$ derives \emph{Hoare quadruples} $\{A\}\alpha\{B\}:S$; it means that the execution of a \emph{hybrid program} $\alpha$, started when a \emph{precondition} $A$ is true, terminates and makes a \emph{postcondition} $B$ true. The \emph{safety condition} $S$  specifies a property that should be true throughout the execution of $\alpha$. 

Note that \cref{lem:condSafe} of RSS is naturally mapped to a Hoare quadruple: if we let $A$ be an RSS condition and $\alpha$ be a proper response, then $S$ (expressing collision-freedom) is ensured throughout. Moreover,  the postcondition $B$ can express the \emph{goal} of $\alpha$, such as to stop at a desired position. This extension of RSS, where RSS rules can guarantee not only safety but also goal achievement, is called \emph{GA-RSS}  in~\cite{HasuoEHDBKPZPYSIKSS23}. 

\begin{figure}[tbp]
\begin{minipage}{.65\columnwidth}
\centering
   \includegraphics[scale=.18]{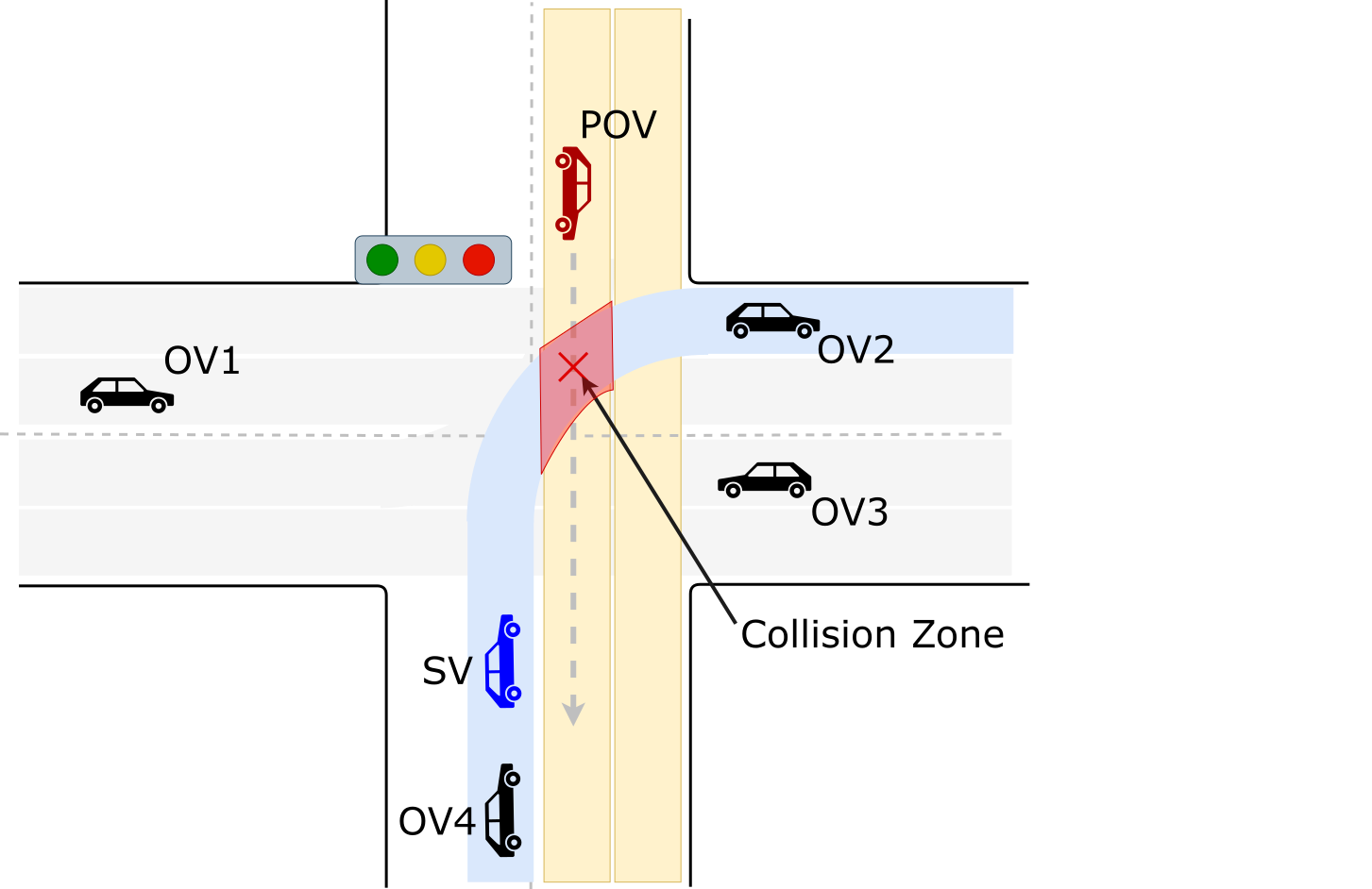}
 \caption{the intersection driving scenario, with the \emph{subject vehicle} \SV{} making a right turn and the \emph{principal other vehicle} \POV{} coming from the opposite}
 \label{fig:intersection}
\end{minipage}
\hfill
\begin{minipage}{.32\columnwidth}
\centering
\includegraphics[bb=0 0 337 458,clip,scale=.25]{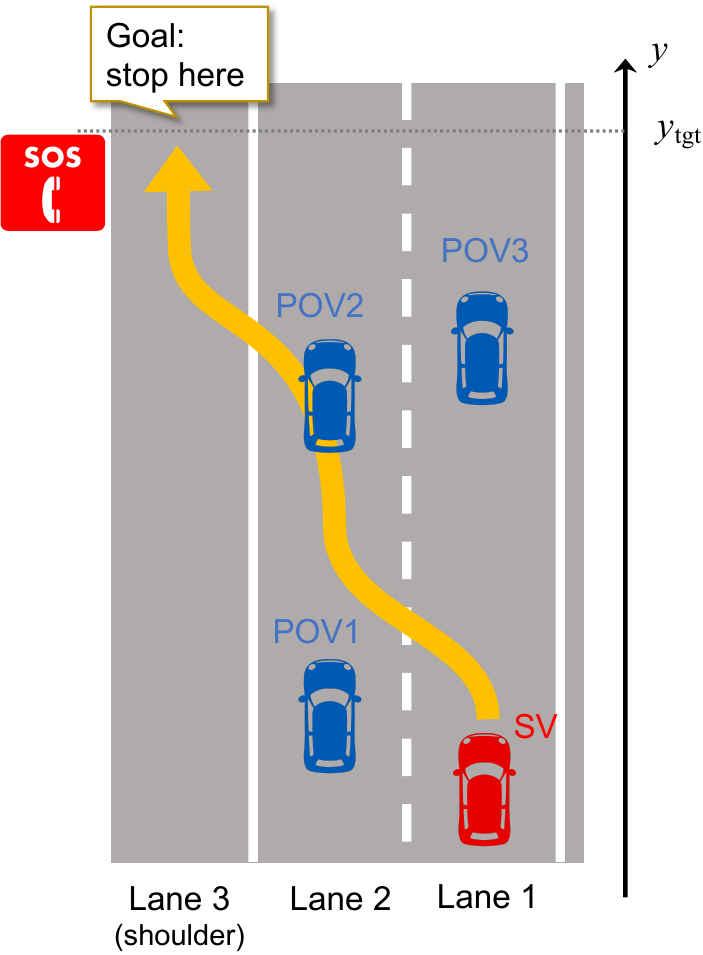}
 \caption{pull over}
 \label{fig:pullover}
\end{minipage}
 \vspace{-2em}
\end{figure}

Another  benefit of formalization by $\dHL$ is \emph{compositional} reasoning. In Floyd--Hoare logic, one can derive
 $\{A\}\,\alpha;\beta\,\{C\}$ (a property of composition $\alpha;\beta$) from 
 $\{A\}\,\alpha\,\{B\}$ and
 $\{B\}\,\beta\,\{C\}$ (properties of  components).
Via similar compositionality in $\dHL$, we devised in~\cite{HasuoEHDBKPZPYSIKSS23} a workflow in which a complex scenario is split into  subscenarios, and RSS rules are derived in a divide-and-conquer manner.

As a case study, we derived an RSS rule for the \emph{pull over} scenario (\cref{fig:pullover}). It is a complex scenario with a goal that requires high-level maneuver planning. It is an important MRM, too, for an ADV exiting its ODD.

\subsection{Automated Derivation of RSS Rules for Intersections}
Our goal in this paper is a formal framework for deriving RSS rules for \emph{intersection scenarios} such as \cref{fig:intersection}. This class of driving scenarios exhibit unique challenges; to cope with them, we introduce a new theory that extends $\dHL$ and emphasizes automation, together with its implementation.

 Intersection scenarios exhibit complex interaction between vehicles.
This makes modeling hard: there are many ``discrete modes'' due to the interaction,  making manual modeling effort error-prone. Desired here is a \emph{compositional modeling formalism} where one can model each vehicle separately, a feature absent in~\cite{HasuoEHDBKPZPYSIKSS23}. Another challenge is in verification: many discrete modes lead to complex case distinction and proof complexity; therefore \emph{proof automation} is highly desired.

\begin{auxproof}
 Specifically, one unique challenge is in modeling.  Complex interaction results in many ``discrete modes,'' and enumerating them by a modeler's manual effort is expensive and error-prone. Desired here is a formalism in which one can model each vehicle individually and let them automatically generate the whole system model. The modeling formalism in $\dHL$---called hybrid programs, see \cref{subsec:prelimHybridProg}---does not have this feature. 

Another challenge is in verification. A number of ``discrete modes'' calls for massive case distinction in safety proofs, and manual efforts for such proofs can be highly expensive. An automated workflow is desired here. 

\end{auxproof}

Our answer to these challenges is \emph{hybrid control flow graphs (hCFGs)} as a modeling formalism (\cref{sec:hCFG}). They are to hybrid programs in~\cite{HasuoEHDBKPZPYSIKSS23} what control flow graphs are to usual imperative programs; we introduce a  synchronization mechanism so that multiple hCFGs compose. For reasoning, we adapt $\dHL$ and introduce the notion of \emph{Hoare annotation}  (\cref{sec:hoareAnnotations}). We propose an automated reasoning algorithm, and present its implementation and experimental evaluation.

\subsection{Contributions}\label{subsec:introContrib}
We present the first framework for deriving RSS rules and proving their safety for intersection scenarios.  Our technical contributions are as follows.
1)
 We introduce the modeling formalism of hCFGs that enables compositional agent-wise modeling (\cref{sec:hCFG}). Their semantics is via  translation to hybrid programs~\cite{HasuoEHDBKPZPYSIKSS23}. 
2)
 We introduce the reasoning formalism of Hoare annotations (\cref{sec:hoareAnnotations}), an automated reasoning algorithm, and its implementation (\cref{sec:impl}). 
3)
 We apply the above framework to the intersection situation in \cref{sec:intersectionModeling}---it consists of the driving scenario~\cref{fig:intersection} and a specific  proper response $\alpha$---and let our implementation automatically derive an RSS condition $A$. 
4)
We experimentally evaluate the RSS rule $(A,\alpha)$.

\subsection{Organization, Notations and Terminologies}
After reviewing $\dHL$~\cite{HasuoEHDBKPZPYSIKSS23} in \cref{sec:prelimdFHL}, we introduce hCFGs in \cref{sec:hCFG}, with their composition and semantics. The intersection case study is given  in \cref{sec:intersectionModeling}. We go back to the general theory and introduce  Hoare annotations in \cref{sec:hoareAnnotations}.  Our algorithm is given in \cref{sec:impl}; in \cref{sec:expr} is the experimental evaluation.

The cardinality of a set $X$ is denoted by $|X|$. The set of polynomials over a set $X$ of variables is denoted by $\mathbb{R}[X]$. The syntactic equality is denoted by $\equiv$. We let $[i,j]=\{i, i+1, \dotsc, j\}$ for integers $i,j$ such that $i\le j$. The propositional constants $\true, \false$ denote truth and falsehood, respectively.

By a \emph{driving scenario} we refer to an environmental setting in which the subject vehicle \SV{} drives. A \emph{driving situation} is a combination of a driving scenario and \SV{}'s control strategy. The latter is typically some MRM (a proper response in RSS).

\subsection{Related and Future Work}\label{subsec:relatedWork}
Some RSS rules are implemented and offered as a library~\cite{GassmannOBLYEAA19IV}. However, its coverage of various driving scenarios seems limited. For example, \cite{KarimiD22} reports that CARLA's autopilot safeguarded by the RSS library exhibits dangerous behaviors in an intersection scenario. This is because the RSS rule used there is essentially the one in \cref{ex:RSSSafetyDistance} and does not address the intersection scenario. 

\begin{auxproof}
 Some RSS rules have been implemented and are offered as a library~\cite{GassmannOBLYEAA19IV}. Integration of the goal-aware RSS rules we derive in this paper, in the library, is future work. One advantage of doing so is that the GA-RSS rules will then accommodate varying road shapes.

 Need to cite~\cite{KarimiD22}
 \begin{itemize}
 \item about test-case generation
 \item novelty is an iterative generation of increasingly challenging test cases
 \item demonstrated that auto-pilot-plus-RSS can exhibit unsafe behaviors. This is because, the RSS rules implemented in a safety architecture is essentially for the one-way traffic scenario (\cref{ex:RSSSafetyDistance}) and does not fully take  the studied intersection scenario into account
 \end{itemize}
\end{auxproof}

There are some extensions of RSS. Besides our formalized, compositional and goal-achieving one~\cite{HasuoEHDBKPZPYSIKSS23}, they study 1) parameter selection for balancing safety and progress~\cite{KonigshofOSS22}, 2) an empirical (not logical) safeguard layer in addition to RSS~\cite{OborilS21},
\begin{auxproof}
  invoked in case parameter values are too liberal~\cite{OborilS21} (this layer is empirical and does not come with logical safety proofs), 
\end{auxproof}
3) extension to unstructured traffic with vulnerable road users~\cite{PaschOGS21}, and 4)    swerves as evasive manoeuvres~\cite{deIacoSC20IV}. The formalization in~\cite{HasuoEHDBKPZPYSIKSS23} has been extended to verification of safety architectures, too~\cite{EberhartDHH23}.
All these extensions are orthogonal to the formalized approach of the current work, leaving their integration as future work.

\begin{auxproof}
 Need to cite~\cite{KonigshofOSS22}
 \begin{itemize}
 \item studies parameter values for RSS rules, such as the maximum acceleration rate, in order to balance safety and progress
 \item focus on lateral safety (unlike the longitudinal safety e.g.\ in \cref{ex:RSSSafetyDistance})
 \item The presented technique is empirical
 \end{itemize}

 \cite{OborilS21}
 \begin{itemize}
 \item $\text{RSS}^{+}$: a proactive extension of RSS
 \item takes a proactive measure in case the situation goes beyond the assumptions for the RSS rules---typically because of too strong assumptions on parameter values 
 \item Its essence is an additional (statistical) layer on top of the RSS-based safety architecture. The proper response suggested by this additional layer, however, does not come with a mathematical safety guarantee, unlike the proper response of the original RSS
 \item (Our hierarchical RSS work is the same in that it adds another layer, but a big difference is that our additional layer is logical again, and the construction is compositional)
 \end{itemize}

 \cite{PaschOGS21}
 \begin{itemize}
 \item Expands the application domain of RSS from structured driving scenarios to unstructured ones with vulnerable road users (VRU)
 \item The main technical focus is the choice of kinematic models for different class of VRUs, and their parameter selection
 \item This can be integrated with the current work
 \end{itemize}

 Recent extensions of RSS include a risk-aware one~\cite{OborilS20IV}  and one that allows swerves as evasive manoeuvres~\cite{deIacoSC20IV}. These extensions shall be pursued in our current goal-aware framework. In particular, allowing swerves should be possible, and it will significantly improve the progress of a RSS-supervised controller.
\end{auxproof}


Some recent works formalize traffic rules in temporal logic, so that they can be effectively monitored~\cite{DBLP:conf/ivs/MaierhoferMA22,DBLP:conf/ivs/LinA22}.  Rules are given externally in~\cite{DBLP:conf/ivs/MaierhoferMA22,DBLP:conf/ivs/LinA22}, while our goal is to derive such rules and prove their safety. 

An automated driving controller typically aims also at requirements other than safety (such as comfort). For this purpose, use of \emph{hierarchically structured} human-interpretable
rules is advocated in~\cite{VeerLCCP2023}. In our previous work~\cite{EberhartDHH23} we also investigated hierarchically structured RSS rules, this time for the purpose of graceful degradation of safety. Combination of~\cite{EberhartDHH23} and the current work will not be hard.

\begin{auxproof}
 
 Need to cite~\cite{DBLP:conf/ivs/MaierhoferMA22}, ``Formalization of Intersection Traffic Rules in Temporal Logic''. 
 \begin{itemize}
 \item Formalizing existing traffic rules in temporal logic. It is not about proving the safety of some control strategy.
 \end{itemize}

 Need to cite~\cite{DBLP:conf/ivs/LinA22}, ``Rule-Compliant Trajectory Repairing using Satisfiability Modulo Theories.'' Sounds like our use of RSS rules in the safety architecture. Say how ours differs from theirs. As I understand, their ``rule'' is a traffic rule such as ``give way to pedestrians.''
 \begin{itemize}
 \item Rules are given
 \item They discuss how to repair trajectories
 \item In contrast, ours is about designing rules and proving their safety
 \end{itemize}
\end{auxproof}

Formal verification of ADV decision making in intersections is studied in~\cite{ChouhanB20,SaraogluPJ22,Rajhans13PhD}. They all take model-checking approaches, unlike the current work that emphasizes theorem proving. The work~\cite{ChouhanB20} employs statistical model checking, where correctness guarantees are by sampling and thus not absolute. The work~\cite{SaraogluPJ22} abstracts away continuous dynamics, unlike our explicit modeling by ODEs. Rigorous and fine-grained verification is pursued in~\cite{Rajhans13PhD} where high-level decision making and low-level continuous dynamics are holistically analyzed by \emph{multi-model heterogeneous verification}. It is not clear how this method generalizes to numerous variations of intersection scenarios (while their $\dHL$ modeling will not be hard). More comparison  is future work.

 Rule-based decision making  for intersection scenarios is proposed in~\cite{AksjonovK21}, but it does not come with safety proofs. From the RSS point of view, the work
is suggesting proper responses, for which safety-guaranteeing RSS conditions can be derived using our framework. Doing so is future work.

For intersection safety, statistical and learning-based approaches are actively pursued, e.g.~\cite{Gao0CWHXWW0K21}. They are different from our logical approach that allows rigorous safety proofs.

\begin{auxproof}
 \cite{AksjonovK21}
 \begin{itemize}
 \item Proposes a rule-based decision making system for intersection scenarios
 \item No logical formalization, no safety proofs
 \item In relation to the current work, it can be seen as a suggestion of a class of proper responses
 \end{itemize}
\end{auxproof}


Formal verification of ADV safety is pursued in e.g.~\cite{RizaldiISA18,RoohiKWSL18arxiv,KrookSLFF19}. The works~\cite{RizaldiISA18,RoohiKWSL18arxiv} adopt much more fine-grained modeling compared to the current work and~\cite{HasuoEHDBKPZPYSIKSS23}; a price for doing so seems to be scalability and flexibility. The work~\cite{KrookSLFF19} makes discrete (and thus coarse-grained) modeling.

\begin{auxproof}
 \begin{itemize}
 \item 
 many optimisation- and learning-based planning algorithms for safe driving, such as~\cite{mcnaughton2011motion} (they do not offer rigorous safety guarantee),
 \item 
 testing-based approaches for ADS safety, such as~\cite{LuoZAJZIWX21ASEtoAppear} (they do not offer rigorous safety guarantee, either), and
 \item 
 runtime verification approaches for ADS safety by reachability analysis, such as~\cite{LiuPA20IV,PekA18IROS}.
 \end{itemize}

 The problem of formally verifying correctness of RSS rules is formulated and investigated in~\cite{RoohiKWSL18arxiv}. Their formulation is based on a rigorous notion of signal; they argue that none of the existing \emph{automated} verification tools is suited for the verification problem. This concurs with our experience so far---in particular, formal treatment of other participants' responsibilities (in the RSS sense) seems to require human intervention. At the same time, in our preliminary manual verification experience in \KeYmaeraX, we see a lot of automation opportunities. Developing proof tactics dedicated to those will ease manual verification efforts.

 Formal (logical, deductive) verification of ADS safety is also pursued in~\cite{RizaldiISA18} using the interactive theorem prover Isabelle/HOL~\cite{NipkowPW02}. The work uses a white-box model of a controller, and a controller must be very simple. This is unlike RSS and the current work, which allows black-box \AC{}s and thus accommodates various real-world controllers such as sampling-based path planners (\cref{subsec:introRSSSafetyArchitecture}).

 In the presence of perceptual uncertainties (such as  errors in position measurement and object recognition), it becomes harder for  \BC{}s and \DM{}s to ensure safety. Making  \BC{}s tolerant of perceptual uncertainties is pursued in~\cite{SalayCEASW20PURSS,DBLP:conf/nfm/KobayashiSHCIK21}. One way to adapt \DM{}s is to enrich their input so that they can better detect  potential hazards. Feeding DNNs' confidence scores is proposed in~\cite{AngusCS19arxiv};  in~\cite{ChowRWGJALKC20}, it is proposed for \DM{}s to look at inconsistencies between perceptual data of different modes.


 Need to cite~\cite{BannourNC21}
 \begin{itemize}
 \item No need to cite
 \item ``Logical'' in the title refers to a level of scenarios (functional---logical---concrete)
 \item their method relies on symbolic automata and constraint solving
 \end{itemize}

 \cite{LauerS22}
 \begin{itemize}
 \item No need to cite
 \item It says ``verification'' but is about test scenario generation
 \end{itemize}
\end{auxproof}

\section{PRELIMINARIES: HYBRID PROGRAMS AND THE PROGRAM LOGIC \dHL}\label{sec:prelimdFHL}
We review the  logic \dHL{} (differential Floyd--Hoare logic), introduced in~\cite{HasuoEHDBKPZPYSIKSS23} for the purpose of 1) modeling driving situations as imperative programs  (called \emph{hybrid programs}), and 2) to reason about their goal achievement and safety.  For the framework  later in \S\ref{sec:hCFG}--\ref{sec:hoareAnnotations},  \dHL{} is  both a main inspiration and an important technical building block. 

The following collects the definitions needed in this paper. Further details are found in~\cite{HasuoEHDBKPZPYSIKSS23}.

\subsection{Hybrid Programs}\label{subsec:prelimHybridProg}

\begin{definition}[assertion]\label{def:assertion}
  Let $\Variables$ be a fixed set of variables. A \emph{term} over $\Variables$
is a rational polynomial $f\in \mathbb{Q}[\Variables]$ on $\Variables$. 
  \emph{Assertions} over $\Variables$ are generated by the grammar
  \[A,B\; ::=\; \true \mid \false \mid e \sim f \mid A \land B \mid A \lor B \mid \lnot A \mid A \limply B\]
  where $e$, $f$ are terms and
  $\sim \ \in \set{=, \leq, <, \neq}$.

 The \emph{openness} and \emph{closedness} of an assertion can be naturally defined: $x\le 2$ is closed while $1< x \land x\le 2$ is neither open nor closed, for example. See~\cite{HasuoEHDBKPZPYSIKSS23}.

\begin{auxproof}
   An assertion can be \emph{open} or \emph{closed} (or both, or none).
  \emph{Openness} and \emph{closedness} are defined
  recursively: $\true$ and $\false$ are both open and closed, $e < f$ and $e \neq f$
  are open, $e \leq f$ and $e = f$ are closed, $A \land B$ and $A \lor
  B$ are open (resp.\ closed) if both components are, $\neg A$ is open
  (resp.\ closed) if $A$ is closed (resp.\ open), and $A \limply B$ is
  open (resp.\ closed) if $A$ is closed and $B$ open (resp.\ $A$ open
  and $B$ closed).
 Note that closed assertions describe closed subsets of $\R^\Variables$.
\end{auxproof}


\end{definition}

\emph{Hybrid programs} are a combination of usual imperative programs 
and differential equations for continuous dynamics.

\begin{definition}
\label{def:hybridPrograms}
  \emph{Hybrid programs} 
  are defined  by
  \begin{align*}
  	\alpha,\beta \quad ::=\quad
  		& \skipClause{} \mid
        \alpha;\beta \mid
        \assignClause{\var}{\term} \mid
        \ifThenElse{\asserta}{\alpha}{\beta} \mid \\
      & \whileClause{\asserta}{\alpha} \mid
        \dwhileClause{\asserta}{\odeClause{\vars}{\funs}}.
  \end{align*}
   Here $x\in\Variables$ is a variable, $e$ is a term, and $A$ is an assertion.
  In $\dwhileClause{\asserta}{\odeClause{\vars}{\funs}}$, $\vars$ and
  $\funs$ are lists of the same length, respectively of (distinct)
  variables and terms, and we require that $\asserta$ be an open assertion.
  We sometimes drop the braces in
  $\dwhileClause{\asserta}{\odeClause{\vars}{\funs}}$ for readability.
\end{definition}
A
\emph{differential while} clause $\dwhileClause{\asserta}{\odeClause{\vars}{\funs}}$
 encodes the continuous dynamics according to a system $\odeClause{\vars}{\funs}$ of ODEs; it executes until the \emph{guard} condition $\asserta$ is
falsified.
Openness of $\asserta$ ensures that, if $\asserta$ is falsified at some point, then
there is the smallest time $t_{0}$ when it is falsified, at which time the command's continuation (if any) starts to execute.

The syntax of hybrid programs is inspired by $\dL$~\cite{Platzer18}, but comes with
significant changes. See~\cite{HasuoEHDBKPZPYSIKSS23} for comparison.

\begin{definition}[semantics of hybrid programs] 
  A \emph{store} $\rho\colon \Variables\to \mathbb{R}$ is a function from variables to reals.
  The \emph{value} $\sem{\term}{\store}$ of a term $\term$ in a store
  $\store$ is a real defined as usual by induction on $e$ (see for
  example~\cite[\S{}2.2]{Winskel93}).
  The \emph{satisfaction} relation between stores $\rho$ and
assertions $A$, denoted by $\store \vDash A$, is also defined as usual
  (see~\cite[\S{}2.3]{Winskel93}).

  A \emph{state} is a pair $\state{\coma}{\store}$ of a hybrid program
  and a store.
  The \emph{reduction} relation $\to$ on states is defined as usual, using several rules found in~\cite{HasuoEHDBKPZPYSIKSS23} (see \cref{ex:hybridProgSem} later). 
  A state $s$ \emph{reduces} to $\statea'$ if $\statea \red{}^*
  \statea'$, where $\red{}^*$ is the reflexive transitive closure of
  $\red{}$.
  A state $\statea$ \emph{converges} to a store $\store$, denoted by
  $\converge{\statea}{\store}$, if there exists a reduction sequence
  $\statea \red{}^* \state{\skipClause}{\store}$.
\end{definition}

\begin{example}\label{ex:hybridProgSem}
   The state $\state{\coma}{\store}$, where
\begin{math}
     \coma\, \equiv\, \left(\,\dwhileClause{\var > 0}{\odeClause{\var}{-1}} \,; \;
    \assignClause{\var}{\var-1}\right)
\end{math}
  and $\store(\var) = 2$, can reduce
1) to $\state{\coma}{\update{\store}{\var}{v}}$ for any $v \in
      (0,2]$,
2) to
      $\state{\assignClause{\var}{\var-1}}{\update{\store}{\var}{0}}$,
 and
3) to $\state{\skipClause}{\update{\store}{\var}{-1}}$.
 Only the last one corresponds to convergence (namely $\converge{\state{\coma}{\rho}}{\update{\store}{\var}{-1}}$).
\end{example}


\subsection{The Program Logic $\dHL$}\label{subsec:prelimdFHL}

\begin{definition}
\label{def:hoareQuad}
  A \emph{Hoare quadruple} is a quadruple
  $\hquad{\safetya}{\asserta}{\coma}{\assertb}$ of three assertions
  $\asserta$, $\assertb$, and $\safetya$, and a hybrid program
  $\coma$.
  It is \emph{valid} if, for all stores $\store$ such that $\store
  \vDash \asserta$,
  \begin{itemize}
    \item there exists $\store'$ such that
      $\converge{\state{\coma}{\store}}{\store'}$ and $\store' \vDash
      \assertb$, and
    \item for all reduction sequences $\state{\coma}{\store} \red{}^*
      \state{\comb}{\store'}$, $\store' \vDash \safetya$.
  \end{itemize}
\end{definition}
Hoare quadruples have safety conditions $S$ 
in addition compared to
 Hoare triples (see e.g.~\cite{Winskel93}). They  specify safety properties which must hold \emph{throughout} execution.

The semantics of Hoare quadruples (\cref{def:hoareQuad}) has some notable features such as accommodation of continuous dynamics and \emph{total correctness} semantics. See~\cite{HasuoEHDBKPZPYSIKSS23} for details.

\begin{auxproof}
 The safety of all intermediate states is ensured by the definition of
 $\statea \to \statea'$ (cf. \cref{ex:hybridProgSem}).
 The interesting case is that of the differential dynamics, where 
 $\statea'$ can
 be the state reached at any point of the dynamics.

 We note that this semantics in \cref{def:hoareQuad} is \emph{total correctness}, rather than
 \emph{partial correctness}, in that it requires termination of a hybrid program $\alpha$. This suits our purpose since in many scenarios we want to ensure goal achievement (modelled by the postcondition $B$, such as stopping in a safe position).
\end{auxproof}


The logic $\dHL$ has several \emph{derivation rules} for  Hoare quadruples. Many of them are standard Hoare logic rules (see e.g.~\cite{Winskel93}), extended in a natural manner to accommodate safety conditions. The rules for the $\dwhileKeyword$ construct come in two versions. One uses  explicit closed-form solutions; this is the one we will use. The other  uses \emph{differential (in)variants}---formulated using Lie derivatives---that correspond to barrier certificates and Lyapunov functions in control theory.
For space reasons, we refer to~\cite{HasuoEHDBKPZPYSIKSS23} for those derivation rules. They are \emph{sound}, in that they only derive valid Hoare quadruples.

\section{HYBRID CONTROL FLOW GRAPHS (hCFGs)}
\label{sec:hCFG}
\subsection{Formalization}\label{subsec:hCFGFormalization}


\begin{definition}
\label{def:hCFG}
A \emph{hybrid control flow graph (hCFG)} is a tuple $\mathcal{G}=(\Loc, \Sigma, \Edge, \Variables, \Flow, \Guard, \Assign, l_{\Init}, \Final)$ where
\begin{itemize}
 \item $\Loc 
    $ is a finite set of \emph{(control) locations};
 \item $\Sigma$ is a finite set of \emph{event names}; 
 \item $\Edge\subseteq \Loc\times\Sigma\times\Loc$ is a finite set of \emph{edges} that represent discrete ``jumps''  (we write $l\xrightarrow{a}l'$ for $(l,a,l')\in E$);
 \item $\Variables 
            $ is a finite set of real-valued variables;
 \item $\Flow$ is a function that associates, to each location $l\in L$, a system 
       \begin{math}
	\dot{\mathbf{x}}=\mathbf{f}
       \end{math} 
     of ODEs, where 1) $\mathbf{x}$ is the list (of length $|\Variables|$) of all variables in $\Variables$, and 2) $\mathbf{f}\in(\mathbb{Q}[\Variables])^{|\Variables|}$ is a list (of the  same length) of polynomials over $\Variables$; 
 \item $\Guard$ is a function that associates, to each edge $e\in E$, its \emph{guard} $A$ that is a closed assertion (\cref{def:assertion});
 \item $\Assign$ is a function that associates, to each edge $e\in E$ and a special label $\Init$ (designating initialization), a (possibly empty) list of \emph{assignment commands} that are of the form $x\coloneqq f$ where $x\in \Variables$ and $f\in \mathbb{Q}[\Variables]$; 
 \item $l_{\Init}\in L$ is an \emph{initial location}; and
 \item $\Final\subseteq L$ is a set of \emph{final locations}.

\end{itemize}
We further assume a fixed order between edges that go out of the same location. This is used later in~\cref{def:translToHybridPrograms}.
\end{definition}
hCFGs extend CFGs with flow dynamics,
much like hybrid programs extend imperative programs~\cite{HasuoEHDBKPZPYSIKSS23}. They are also similar to \emph{hybrid automata (HA)}~\cite{Henzinger96}, with a key difference that invariants and guards in HA are replaced by guards. 

\begin{wrapfigure}[5]{r}{0pt}

\vspace*{-1.5em}

\noindent
\begin{minipage}{8em}
 \begin{equation}\label{eq:edgesFromL}
\small
\begin{array}{l}
  \raisebox{-0.5\height}{\scalebox{.8}{\begin{tikzpicture}[
   grow=right,
   level distance=2cm, sibling distance=-.2cm,
   edge from parent/.style = {draw, -latex},
   edge from parent path={(\tikzparentnode) -- (\tikzchildnode)}]
 \Tree
 [.\node[draw=none]{$l$};
    \edge node[below] {$e_{l,n_{l}}$};
    [.\node[draw=none]{$l'_{l,n_{l}}$};
        ]
    \edge[draw=none];
    [.\node[]{$\vdots$};]
    \edge node[above] {$e_{l,1}$};
    [.\node[draw=none]{$l'_{l,1}$};
        ]
 ]
 \end{tikzpicture}
 }}
\\
 A_{l,i}=\Guard(e_{l,i})
\end{array} 
\end{equation}
\end{minipage}
\end{wrapfigure}
The semantics of guards in hCFGs is different from HA, too. Consider the location $l$ in~\cref{eq:edgesFromL}, where $e_{l,1}, \dotsc, e_{l,n_{l}}$ enumerate  all outgoing edges from $l$, and $A_{l,i}$ are their guards. 

Operationally, (the precise definition is later in~\cref{subsec:hCFGSemantics}) 
\begin{itemize}
 \item the execution stays at $l$---exhibiting the flow dynamics specified by $\Flow(l)$---\emph{exactly as long as} none of $A_{l,1}, \dotsc, A_{l,n_{l}}$ is satisfied; and
 \item \emph{as soon as} one of the guards is satisfied (say $A_{l,i}$), the $i$-th edge is taken and location jump occurs. 
\end{itemize} 
\begin{auxproof}
 In HA terms, one can understand that the invariant $\lnot A_{l,1}\land \cdots\land \lnot A_{l,n_{l}}$ is implicitly asserted at $l$. 
 Note that, this way, the execution of hCFGs is much more deterministic than HA.  Note also that
 the closedness requirement on guards $A_{l,i}$ (\cref{def:hCFG}) ensures that the time of location jump is well-defined. 
\end{auxproof}

\begin{auxproof}
 \begin{example}[delayed braking]
  \label{ex:obstacle}
  This example models a vehicle which comes under the need for braking at time $t=0$, cruises for $\rho$ seconds (the response time), then starts
  braking, hoping to come to a stop before it collides with a wall. We define an
  hCFG using three variables: $t$, $x$ and $v$, representing the amount of time
  elapsed and the vehicle's position and speed, respectively. The wall is at
  position $o$. The hCFG is presented in a diagrammatic form in 
  \cref{fig:delayed-stop}.
  \begin{figure}[tbp]
    \centering
    \resizebox{\columnwidth}{!}{%
      \begin{tikzpicture}[FAstyle]
    \node[state, initial, text width=1.2cm, align=center, initial text = {$\Init$, $t:=0$}] (Cruising)
      {$\mathsf{Cruising}$\\ $\dot{t} = 1$\\ $\dot{x} = v$\\ $\dot{v} = 0$};
    \node[state, right=of Cruising, text width=1.2cm, align=center] (Braking)
      {$\mathsf{Braking}$\\ $\dot{t} = 1$\\ $\dot{x} = v$\\ $\dot{v} = -1$};
    \node[state,  below=2.5cm of Cruising, text width=1.2cm, align=center, bottom color=red!20] (Crashed)
      {$\mathsf{Crashed}$};
    \node[state, accepting, right=of Crashed, text width=1.2cm, align=center] (Stopped)
      {$\mathsf{Stopped}$};
      \path[->]
        (Cruising)
          edge node[text width=2cm]{$\mathsf{Brake}$\\$t \geq \rho $} (Braking)
          edge node[text width=1cm, swap]{$\mathsf{Crash}$\\$x \geq o$} (Crashed)
        (Braking)
          edge node[text width=2cm]{$\mathsf{Stop}$\\$v \leq 0$} (Stopped)
          edge node[text width=2cm]{$\mathsf{Crash}$\\$x \geq o$} (Crashed)
        (Crashed);
      \end{tikzpicture}%
    }
    \caption{hCFG for the delayed braking situation}
    \label{fig:delayed-stop}
  \end{figure}
  Each node of this diagram represents a location of the hCFG. The nodes with a
  double border are final. Each edge is labelled with the corresponding event name
  and  guard (there are no assignment commands except for $\Init$). 
 The flow dynamics at the positions $\mathsf{Crashed}, \mathsf{Stopped}$ is omitted since they are not relevant.

 \end{example}
\end{auxproof}

\subsection{Translation to Hybrid Programs and Formal Semantics}
\label{subsec:hCFGSemantics}

\begin{definition}[hCFGs to hybrid programs]\label{def:translToHybridPrograms}
 Let $\mathcal{G}$ be an hCFG as in~\cref{def:hCFG}, where $L=\{l_{1}, \dotsc, l_{n}\}$, $l_{1}=l_{\Init}$ and $\Final=\{ l_{k+1},l_{k+2},\dotsc,l_{n}\}$. 
The hybrid program $\toProgr{\mathcal{G}}$ is
\begin{displaymath}
 \small
  \begin{array}{l}
  l \coloneqq l_{1}\,; \qquad\hfill   \text{// setting program counter to $l_{1}$\;}
  \\
  \Assign(\Init)\,; \qquad\hfill \text{// initial assignment commands\;}
  \\
  \mathsf{while}\, (l\not\in\Final) \,\{
   \hfill\text{\qquad// until reaching a final location\;}
  \\ 
   \quad      \mathsf{if} (l = l_{1}) \ \mathsf{then} \ \{ B_{1}\}\\
      \quad\mathsf{else} \ \mathsf{if} \ (l = l_{2}) \ \mathsf{then} \ \{ B_{2}\}\\
      \quad\dotsc\\
      \quad\mathsf{else} \ \mathsf{if} \ (l = l_{k}) \ \mathsf{then} \ \{ B_{k}\}\\
  \} 
   \\
   \begin{array}{r}
    \text{where}\\
     \text{$    B_{i} $ is}
   \end{array}
    \left[\footnotesize
    \begin{array}{l}
      \mathsf{dwhile}\, (\lnot A_{l_{i},1}\land \cdots\land \lnot A_{l_{i},n_{l_{i}}})\,\{\;\Flow(l_{i})\;\}\,;
      \\
      \mathsf{if} (A_{l_{i},1}) \ \mathsf{then} \ \{ \Assign(e_{l_{i},1});\; l := l'_{l_{i},1}\}\\
      \mathsf{else} \ \mathsf{if} \ (A_{l_{i},2}) \ \mathsf{then} \ \{ \Assign(e_{l_{i},2});\; l:=l'_{l_{i},2} \}\\
      \mathsf{else} \ \mathsf{if} \ \dotsc\\
      \mathsf{else} \ \mathsf{if} \ (A_{l_{i},n_{l_{i}}}) \ \mathsf{then} \ \{ \Assign(e_{l_{i},n_{l_{i}}}); \;l:=l'_{l_{i},n_{l_{i}}} \}
      \end{array}
      \right]
    \end{array}
\end{displaymath}
\end{definition}
We adopted  the notation~\cref{eq:edgesFromL} in $B_{i}$. There, the $\mathsf{dwhile}$ construct in the first line executes the specified flow dynamics as long as $\lnot A_{l_{i},1}\land \cdots\land \lnot A_{l_{i},n_{l_{i}}}$  is true. Once we leave the flow dynamics, we take a transition $l_{i}\to l'_{l_{i},j}$ whose guard $A_{l_{i},j}$ is satisfied; when there are multiple such,  we use the fixed order of edges (assumed in \cref{def:hCFG}) to resolve the possible nondeterminism, in the way explicated in the $\mathsf{if}\ \cdots \mathsf{else}\ \mathsf{if}\ \cdots$ construct in $B_{i}$.

\begin{definition}[semantics of hCFGs]\label{def:semanticsOfhCFGs}
 Let $\mathcal{G}$ be an hCFG. Its \emph{operational semantics} is defined to be that of the hybrid program $\toProgr{\mathcal{G}}$, where the latter is defined in \cref{subsec:prelimHybridProg}.
\end{definition}

\subsection{Networks of hCFGs}
\label{subsec:networkOfHCDGs}
The following formalism enables compositional modeling. It is much like \emph{networks of timed automata} (see e.g.~\cite{BengtssonY03}) although ours features richer interaction, as  explained later.
\begin{auxproof}
 Many hCFGs of our interest  arise as a combination of multiple agents. For example, an intersection is naturally modeled as a combination of {\SV} 
 and {\POV{}}.
 It is therefore useful if we can write \emph{component hCFGs} for individual agents and let them collectively represent the whole system. Here we introduce a formalism for doing so. It is much like \emph{networks of timed automata} (see e.g.~\cite{BengtssonY03}) although ours features richer interaction (further explanation will come later).
\end{auxproof}

\begin{definition}[network of hCFGs]
 \label{def:networkOfHCFG}
An \emph{open hCFG} is defined in the same way as hCFGs (\cref{def:hCFG}), with the  difference that 1) the flow dynamics $\Flow(l)\equiv(\dot{\mathbf{x}=\mathbf{f}})$ may not cover all variables in $\Variables$ (that is, the length of the list $\mathbf{x}$  may be $< |\Variables|$), and 2) it has no final locations specified. 

A \emph{network of hCFGs} is a tuple $\mathcal{N}=(\mathcal{G}^{(1)},\dotsc, \mathcal{G}^{(k)})$ of open hCFGs (these are called \emph{component hCFGs} of $\mathcal{N}$) such that 1) their location sets $L^{(1)},\dotsc, L^{(k)}$ are disjoint, 
2) they share the same set $\Variables$ of  variables, 
and 3) they satisfy so-called \emph{compatibility conditions}, deferred to 
\cref{appendix:compatCond},
requiring that there are no conflicts in the descriptions of dynamics in different hCFGs. 

\end{definition}
Intuitively, a variable can be changed (by $\Flow$ or $\Assign$) by only one component, but it can be seen (i.e.\ occur in guards) in multiple components.
This ``can't change but can see'' interaction---essential in modeling driving situations---is not present in networks of timed automata~\cite{BengtssonY03}. A consequence of this feature is that a component CFG does not have its own semantics: its behavior depends on variables changed by other components. This does not matter for our purpose (compositional modeling, not compositional verification).

The semantics of a network of hCFGs is defined by the synchronized product of  component hCFGs. There are two modes of synchronization: by shared variables (in  guards), and by shared events (in edge labels). Below is an informal definition; a formal definition is in
\cref{appendix:semNetHCFG}.
\begin{definition}\label{def:semNetworkOfHCFG}
 The \emph{semantics} of a network  $\mathcal{N}=(\mathcal{G}^{(1)},\dotsc, \mathcal{G}^{(k)})$ of hCFGs is defined to be the semantics (\cref{def:semanticsOfhCFGs}) of its \emph{synchronized product} $\mathcal{G}_{\mathcal{N},\Final}$ with a given set $\Final$ of final locations ($\Final$ is specified separately). The hCFG $\mathcal{G}_{\mathcal{N},\Final}$ is defined in a usual manner. Its location is a tuple $(l^{(1)},\dotsc, l^{(k)})$ of locations of components. When an event name $a$ is shared by multiple components, we require that \emph{all} these components  synchronize. The flow dynamics, guards, and assignment commands are defined by suitably joining those of components. 
\end{definition}

\section{THE INTERSECTION SITUATION: MODELING}\label{sec:intersectionModeling}
\todoil{use the new model}
Here we describe our hCFG modeling of the main case study, namely the \emph{intersection} driving situation. 

\subsection{The Intersection Situation, Informally}\label{subsec:intersectionInformally}
\todoil{(Ichiro) James, can you name the vehicles in the figure? I think you can call them OV1, ..., OV4.}
The driving situation  is illustrated in \cref{fig:intersection}. There is an intersection with traffic lights, and the lights are green for the vertical road---both upwards and downwards---and they are red for the horizontal one. 
There may be many vehicles, but we assume that only the following two are relevant.
\begin{description}
 \item[\SV{}] The \emph{subject vehicle} making a right turn.
 \item[\POV{}] The \emph{principal other vehicle} is driving straight downwards, potentially causing a collision with \SV{}.
\end{description}
The other  vehicles  in \cref{fig:intersection} are irrelevant, physically (far  away) or by responsibility (red traffic light or  behind \SV{}). 

\begin{auxproof}
 for the following reasons.
 1) The one on the left is irrelevant since it is supposed to follow the red traffic light. It should not  come into the intersection.
 2) The one behind SV is irrelevant since it is this vehicle's responsibility to avoid back collision to \SV{}, and its potential collision with \POV{} is none of \SV{}'s concern.
 3) The leftward-facing one  on the right  is irrelevant for the same reason as the one on the left.
 4) The rightward-facing vehicle on the right is assumed to be far ahead of \SV{} so that \SV{} will never back-collide into it. 
 To be precise, we assume that this vehicle is
 at least the RSS safety distance ahead of \SV{} (\cref{ex:RSSSafetyDistance}). Since we assume (as described below) that \SV{}'s control is nothing but the RSS condition in \cref{ex:RSSSafetyDistance}, \SV{} is guaranteed not to back-collide into this vehicle. 
\end{auxproof}

The problem of this case study is as follows. \emph{We focus on one specific minimum risk maneuver (MRM), namely for \SV{} to stop as quickly as possible. We ask if this MRM is safe.} In RSS terms (\cref{subsec:introRSS}), the problem is to compute an RSS condition for the max-brake proper response.

\begin{auxproof}
  This problem can be stated as follows, more rigorously using  RSS terms. We focus on a single proper response (namely the max-brake one), and we would like to compute its RSS condition---a condition that, if it is true at the beginning, guarantees the safe completion of that specific proper response. This RSS condition should be expressed as a logical assertion over the positions  and velocities of \SV{}, \POV{}.
\end{auxproof}

Note that  this case study is only part of the  safety analysis of the intersection scenario in \cref{fig:intersection}. There are other conceivable proper responses, such as accelerating to leave the intersection quickly. For each such proper response, we can conduct a similar analysis to find its RSS condition. A catalogue of proper responses and their RSS conditions, obtained this way, can be used for the maneuver planning of \SV{}: we can choose a proper response whose RSS condition is  true; we can use some preference if there are multiple such. This conditional combination of proper responses is what is done in~\cite[\S{}IV-E]{HasuoEHDBKPZPYSIKSS23}.

We assume that \SV{} and \POV{} maintain their lanes (which we can confirm by their turn signals off); therefore only two lanes are relevant to us.

\subsection{An hCFG Modeling}
The intersection situation exhibits complex interaction between \SV{} and \POV{}.
 Such interaction can be \emph{causal} (if  \SV{} is in the collision zone then \POV{} must act to avoid collision) or \emph{temporal} (\SV{} may enter the collision zone before \POV{}, after it, or simultaneously). We use a network of hCFGs (\cref{subsec:networkOfHCDGs}) to deal with the complication.


\begin{auxproof}
 \todoil{(Ichiro) I wrote the following to explain the current model, and it should be fine for the time being. However, one can ask the following questions; a refined analysis is desired to address them.
 \begin{itemize}
 \item \POV{} has the precedence in this intersection scenario. Thus perhaps ``POV reacting to SV in the CZ'' is not relevant.
 \end{itemize}
 }

 \todoil{Simplify writing}
 For one thing, there is a causal relationship between the behaviors of \SV{} and \POV{}.
 In RSS, every traffic participant is bound by a duty of care; in the current specific driving scenario, \POV{} is required to engage braking to avoid collision when there is another vehicle (such as \SV{}) ahead of it. This way \SV{}'s behavior can affect \POV{}'s, which in turn may  help \SV{} itself avoiding a collision. 

 For another, our modeling encompasses reaction time $\rho$, which is needed for realistic analysis and is done e.g.\ in~\cite{ShalevShwartzSS17RSS} (see \cref{ex:RSSSafetyDistance}). This complicates the temporal relationship between events on the \SV{} side and those on the \POV{} side. 
\end{auxproof}

\begin{auxproof}
 Therefore, it helps our modeling efforts if we can describe the two agents (\SV{} and \POV{}) separately, and let them synchronize in a suitable manner and automatically generate the whole system model, instead of manually resolving all the interaction and handcrafting the whole system model. The formalism of networks of hCFGs (\cref{subsec:networkOfHCDGs}) allows us to do so.
\end{auxproof}

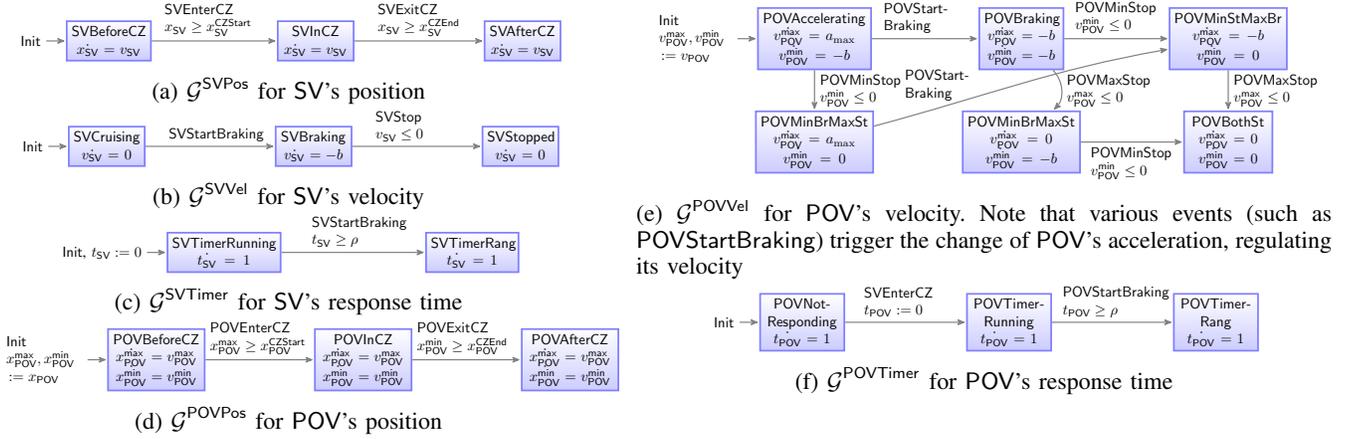
\begin{figure*}[tbp]
 \centering
\begin{minipage}[t]{.44\textwidth}

\vspace{-7em}
 \begin{subfigure}[t]{\textwidth}
 \centering
 \scalebox{.55}{%
    \begin{tikzpicture}[FAstyle]
          \node[state, initial, text width=1.8cm, align=center, initial text = {$\Init$}] (BeforeCZ)
            {$\mathsf{SVBeforeCZ}$\\$\dot \xSV = \vSV$};
          \node[state, right=of BeforeCZ, text width=1.6cm, align=center] (InCZ)
            {$\mathsf{SVInCZ}$\\$\dot \xSV = \vSV$};
          \node[state, right=of InCZ, text width=1.6cm, align=center] (AfterCZ)
            {$\mathsf{SVAfterCZ}$\\$\dot \xSV = \vSV$};
            \path[->]
              (BeforeCZ)
                edge node[text width=2.4cm]{$\mathsf{SVEnterCZ}$\\$\xSV \geq \CZStartSV $} (InCZ)
              (InCZ)
                edge node[text width=2cm]{$\mathsf{SVExitCZ}$\\$\xSV \geq \CZEndSV$} (AfterCZ);
        \end{tikzpicture}%
        }
  \caption{$\mathcal{G}^{\mathsf{SVPos}}$ for \SV{}'s position}
  \label{fig:hCFGSVPos}
 \end{subfigure}
 \begin{subfigure}[t]{\textwidth}
 \centering
 \scalebox{.55}{
          \begin{tikzpicture}[FAstyle]
          \node[state, initial, text width=1.6cm, align=center, initial text = {$\Init$}] (Any)
          {$\mathsf{SVCruising}$\\$\dot\vSV = 0$};
          \node[state, right=of BeforeCZ, text width=1.6cm, align=center] (Braking)
            {$\mathsf{SVBraking}$\\$\dot\vSV = -b$};
          \node[state, right=of InCZ, text width=1.6cm, align=center] (Stopped)
            {$\mathsf{SVStopped}$\\$\dot\vSV = 0$};
            \path[->]
              (Any)
                edge node[text width=2cm]{$\mathsf{SVStartBraking}$} (Braking)
              (Braking)
                edge node[text width=2cm]{$\mathsf{SVStop}$\\$\vSV \le 0$} (Stopped);
              \end{tikzpicture}
        }
  \caption{$\mathcal{G}^{\mathsf{SVVel}}$ for \SV{}'s velocity}
  \label{fig:hCFGSVVel}
 \end{subfigure}
 \begin{subfigure}[t]{\textwidth}
 \centering
  \scalebox{.55}{
    \begin{tikzpicture}[FAstyle]
          \node[state, initial, text width=2.5cm, align=center, initial text = {$\Init$, $\tSV :=0$}] (Running)
            {$\mathsf{SVTimerRunning}$\\$\dot\tSV  = 1$};
          \node[state, right=6cm of BeforeCZ, text width=2cm, align=center] (Rang)
          {$\mathsf{SVTimerRang}$\\$\dot\tSV = 1$};
            \path[->]
              (Running)
              edge node[text width=2cm]{$\mathsf{SVStartBraking}$\\ $\tSV \ge \rho$} (Rang);
            \end{tikzpicture}
  }
  \caption{$\mathcal{G}^{\mathsf{SVTimer}}$ for \SV{}'s response time}
  \label{fig:hCFGSVTimer}
 \end{subfigure}
\begin{subfigure}[t]{\textwidth}
 \centering
  \scalebox{.55}{%
    \begin{tikzpicture}[FAstyle]
          \node[state, initial, text width=2.1cm, align=center, initial text = {$	   \begin{array}{l}
	    \Init\\
            \xPOVmax, \xPOVmin \\\,:= \xPOV
	   \end{array}$}] (BeforeCZ)
            {$\mathsf{POVBeforeCZ}$\\$\dot\xPOVmax = \vPOVmax$\\$\dot\xPOVmin = \vPOVmin$};
          \node[state, right=of BeforeCZ, text width=2.1cm, align=center] (InCZ)
            {$\mathsf{POVInCZ}$\\$\dot\xPOVmax = \vPOVmax$\\$\dot\xPOVmin = \vPOVmin$};
          \node[state, right=of InCZ, text width=2.1cm, align=center] (AfterCZ)
            {$\mathsf{POVAfterCZ}$\\$\dot\xPOVmax = \vPOVmax$\\$\dot\xPOVmin = \vPOVmin$};
            \path[->]
              (BeforeCZ)
                edge node[text width=2.4cm]{$\mathsf{POVEnterCZ}$\\$\xPOVmax \ge \CZStartPOV$} (InCZ)
              (InCZ)
                edge node[text width=2.4cm]{$\mathsf{POVExitCZ}$\\$\xPOVmin \ge \CZEndPOV$} (AfterCZ);
        \end{tikzpicture}%
        }
  \caption{$\mathcal{G}^{\mathsf{POVPos}}$ for \POV{}'s position}
  \label{fig:hCFGPOVPos}
\end{subfigure}
\end{minipage}
\hfill
\begin{minipage}[t]{.52\textwidth}
\begin{subfigure}[t]{\textwidth}
 \centering
 \scalebox{.55}{
          \begin{tikzpicture}[FAstyle]
          \node[state, initial, text width=2.5cm, align=center, initial text = {$
	   \begin{array}{l}
	    \Init\\
            \vPOVmax,\vPOVmin \\\,:= \vPOV
	   \end{array}$}] (Any)
          {$\mathsf{POVAccelerating}$\\$\dot\vPOVmax = \amax$\\$\dot\vPOVmin = -b$};
          \node[state, below=2.5cm of Any, text width=2.6cm, align=center] (MaxAcceleratingMinStopped)
            {$\mathsf{POVMinBrMaxSt}$\\$\dot\vPOVmax = \amax$\\$\dot\vPOVmin = 0$};
          \node[state, right=of BeforeCZ, text width=1.8cm, align=center] (Braking)
            {$\mathsf{POVBraking}$\\$\dot\vPOVmax = -b$\\$\dot\vPOVmin = -b$};
          \node[state, right=of InCZ, text width=2.6cm, align=center] (MinStoppedMaxBraking)
            {$\mathsf{POVMinStMaxBr}$\\$\dot\vPOVmax = -b$\\$\dot\vPOVmin = 0$};
          \node[state, below=2.5cm of MinStoppedMaxBraking, text width=1.9cm, align=center] (BothStopped)
            {$\mathsf{POVBothSt}$\\$\dot\vPOVmax = 0$\\$\dot\vPOVmin = 0$};

            \path[->]
              (Any)
                edge node[text width=2cm]{$\mathsf{POVStart}$-$\mathsf{Braking}$} (Braking)
              (Braking)
                edge node[text width=2cm]{$\mathsf{POVMinStop}$\\$\vPOVmin \le 0$} (MinStoppedMaxBraking)
              (MinStoppedMaxBraking)
                edge node[text width=2cm]{$\mathsf{POVMaxStop}$\\$\vPOVmax \le 0$} (BothStopped);

          \node[state, below=2.5cm of Braking, text width=2.6cm, align=center] (MaxStoppedMinBraking)
            {$\mathsf{POVMinBrMaxSt}$\\$\dot\vPOVmax = 0$\\$\dot\vPOVmin = -b$};
            \path[->]
	      (Any)
                edge node[text width=2cm]{$\mathsf{POVMinStop}$\\$\vPOVmin \le 0$} (MaxAcceleratingMinStopped)
	      (MaxAcceleratingMinStopped)
                edge [out=15, in=190] node[text width=2cm, pos=0.23, above]{$\mathsf{POVStart}$-\\$\mathsf{Braking}$} (MinStoppedMaxBraking);

	   \path[->]
	     (Braking)
                edge [out=315,in=45] node[text width=2cm, right]{$\mathsf{POVMaxStop}$\\$\vPOVmax \le 0$} (MaxStoppedMinBraking)
              (MaxStoppedMinBraking)
                edge node[text width=2cm,below]{$\mathsf{POVMinStop}$\\$\vPOVmin \le 0$} (BothStopped);
              \end{tikzpicture}
        } 
  \caption{$\mathcal{G}^{\mathsf{POVVel}}$ for \POV{}'s velocity. Note that various events (such as $\mathsf{POVStartBraking}$) trigger the change of \POV{}'s acceleration, regulating its velocity}
  \label{fig:hCFGPOVVel}
\end{subfigure}
\begin{subfigure}[t]{\textwidth}
 \centering
  \scalebox{.55}{
        \begin{tikzpicture}[FAstyle]
         \node[state, initial, text width=1.8cm, align=center, initial text = {$\Init$}] (NotResponding)
            {$\mathsf{POVNot}$-\\$\mathsf{Responding}$\\$\dot\tPOV = 1$};
          \node[state, right=of NotResponding, text width=1.8cm, align=center] (Running)
            {$\mathsf{POVTimer}$-\\$\mathsf{Running}$\\$\dot\tPOV = 1$};
          \node[state, right=of Running, text width=1.8cm, align=center] (Rang)
          {$\mathsf{POVTimer}$-\\$\mathsf{Rang}$\\$\dot\tPOV = 1$};
          \path[->]
            (NotResponding) edge node[text width=2cm]{$\mathsf{SVEnterCZ}$\\ $\tPOV := 0$} (Running)
              (Running)
              edge node[text width=2.4cm]{
	                                $\mathsf{POVStartBraking}$
                                        \\ $\tPOV \ge \rho$} (Rang);
            \end{tikzpicture}
  }
  \caption{$\mathcal{G}^{\mathsf{POVTimer}}$ for \POV{}'s response time}
  \label{fig:hCFGPOVTimer}
\end{subfigure}
\end{minipage}
\caption{the network $\mathcal{N}^{\Int}$ of hCFGs for the intersection situation, consisting of six open hCFGs;  final locations are as in~\cref{eq:intersectionFinalLocation}}
\label{fig:intersectionHCFG}
\end{figure*}

Concretely, our model is the network of hCFGs presented in~\cref{fig:intersectionHCFG}. 
 We use the following variables and constants.
\begin{itemize}
 \item The positions and velocities of \SV{} and \POV{} are expressed in the lane coordinates, for the lanes shown in \cref{fig:intersection}. We use the variables $\xSV,\xPOV,\vSV,\vPOV$ for them. \POV{}'s behavior has nondeterminism: it can accelerate, brake, or cruise, unless it brakes to avoid collision. We model two extremes, using the variables $\xPOVmax, \xPOVmin$, etc.
 \item The region in which these two lanes intersect is called the \emph{collision zone} (\CZ).
  \CZ{} is expressed by intervals in the lane coordinates, namely by $[\CZStartSV,\CZEndSV]$ and $[\CZStartPOV,\CZEndPOV]$, respectively.
 \item The maximum braking rate $b>0$ is a constant.
\begin{auxproof}
  and shared by \SV{} and \POV{}. We assume that this is the maximum comfortable braking rate (as opposed to the emergency one, see \cref{ex:RSSSafetyDistance}), meaning that we are after an RSS condition that allows collision avoidance only relying on comfortable braking. 
\end{auxproof} 
\item For dealing with the response time $\rho$ (that is a constant), \SV{} and \POV{} have their \emph{timer variables} $\tSV,\tPOV$.
\end{itemize}

Let us go into each component open hCFG. 

 The components
 $\mathcal{G}^{\mathsf{SVPos}},
 \mathcal{G}^{\mathsf{POVPos}}
 $
 describe the positions, with different locations ($\mathsf{SVBeforeCZ}, \mathsf{SVInCZ}$, etc.) designating the vehicles' (discrete) position relative to \CZ{}. The ODE for the flow dynamics stays the same in each component, where the values of $\vSV,\vPOV$ are governed by other components. Note that $\mathsf{POVInCZ}$ is modeled conservatively: it starts when the fastest possible \POV{} reaches \CZ{}; it ends when the slowest possible leaves \CZ{}. 

 The components
 $\mathcal{G}^{\mathsf{SVVel}},
 \mathcal{G}^{\mathsf{POVVel}}
 $ describe the velocities. Here, we have multiple locations to accommodate different flow dynamics (i.e.\ different acceleration). The events $\mathsf{SVStartBraking}, \mathsf{POVStartBraking}$ come with no explicit guards, but they are implicitly guarded by the conditions $\tSV\ge \rho, \tPOV\ge \rho$ for the same events in the timer components  $\mathcal{G}^{\mathsf{SVTimer}},
 \mathcal{G}^{\mathsf{POVTimer}}$. Recall from \cref{def:semNetworkOfHCFG} that components must synchronize upon shared events.


The component $ \mathcal{G}^{\mathsf{POVVel}}$ is  more complex than
 $ \mathcal{G}^{\mathsf{SVVel}}$ since  two extreme \POV{}s may stop at different moments. It is semantically clear that the location $\mathsf{POVMinBrMaxSt}$ is vacuous. We include it so that the modeling is systematic;  our reasoning algorithm indeed detects that it is vacuous.

The components
$\mathcal{G}^{\mathsf{SVTimer}},
 \mathcal{G}^{\mathsf{POVTimer}}$ govern the timer variables.
The one for \SV{} ticks from the beginning, counting up to $\rho$, at which the event $\mathsf{SVStartBraking}$ is enabled in $\mathcal{G}^{\mathsf{SVTimer}}$, hence in $\mathcal{G}^{\mathsf{SVVel}}$  by synchronization. 
In contrast, the one for \POV{} starts counting
only when the event $\mathsf{SVEnterCZ}$ occurs, which must synchronize 
with the same event 
in $\mathcal{G}^{\mathsf{SVPos}}$. 
This way,
$\mathcal{G}^{\mathsf{POVTimer}}$ starts ticking 
when it sees \SV{} in \CZ{}.

\begin{auxproof}
 Overall, we have three components describing \SV{}, and three for \POV{}. The event names $\mathsf{SVStartBraking}$, $\mathsf{POVStartBraking}$, $\mathsf{SVEnterCZ}$ are shared by multiple components for the synchronization purpose. 
\end{auxproof}

Most of the component hCFGs are straightline and they may seem deterministic. Nevertheless, the whole system comes with nondeterminism due to the product construction---for example, at the initial state, the next event can be one of $\mathsf{SVEnterCZ}$, $\mathsf{POVEnterCZ}$, 
etc., depending on the values of variables.

We note that each component is simple and clearly presents its intuition. It is also easy to compare corresponding components for different vehicles, manifesting their similarities and differences. These features of hCFGs greatly aid  \emph{explainability} and \emph{accountability} of the modeling process.

The final locations are defined as follows.
\begin{equation}\label{eq:intersectionFinalLocation}
\small
\begin{array}{l}
  \Final^{\Int}
 \;:=\;
 \bigl\{\,(l^{\mathsf{SVPos}}, 
 \dotsc,
 l^{\mathsf{POVTimer}})\,\big|\,
\\
\; l^{\mathsf{SVPos}} = \mathsf{SVAfterCZ} 
 \lor
 l^{\mathsf{POVPos}} = \mathsf{POVAfterCZ}
\\
\;
 \lor
 (l^{\mathsf{SVVel}} = \mathsf{SVStopped} \land
 l^{\mathsf{POVVel}} = \mathsf{POVBothSt} )
 \,
 \bigr\}
\end{array}
\end{equation}
  In all these cases there is no possibility of further collision, or the
  situation will no longer evolve.

\section{REASONING OVER hCFGs}\label{sec:hoareAnnotations}
\begin{definition}[Hoare annotation]\label{def:hoareAnnotation}
Let $\mathcal{G}$ be an hCFG in \cref{def:hCFG}. We adopt the notations in
 \cref{def:translToHybridPrograms} 
for edges,  etc.

A \emph{Hoare annotation} $\gamma$ for $\mathcal{G}$ is a function that associates, to each location $l$ of $\mathcal{G}$, an assertion $\gamma(l)$ over $\Variables$ (cf.\ \cref{def:assertion}).

Let $S$ be an assertion, and $\Unsafe$ be a set of locations.  A Hoare annotation $\gamma$ for $\mathcal{G}$ is  \emph{valid under a safety condition $S$ and unsafe locations $\Unsafe$} if the following holds.

If $l\in\Final$, then the  implication $\gamma(l)\Rightarrow S$ is valid.

If $l\in\Unsafe$, then  $\gamma(l)\Rightarrow \false$ is valid.

Otherwise,  let $l\in L\setminus (\Final\cup\Unsafe)$ be a location that is not final nor unsafe; let its successors and guards denoted as in~\cref{eq:edgesFromL}. Then there exist assertions $C_{l,1},\dotsc, C_{l,n_{l}}$ such that
\begin{itemize}
 \item the Hoare quadruple
       \begin{equation}\label{eq:hoareAssertionValidityQuadruple}
\begin{array}{r}
 	\Bigl\{C_{l,i}\Bigr\} 
	\;
 \left(
{\scriptsize \begin{array}{l}
 	\mathsf{dwhile}\, (\lnot A_{l,i})\,\{
	 \\\quad\Flow(l)
         \\\}
\end{array}}      
 \right)
\, 
	\Bigl\{\,
 \gamma(l'_{l,i})\,\bigl[\,\Assign(e_{l,i})\,\bigr]
        \,\Bigr\} 
\\\qquad\quad
	\;:\; S \land \bigwedge_{j\in [1,n_{l}]\setminus\{i\}} \lnot A_{l,j}
\end{array}       
\end{equation}
       is valid for each $i\in [1,n_{l}]$, where $\gamma(l'_{l,i})\,\bigl[\,\Assign(e_{l,i})\,\bigr]$ is the assertion defined below, and

 \item
 the   implication $\gamma(l)\Rightarrow C_{l,1}\lor\cdots\lor C_{l,n_{l}}$ is valid.
\end{itemize}

\begin{auxproof}
 The safety condition $S \land \bigwedge_{j\in [1,n_{l}]\setminus\{i\}} \lnot A_{l,j}$ is too restrictive when multiple guards (say $i,i'$) are enabled for the first time simultaneously and $i$ is the smallest index for such. However, dropping $i'$ from the conjunct makes the Hoare quadruple too weak, allowing the ``leak'' to the $i'$-th edge before the $i$-th guard is enabled. This is another occurrence of the general closure problem.

Perhaps one possible uniform solution is that, for $\{A\}\alpha\{B\}\colon S$, we require $S$ to be true at all points of execution of $\alpha$ \emph{except for} the very final state. If we want $S$ at the final state, then we can include it in the postcondition $S$. Doing so would require changing the (LImp) rule, etc.
\end{auxproof}
In~\cref{eq:hoareAssertionValidityQuadruple}, the assertion $\gamma(l'_{l,i})\,\bigl[\,\Assign(e_{l,i})\,\bigr]$ is obtained from $\gamma(l'_{l,i})$ by the substitution specified in $\Assign(e_{l,i})$. That is, letting $\Assign(e_{l,i})\equiv (x_{1}:= a_{1};\, \dotsc\,;\, x_{k}:= a_{k})$, we define
\begin{math}
   \gamma(l'_{l,i})\,\bigl[\,\Assign(e_{l,i})\,\bigr]
 :\equiv
 \gamma(l'_{l,i})\,[a_{k}/x_{k}]\,[a_{k-1}/x_{k-1}]\,\cdots\,[a_{1}/x_{1}].
\end{math}
\end{definition}

Some explanations are in order. 

The basic idea is that $\gamma$ is an \emph{invariant}: if  true now, then it is true again in the next location. Precisely, if $\gamma(l)$ is true when   $l$ is reached, then $\gamma(l')$ should be true at the ``jump'' moment that the successor $l'$ is reached. The definition is so that $\gamma$'s truth guarantees the (global) safety condition $S$ and the avoidance of unsafe locations. The condition $S$ is true throughout the flow dynamics, too, since $S$ is in the quadruple~\cref{eq:hoareAssertionValidityQuadruple}.


\begin{auxproof}
 The invariant $\gamma$ ensures that the execution never reaches an unsafe location $l\in\Unsafe$, too, by the condition $\gamma(l)\Rightarrow \false$. (We note that avoidance of unsafe locations can be encoded in $S$ by introducing new variables. However, explicit unsafe locations aid presentation.)
\end{auxproof}

Therefore, for $\gamma$, we have that 1)  it is an invariant, and 2) its truth guarantees  truth of $S$ and avoidance of $\Unsafe$. Consequently, if $\gamma$ holds initially, 
then $S$ is true and $\Unsafe$ is avoided all the time. This is stated formally in \cref{thm:validityAndSemantics}.

 Branching according to guards requires some care (cf.\ \cref{eq:edgesFromL}); this is why we split $\gamma(l)$ into \emph{edge-wise preconditions} $C_{l,i}$.
 The assertion $C_{l,i}$ ensures the following, since~\cref{eq:hoareAssertionValidityQuadruple} is valid.
\begin{itemize}
 \item The $i$-th edge $e_{l,i}$ is taken after the flow dynamics $\Flow(l)$  (due to $\bigwedge_{j\in [1,n_{l}]\setminus\{i\}} \lnot A_{l,j}$ in the safety condition of~\cref{eq:hoareAssertionValidityQuadruple}).
 \item Moreover, after taking the edge $e_{l,i}$ and executing 
the associated assignment 
 $\Assign(e_{l,i})$,  $\gamma(l'_{l,i})$ holds at the successor $l'_{l,i}$.  This is because the assertion $\gamma(l'_{l,i})[\Assign(e_{l,i})]$ in~\cref{eq:hoareAssertionValidityQuadruple} is defined so that the quadruple 
\begin{math}
  \bigl\{\,\gamma(l'_{l,i})\,\bigl[\,\Assign(e_{l,i})\,\bigr]\,\bigr\}
 \;\Assign(e_{l,i})\;
 \bigl\{\,\gamma(l'_{l,i})\,\bigr\} \;:\, S
\end{math} is valid (confirmed easily by the \textsc{(Assign)} rule in~\cite{HasuoEHDBKPZPYSIKSS23}).
\end{itemize}  
\begin{auxproof}
 For the latter point, we note that 
 the assertion $\gamma(l'_{l,i})\,\bigl[\,\Assign(e_{l,i})\,\bigr]$  defined in~\cref{eq:substAssert} makes the following Hoare quadruple  valid:
 \begin{equation}\label{eq:substAssertQuadruple}
 \Bigl\{\,\gamma(l'_{l,i})\,\bigl[\,\Assign(e_{l,i})\,\bigr]\,\Bigr\}
 \;\Assign(e_{l,i})\;
 \Bigl\{\,\gamma(l'_{l,i})\,\Bigr\} \;:\, S.
 \end{equation}
 The validity of \cref{eq:substAssertQuadruple} follows from the \textsc{(Assign)} rule (see~\cite{HasuoEHDBKPZPYSIKSS23}), using  the validity of $\gamma(l'_{l,i})[\Assign(e_{l,i})]\Rightarrow S$ (by~\cref{eq:hoareAssertionValidityQuadruple}) and $\gamma(l'_{l,i})\Rightarrow S$ (shown below).


We also note that, throughout any execution that respects $\gamma$, the safety condition $S$ is true all the time. It is true during the flow dynamics thanks to the conjunct $S$ in the safety condition in~\cref{eq:hoareAssertionValidityQuadruple}. For jump moments, note first that we have $C_{l,i}\Rightarrow S$ hold for each $i$, due to~\cref{eq:hoareAssertionValidityQuadruple} and the definition of validity of Hoare quadruples (see \cref{def:hoareQuad}; if $\{A\}\alpha\{B\}\colon S$ is valid then so is $A\Rightarrow S$). Therefore $\gamma(l)\Rightarrow S$ for each non-final $l$. For a final location $l$, $\gamma(l)\Rightarrow S$ is explicitly required in~\cref{def:hoareAnnotation}.
\end{auxproof}

In particular,
when a non-final location $l$ has no outgoing edge, then $\gamma(l)=\false$, since there is no
 $C_{l,i}$ available.
\begin{auxproof}
  (such as $\mathsf{Crashed}$ in~\cref{fig:delayed-stop}), there is no $C_{l,i}$ thus $C_{l,1}\lor\cdots\lor C_{l,n_{l}}$ is naturally defined to be $\false$. This forces $\gamma(l)=\false$, too, due to $\gamma(l)\Rightarrow C_{l,1}\lor\cdots\lor C_{l,n_{l}}$.
\end{auxproof}

The following result semantically justifies 
 \cref{def:hoareAnnotation}. Its proof follows from the discussions so far. We need acyclicity of $\mathcal{G}$ for ensuring termination of $\mathcal{G}$'s execution (but not for safety).
\begin{auxproof}
 ; acyclicity is obvious in many driving situations.\footnote{One notable exception is a roundabout, for whose termination we need another mechanism such as a ranking function over pairs of a location and a store. We note that, even without acyclicity, the quadruple~\cref{eq:validityAndSemantics} is valid in the partial correctness sense and thus the safety condition $S$ is satisfied.}
\end{auxproof}
\begin{theorem}[semantic validity]\label{thm:validityAndSemantics}
Let $\gamma$ be a valid Hoare annotation for an hCFG $\mathcal{G}$ under  $S$ and $\Unsafe$. Assume that $\mathcal{G}$ is acyclic, that is, that there is no cycle $l_{1}\xrightarrow{a_{1}}\cdots \xrightarrow{a_{k}} l_{k+1}$ in the graph $(L,E)$. Then the Hoare quadruple
\begin{equation}\label{eq:validityAndSemantics}
\begin{array}{r}
  \Bigl\{\,\gamma(l_{\Init})\,\bigl[\,\Assign(\Init)\,\bigr]\,\Bigr\}\;
 [\mathcal{G}]\;
 \Bigl\{\true\Bigr\} \;
\\\quad
 \colon\; S \land\bigwedge_{l_{i}\in \Unsafe}\lnot(l=l_{i})
\end{array}
\end{equation}
is valid in \dHL{}. Here $[\mathcal{G}]$ is the translation in \cref{def:translToHybridPrograms}, $l$ is the program counter variable in $[\mathcal{G}]$, and $\gamma(l_{\Init})\,[\Assign(\Init)]$ is defined by substitution much like in~\cref{def:hoareAnnotation}.  \qed
\end{theorem}

\begin{auxproof}
 \section{THE INTERSECTION SITUATION: REASONING}\label{sec:intersectionReasoning}
 Here we use the reasoning formalism in \cref{sec:hoareAnnotations} in our intersection case study. Its model was given 
 in \cref{sec:intersectionModeling}. 

 To specify safety requirements, we use the following definitions, identifying a collision as a joint occupancy of  \CZ{}.
 \begin{displaymath}
 \begin{array}{l}
  S^{\Int} \;:=\; \true
 \\
    \Unsafe^{\Int}
 \;:=\;
 \bigl\{\,(l^{\mathsf{SVPos}}, 
 \dotsc,
 l^{\mathsf{POVTimer}})\,\big|\,
 \\
 \qquad l^{\mathsf{SVPos}} = \mathsf{SVInCZ} 
 \land
 l^{\mathsf{POVPos}} = \mathsf{POVInCZ} 
 \,
 \bigr\}
 \end{array}
 \end{displaymath}
 The  hCFG $\mathcal{G}_{\mathcal{N}^{\Int},\Final^{\Int}}$ models the driving situation, and in particular the max-brake proper response. We are after its RSS condition, that is, a precondition that ensures safety, the latter being an assertion over the variables $\xSV,\xPOV,\vSV,\vPOV$. By \cref{thm:validityAndSemantics}, we look for 
 a valid Hoare annotation $\gamma$. We solve this problem by the symbolic algorithm we implemented, discussed in \cref{sec:impl}.

 \begin{auxproof}
 Now our goal is to find a valid Hoare annotation $\gamma$ for the hCFG $\mathcal{G}_{\mathcal{N}^{\Int},\Final^{\Int}}$ under $S^{\Int}$ and $\Unsafe^{\Int}$. We are especially interested in the assertion $\gamma(l^{\Int}_{\Init})\bigl[\Assign^{\Int}(\Init^{\Int})\bigr]$, since it is an RSS condition that guarantees the safe execution of the ``maximum braking'' proper response, by~\cref{eq:validityAndSemantics}. Note that this assertion is over the variables $\xSV,\xPOV,\vSV,\vPOV$: the substitution $\bigl[\Assign^{\Int}(\Init^{\Int})\bigr]$ (namely $[0/\tSV]$) removes occurrences of $\tSV$ in $\gamma(l^{\Int}_{\Init})$; and $\tPOV$ is clearly irrelevant. Summarizing, the assertion $\gamma(l^{\Int}_{\Init})\bigl[\Assign^{\Int}(\Init^{\Int})\bigr]$ tells us when it is safe to execute the ``maximum braking'' MRM.
 \end{auxproof}



\end{auxproof}

\section{ALGORITHM AND IMPLEMENTATION}\label{sec:impl}
\todoil{say we use user annotation (which is trivial)}
Here we discuss our algorithm for finding an RSS condition. It operates on a given proper response that is modeled, together with its driving scenario, modeled as a network $\mathcal{N}$ of hCFGs. The algorithm's outline is in \cref{alg:ourAlgorithm}. 
The algorithm returns an assertion $A$
 that makes the quadruple
 \begin{equation}\label{eq:implOutputQuad}
 \begin{array}{r}
  \bigl\{\,A\,\bigr\}\;
 [\mathcal{G}_{\mathcal{N},\Final}]\;
 \bigl\{\true\bigr\} \;
 \colon\; S \land\bigwedge_{l_{i}\in \Unsafe}\lnot(l=l_{i})
 \end{array}
 \end{equation}
  valid. This $A$ is thought of as an RSS condition for the safety of the proper response.
Our algorithm relies on \cref{thm:validityAndSemantics} and symbolically synthesizes a  valid Hoare annotation $\gamma$, which in turn yields a desired assertion as $\gamma(l_{\Init})\,[\Assign(\Init)]$.
The synthesis works backwards within the hCFG $\mathcal{G}_{\mathcal{N}}$, gradually defining $\gamma(l)$ for earlier $l$. We assume that the hCFG is acyclic.

\begin{auxproof}
 Letting $\mathcal{N}$ be a model of a driving situation (which consists of a driving scenario and \SV{}'s proper response), the output $A$ is thought of as an RSS condition for the safety of the proper response.
\end{auxproof}

\begin{algorithm}[tbp]
\SetAlgorithmName{Procedure}{procedure}{List of Procedures}
\LinesNumbered
\caption{our algorithm}
\label{alg:ourAlgorithm}
\KwIn{
 a network $\mathcal{N}$  of hCFGs, an assertion $S$,  and sets $\Final,\Unsafe$ of locations
}
\KwOut{an assertion $A$ such that \cref{eq:implOutputQuad} is valid}

\emph{hCFG generation}:
generate the hCFG  $\mathcal{G}_{\mathcal{N},\Final}$, following~\cref{def:semNetworkOfHCFG}

\emph{Synthesizing $\gamma$, base case}:
for each  unsafe location $l$, let $\gamma(l):=\false$; for each final location $l$ that is not unsafe, let $\gamma(l):=S$

\emph{Synthesizing $\gamma$, step case}: 
traverse the hCFG  $\mathcal{G}_{\mathcal{N},\Final}$ backwards, finding $C_{l,i}$ in \cref{eq:hoareAssertionValidityQuadruple} and taking their disjunction to define $\gamma(l)$, as in \cref{def:hoareAnnotation}

\Return{$\gamma(l_{\Init})\,[\Assign(\Init)]$}
\end{algorithm}

The most technical step is the step case of the synthesis of $\gamma$, where we search for a precondition $C_{l,i}$ that makes the quadruple~\cref{eq:hoareAssertionValidityQuadruple} valid. To do so, we use the rules of $\dHL$~\cite{HasuoEHDBKPZPYSIKSS23}; note that most Hoare-style rules can be used for calculating weakest preconditions. To deal with
 the $\mathsf{dwhile}$ clauses,  our implementation  uses the rule (\textsc{DWh-Sol}) that exploits closed-form solutions. This makes the symbolic reasoning simpler than with the other rule (\textsc{DWh}) that requires  differential (in)variants. The dynamics in our case study (\cref{sec:intersectionModeling}) have closed-form solutions, too, given by quadratic formulas.

Our implementation is in Haskell, with approximately 1K LoC. It uses Z3 for checking validity of assertions, and Mathematica for solving ODEs.

\section{EXPERIMENTS}\label{sec:expr}
Our implementation of \cref{alg:ourAlgorithm} successfully found a valid Hoare annotation $\gamma$, and hence an RSS condition, for the intersection case study (\S{}\ref{sec:intersectionModeling}). This RSS condition $A$ is an assertion over $\xSV,\vSV,\xPOV,\vPOV$ (and parameters such as $b$) that, if true initially, guarantees a safe execution of the max-brake proper response. 

We used the following  safety specifications, identifying a collision as a joint occupancy of  \CZ{}: $  S^{\Int} := \true$ (no condition on the continuous dynamics); 
\begin{math}
     \Unsafe^{\Int}
 :=
 \bigl\{\,(l^{\mathsf{SVPos}}, 
 \dotsc,
 l^{\mathsf{POVTimer}})\,\big|\,
 l^{\mathsf{SVPos}} = \mathsf{SVInCZ} 
 \land
 l^{\mathsf{POVPos}} = \mathsf{POVInCZ} 
 \,
 \bigr\}
\end{math}.
The automated derivation took a couple of minutes, most of which was spent for validity checking by Z3.  The synthesis of $\gamma$ was assisted by some user-provided assertions, although these assertions are  straightforward (such as $\vSV\ge 0\land \CZStartSV\le \xSV$ for the location $\mathsf{SVInCZ}$ in $\mathcal{G}^{\mathsf{SVPos}}$). Fully automatic synthesis (without assistance) should not be hard; we leave it as future work.



We conducted experiments to evaluate the RSS condition $A$ we obtained for the intersection situation. We posed the following research questions.
\begin{description}
  \item[\bfseries RQ1] Does the RSS condition $A$ indeed guarantee safety ($A$ true $\Rightarrow$ no collision)?  This is theoretically guaranteed by \cref{thm:validityAndSemantics}, but we want to experimentally confirm.
  \item[\bfseries RQ2] Is $A$ not overly conservative? That is, when it is false ($A$ says there is risk), is there a real risk of collision?

\end{description}
Another  question  is about the ease of modeling. We believe that the compositional modeling in~\cref{sec:intersectionModeling} answers it positively.

To address the RQs, we implemented a  simulator of the intersection situation (cf.\ \cref{fig:intersection}), on which we conducted a number of simulations for different parameter values. The parameter space consists of the following.
\begin{itemize}
 \item  Initial values of positions and velocities of \SV{} and \POV{}, chosen from the ranges $\xSV,\xPOV\in\{5, 10, 
\dotsc,
 45\} [\SI{}{\metre}]$ and $\vSV,\vPOV\in\{3, 6, 
\dotsc,
18\} [\SI{}{\metre\per\second}]$.  The numbers  are in lane coordinates;  positions are measured from  \CZ{}'s center. (Note that some initial values are inevitably unsafe.)
 \item There is nondeterminism in \POV{}'s behavior, namely in what it does before it needs to brake (in the location $\mathsf{POVAccelerating}$ of \cref{fig:intersectionHCFG}). We consider \POV{}s that maintain the same acceleration rate $\aPOV$, and $\aPOV$ is chosen from $\{-5, -4,
\dotsc,0, 1, 2\} [\SI{}{\metre\per\second^2}]$.
\end{itemize}

Note that the RSS condition $A$ is formulated in terms of (the initial values of) $\xSV,\xPOV,\vSV,\vPOV$; it does not refer to $\aPOV$ since $A$ needs to guarantee safety for all possible \POV{}'s behavior. Therefore, we refer to a choice of the values of $\xSV,\xPOV,\vSV,\vPOV$ as an \emph{instance}; each \emph{simulation} is run under an additional choice of the value of $\aPOV$.

As for constants, we used
$b=\SI{5}{\metre\per\second^2}, \amax=\SI{2}{\metre\per\second^2}$, and $\rho =\SI{0.3}{\second}$. In a simulation, \SV{} executed the max-brake proper response (cruise for $\rho$ and brake at $b$); \POV{} accelerated at $\aPOV$, braking when it sees \SV{} in \CZ{}.

We ran 
23,328
 simulations, that is, 8 simulations (for different \POV{} behaviors) for each of 2,916 instances.
The  simulation results are summarized in \cref{table:safetystat}, counting 
\begin{itemize}
 \item what we call \emph{RSS complying} instances (i.e.\ in which the initial values of $\xSV,\xPOV,\vSV,\vPOV$ satisfy the RSS condition $A$) vs.\ \emph{RSS non-complying} ones, and
 \item what we call \emph{experimentally safe} instances  (i.e.\  those whose eight simulations under different \POV{} behaviors are all collision-free) vs.\ \emph{experimentally unsafe} ones.
\end{itemize}

 \begin{table}[tbp]
 \caption{Safety, by  RSS rules and by experiments}
 \label{table:safetystat}
    \centering
    \begin{tabular}{lcc}
        \toprule
        {} & RSS complying &RSS non-complying \\
        \midrule
        Experimentally unsafe &  0  &   558  \\
	\cmidrule{1-1}
        Experimentally safe & 2296 &   62 \\
        \bottomrule
    \end{tabular}

\vspace{-1.5em}
\end{table}

\begin{wrapfigure}[11]{r}{0pt}
\centering
  \includegraphics[width=.25\textwidth]{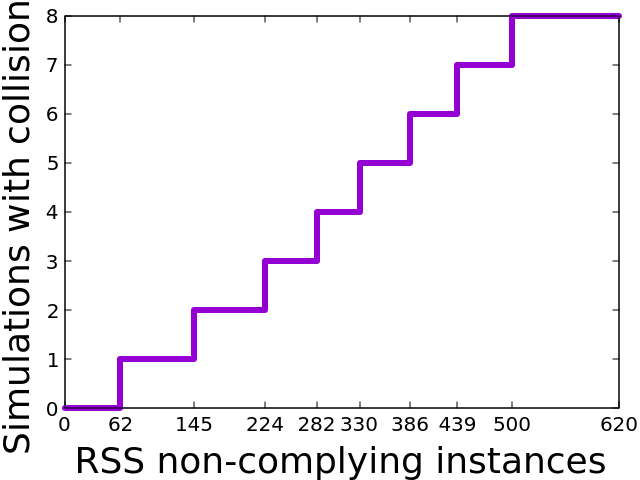}
	
\caption{unsafe simulations per RSS non-complying instance}
  \label{fig:expr}
\end{wrapfigure}

\noindent Furthermore, in \cref{fig:expr}, we present how many simulations exhibit a collision for each RSS non-complying instance. (There is no collision in RSS complying instances.)


The experimental results yield the following answers to the research questions.

On \textbf{RQ1}, we  confirmed experimentally that all RSS complying instances are safe. This is no surprise
given the theoretical guarantee in \cref{thm:validityAndSemantics}.

On \textbf{RQ2}, we  see only a small number of \emph{false positive} instances, i.e.\ those which are RSS non-complying but experimentally safe. According to 
\cref{table:safetystat}, the precision of the RSS condition is 
90\%
and its recall is 100\%. This high precision indicates that the RSS condition $A$, synthesized by our algorithm, is not overly conservative. 

In \cref{fig:expr}, it is observed that the number of unsafe simulations vary greatly---depending on \POV{}'s behavior---and still the RSS condition $A$ successfully spots all unsafe instances. 

Let us consider, for comparison, scenario-based testing methods for detecting  unsafe instances. For such methods, spotting an unsafe instance with fewer unsafe simulations will be a challenge, since they have to find a rare \POV{} behavior that causes a collision. In contrast, \POV{} behaviors are \emph{universally quantified} in our formal verification approach, in the form of the two extreme behaviors in \cref{fig:intersectionHCFG}. Therefore our approach can spot unsafe instances regardless of the rarity of collisions.

We note that an instance that is RSS non-complying with respect to $A$ may still be safe with respect to other proper responses than the max-brake one.
See \cref{subsec:intersectionInformally}. 

\balance
\section{Conclusions}
Building on RSS~\cite{ShalevShwartzSS17RSS} and its formalization~\cite{HasuoEHDBKPZPYSIKSS23}, we presented the first framework for automated formal verification of safety of intersection situations. The modeling formalism of hCFGs is  suited to the multi-agent character of intersection situations; the reasoning formalism of Hoare annotations demonstrates the power of program logic. We believe this is an important step forward towards RSS's coverage of a wide variety of real-world traffic scenarios, and hence towards the public acceptance of automated driving where  its safety is mathematically proved.


\bibliographystyle{IEEEtran}
\bibliography{myrefs_shorten}

\clearpage

\nobalance

\appendix

\subsection{Compatibility Conditions in \cref{def:networkOfHCFG}}
\label{appendix:compatCond}

\begin{itemize}
 \item  For each tuple $(l^{(1)},\dotsc, l^{(k)})$ of locations of the components  (where $l^{(i)}\in L^{(i)}$ for each $i$), their flow dynamics has no conflicts, and moreover it is complete. That is, letting $\Flow(l^{(i)})\equiv(\dot{\mathbf{x}}^{(i)}=\mathbf{f}^{(i)})$, 
1) if the same variable occurs in $\mathbf{x}^{(i)}$ and $\mathbf{x}^{(j)}$ with $i\neq j$, then the polynomials on their right-hand side (in $\mathbf{f}^{(i)}, \mathbf{f}^{(j)}$) coincide, and
 2) the union of $\mathbf{x}^{(1)},\dotsc, \mathbf{x}^{(k)}$ is the set $\Variables$ of variables. 
 \item Similarly,  for each tuple $(e^{(1)},\dotsc, e^{(k)})$ of edges of the component hCFGs, there is no conflict in their assignment commands, meaning that $\Assign(e^{(1)}), \dotsc, \Assign(e^{(k)})$ have the same expressions on the right-hand side if the variable on  the left-hand side is the same. We require the same compatibility for initial assignment commands, too.
\begin{auxproof}
  that is, that there is no conflict among $\Assign(\Init^{(1)}), \dotsc, \Assign(\Init^{(k)})$.
\end{auxproof}

\end{itemize}

\subsection{Semantics of Networks of hCFGs, Formalized}
\label{appendix:semNetHCFG}

\begin{definition}\label{def:semNetworkOfHCFGinDetail}
 The \emph{semantics} of a network  $\mathcal{N}=(\mathcal{G}^{(1)},\dotsc, \mathcal{G}^{(k)})$ of hCFGs is defined to be the semantics (\cref{def:semanticsOfhCFGs}) of its \emph{synchronized product} $\mathcal{G}_{\mathcal{N},\Final}$ with a given set $\Final$ of final locations. The latter is the hCFG defined below. 
\begin{itemize}
 \item Its locations are tuples $(l^{(1)},\dotsc, l^{(k)})$ of locations of components. Here $l^{(i)}\in L^{(i)}$ for each $i$.
 \item Its set $E_{\mathcal{N},\Final}$ of event names the union $E^{(1)}\cup\cdots \cup E^{(k)}$ of those of the components.
 \item Its set $\Variables$ of variables is the one of all components. 
 \item We have an edge 
\begin{math}
 (l^{(1)},\dotsc, l^{(k)})
 \xrightarrow{a}
 (l'^{(1)},\dotsc, l'^{(k)})
\end{math}
       if 1) $l^{(i)}\xrightarrow{a} l'^{(i)}$ is an edge in $\mathcal{G}^{(i)}$ for each $i$ such that $a\in E^{(i)}$, and 2) $l^{(i)}=l'^{(i)}$ otherwise, i.e.\ for every $i$ such that $a\not\in E^{(i)}$. In this case, we say that the component edges $l^{(i)}\xrightarrow{a} l'^{(i)}$ \emph{enable} the edge in $\mathcal{G}_{\mathcal{N},\Final}$.  The intuition is that, if an event name $a$ is shared by multiple components, then \emph{all} these components must synchronize in the event. 
 \item Its flow dynamics at $(l^{(1)},\dotsc, l^{(k)})$ is the union of the flow dynamics of $l^{(1)},\dotsc, l^{(k)}$. By the compatibility condition (\cref{def:networkOfHCFG}), this union is well-defined and covers all variables in $\Variables$, making $\mathcal{G}_{\mathcal{N},\Final}$ a (proper) hCFG.
 \item Its guard for an edge \begin{math}
 (l^{(1)},\dotsc, l^{(k)})
 \xrightarrow{a}
 (l'^{(1)},\dotsc, l'^{(k)})
\end{math}
is the conjunction of those guards associated with the enabling component edges $l^{(i)}\xrightarrow{a} l'^{(i)}$. 
 \item Its assignment command for an edge in $\mathcal{G}_{\mathcal{N},\Final}$ is the concatenation of the assignment commands for the enabling component edges. There are no overriding assignments thanks to the compatibility condition (\cref{def:networkOfHCFG}). The initial assignment commands are defined similarly.
 \item Its initial location is the tuple $(l^{(1)}_{\Init},\dotsc, l^{(k)}_{\Init})$.
 \item Its set $\Final$ of final locations is specified separately. It is a set of locations, that is, $\Final\subseteq L^{(1)}\times\cdots \times L^{(k)}$.

\end{itemize}
\end{definition}

\begin{auxproof}

\clearpage

\section{Modeling Intersection Situations}
\todo[inline]{Discuss ``poor man's nondeterminism'' BTW, how can we justify that one of the two extreme behaviors (full acceleration or full braking) is the worst-case behavior? An idea: would it be easier to justify ``worst-case is the extreme'' if we distinguish cases ``exit CZ before {\POV{}} enters'' and ``avoid entering CZ''?}

\begin{remark}
 In this paper, we just assume that a finite set of edge cases is given.
To do it properly, one possible way is to have two extreme cars in the same model and do ``logical'' worst case analysis, such as
\begin{displaymath}
 \includegraphics[width=.45\textwidth]{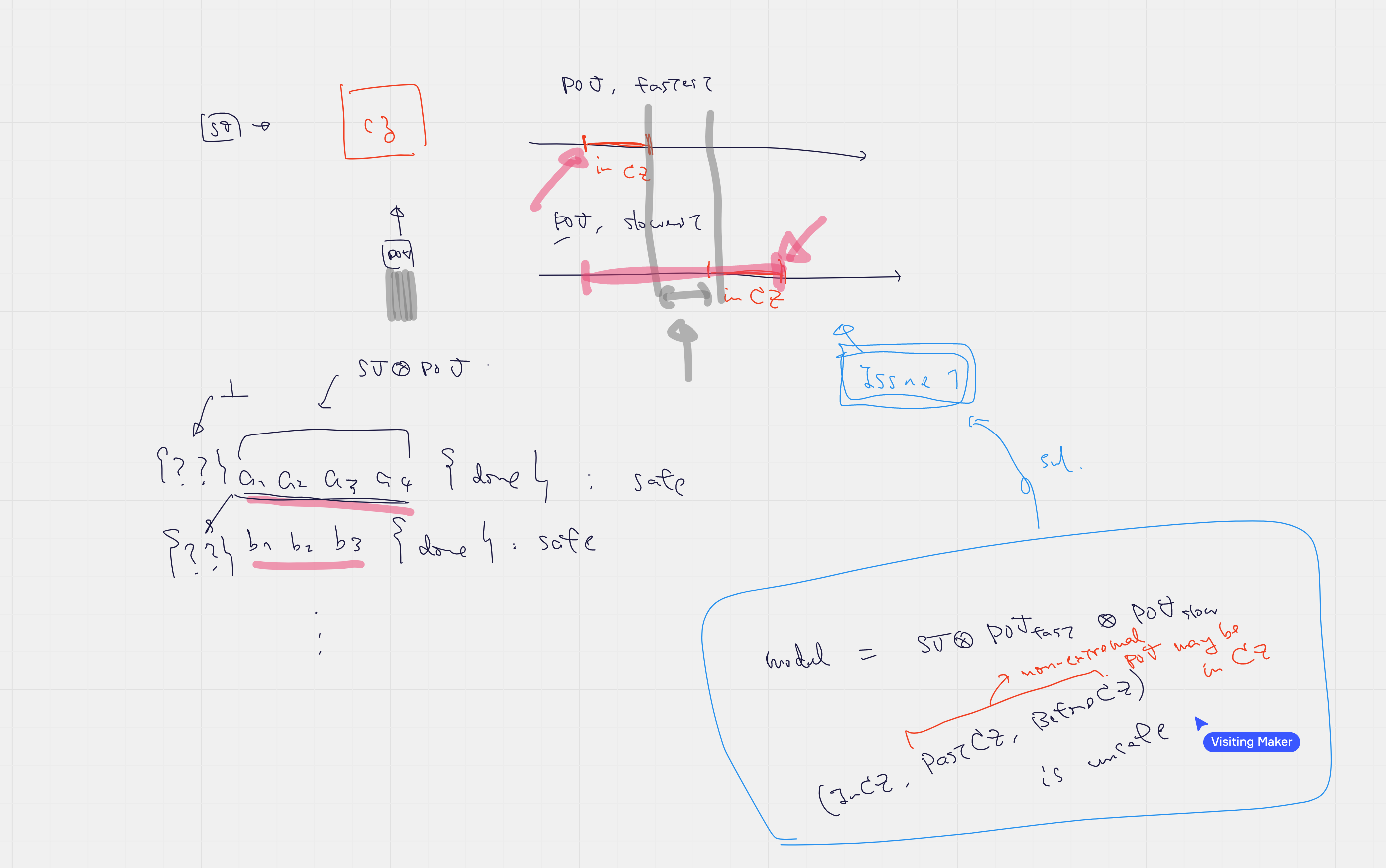}
\end{displaymath}
\end{remark}

\clearpage

\section{a modeling with proper nondeterminism}
\begin{figure*}[tbp]
 \centering
\begin{minipage}[t]{.45\textwidth}
 \begin{subfigure}[b]{\textwidth}
 \centering
 \scalebox{.65}{%
    \begin{tikzpicture}[FAstyle]
          \node[state, initial, text width=1.8cm, align=center, initial text = {$\Init$}] (BeforeCZ)
            {$\mathsf{SVBeforeCZ}$\\$\dot \xSV = \vSV$};
          \node[state, right=of BeforeCZ, text width=1.6cm, align=center] (InCZ)
            {$\mathsf{SVInCZ}$\\$\dot \xSV = \vSV$};
          \node[state, right=of InCZ, text width=1.6cm, align=center] (AfterCZ)
            {$\mathsf{SVAfterCZ}$\\$\dot \xSV = \vSV$};
            \path[->]
              (BeforeCZ)
                edge node[text width=2.4cm]{$\mathsf{SVEnterCZ}$\\$\xSV \geq \CZStartSV $} (InCZ)
              (InCZ)
                edge node[text width=2cm]{$\mathsf{SVExitCZ}$\\$\xSV \geq \CZEndSV$} (AfterCZ);
        \end{tikzpicture}%
        }
  \caption{$\mathcal{G}^{\mathsf{SVPos}}$ for \SV{}'s position}
  \label{fig:hCFGSVPos}
 \end{subfigure}
 \begin{subfigure}[b]{\textwidth}
 \centering
 \scalebox{.65}{
          \begin{tikzpicture}[FAstyle]
          \node[state, initial, text width=1.6cm, align=center, initial text = {$\Init$}] (Any)
          {$\mathsf{SVCruising}$\\$\dot\vSV = 0$};
          \node[state, right=of BeforeCZ, text width=1.6cm, align=center] (Braking)
            {$\mathsf{SVBraking}$\\$\dot\vSV = -b$};
          \node[state, right=of InCZ, text width=1.6cm, align=center] (Stopped)
            {$\mathsf{SVStopped}$\\$\dot\vSV = 0$};
            \path[->]
              (Any)
                edge node[text width=2cm]{$\mathsf{SVStartBraking}$} (Braking)
              (Braking)
                edge node[text width=2cm]{$\mathsf{SVStop}$\\$\vSV \le 0$} (Stopped);
              \end{tikzpicture}
        }
  \caption{$\mathcal{G}^{\mathsf{SVVel}}$ for \SV{}'s velocity}
  \label{fig:hCFGSVVel}
 \end{subfigure}
 \begin{subfigure}[b]{\textwidth}
 \centering
  \scalebox{.65}{
    \begin{tikzpicture}[FAstyle]
          \node[state, initial, text width=2.5cm, align=center, initial text = {$\Init$, $\tSV :=0$}] (Running)
            {$\mathsf{SVTimerRunning}$\\$\dot\tSV  = 1$};
          \node[state, right=6cm of BeforeCZ, text width=2cm, align=center] (Rang)
          {$\mathsf{SVTimerRang}$\\$\dot\tSV = 1$};
            \path[->]
              (Running)
              edge node[text width=2cm]{$\mathsf{SVStartBraking}$\\ $\tSV \ge \rho$} (Rang);
            \end{tikzpicture}
  }
  \caption{$\mathcal{G}^{\mathsf{SVTimer}}$ for \SV{}'s response time}
  \label{fig:hCFGSVTimer}
 \end{subfigure}
\end{minipage}
\hfill
\begin{minipage}[t]{.52\textwidth}
\begin{subfigure}[b]{\textwidth}
 \centering
  \scalebox{.65}{%
    \begin{tikzpicture}[FAstyle]
          \node[state, initial, text width=2.1cm, align=center, initial text = {$\Init$}] (BeforeCZ)
            {$\mathsf{POVBeforeCZ}$\\$\dot\xPOV = \vPOV$};
          \node[state, right=of BeforeCZ, text width=2.1cm, align=center] (InCZ)
            {$\mathsf{POVInCZ}$\\$\dot\xPOV = \vPOV$};
          \node[state, right=of InCZ, text width=2.1cm, align=center] (AfterCZ)
            {$\mathsf{POVAfterCZ}$\\$\dot\xPOV = \vPOV$};
            \path[->]
              (BeforeCZ)
                edge node[text width=2.4cm]{$\mathsf{POVEnterCZ}$\\$\xPOV \ge \CZStartPOV$} (InCZ)
              (InCZ)
                edge node[text width=2.4cm]{$\mathsf{POVExitCZ}$\\$\xPOV \ge \CZEndPOV$} (AfterCZ);
        \end{tikzpicture}%
        }
  \caption{$\mathcal{G}^{\mathsf{POVPos}}$ for \POV{}'s position}
  \label{fig:hCFGPOVPos}
\end{subfigure}
\begin{subfigure}[b]{\textwidth}
 \centering
 \scalebox{.65}{
          \begin{tikzpicture}[FAstyle]
          \node[state, initial, text width=1.8cm, align=center, initial text = {$\Init$}] (Any)
          {$\mathsf{POV}$-\\$\mathsf{Accelerating}$\\$\dot\vPOV = a$};
          \node[state, right=of BeforeCZ, text width=1.8cm, align=center] (Braking)
            {$\mathsf{POVBraking}$\\$\dot\vPOV = -b$};
          \node[state, right=of InCZ, text width=1.9cm, align=center] (Stopped)
            {$\mathsf{POVStopped}$\\$\dot\vPOV = 0$};
            \path[->]
              (Any)
                edge node[text width=2cm]{$\mathsf{POVStart}$-$\mathsf{Braking}$} (Braking)
              (Braking)
                edge node[text width=2cm]{$\mathsf{POVStop}$\\$\vPOV \le 0$} (Stopped);
              \end{tikzpicture}
        } 
  \caption{$\mathcal{G}^{\mathsf{POVVel}}$ for \POV{}'s velocity}
  \label{fig:hCFGPOVVel}
\end{subfigure}
\begin{subfigure}[b]{\textwidth}
 \centering
  \scalebox{.65}{
        \begin{tikzpicture}[FAstyle]
         \node[state, initial, text width=1.8cm, align=center, initial text = {$\Init$}] (NotResponding)
            {$\mathsf{POVNot}$-\\$\mathsf{Responding}$\\$\dot\tPOV = 1$};
          \node[state, right=of NotResponding, text width=1.8cm, align=center] (Running)
            {$\mathsf{POVTimer}$-\\$\mathsf{Running}$\\$\dot\tPOV = 1$};
          \node[state, right=of Running, text width=1.8cm, align=center] (Rang)
          {$\mathsf{POVTimer}$-\\$\mathsf{Rang}$\\$\dot\tPOV = 1$};
          \path[->]
            (NotResponding) edge node[text width=2cm]{$\mathsf{SVEnterCZ}$\\ $\tPOV := 0$} (Running)
              (Running)
              edge node[text width=2cm]{$\mathsf{POVStart}$-\\$\mathsf{Braking}$\\ $\tPOV \ge \rho$} (Rang);
            \end{tikzpicture}
  }
  \caption{$\mathcal{G}^{\mathsf{POVTimer}}$ for \POV{}'s response time}
  \label{fig:hCFGPOVTimer}
\end{subfigure}
\end{minipage}
\caption{(original) the network $\mathcal{N}^{\Int}$ of hCFGs for the intersection driving situation, consisting of six open hCFGs; the set $ \Final^{\Int}$ of final locations are defined in~\cref{eq:intersectionFinalLocation} (to-do: add ``POVMaxStoppedMinBraking.'' This is clearly vacuous but good for systematic modeling)}
\label{fig:intersectionHCFGInterval}
\end{figure*}

\clearpage
\section{Labeled Hybrid Control Flow Graphs (lhCFG)}
\todo{Let's make the theoretical exposition minimal! We are short of space, and they wouldn't understand formalization anyway}

\subsection{Fist draft}

We want to model situations with several agents which interact. The clearest way
to model this is therefore using a form of synchronised composition.

\begin{definition}[event labeling]
  An \emph{event labeling} is a join semilattice $(E, \bot, \leq, \lor)$ equipped with
  a monotone function $s \colon E \to E$, called the \emph{synchroniser}.
\end{definition}

\begin{definition}[discrete machine]
  A \emph{discrete machine} is an action $a \colon E \times S \rightharpoonup S$ of an event
  labeling $E$ on a set $S$ of states, such that
  \[
    \forall e. \forall x. \quad a(e,x) = a(s(e),x)
  \]
  For a state $x \in S$ and an event $e \in E$, we say that $e$ is \emph{enabled} if
  $a(e, x)$ is defined.
\end{definition}
\todoil{Merge into definition of lhCFG}

\begin{definition}
  The \emph{synchronised product} is obtained by combining two machines $M_{1}$
  and $M_{2}$ with two synchronisation functions
  \[
    s_{1,2} \colon E_{1} \to E_{2}, \quad s_{2,1} \colon E_{2} \to E_{1}.
  \]
  The join semilattice of events is $E_{1} \times E_{2}$ (the product join semilattice) with synchronisation:
  \[
    (x, y) \mapsto (x \lor s_{2,1}(y), y \lor s_{1,2}(x)).
  \]
  The action $a \colon (E_{1} \times E_{2}) \times (S_{1} \times S_{2}) \to S_{1} \times S_{2}$ is
  defined by
  \[
    a ((e_{1}, e_{2}), (x_{1}, x_{2})) = (f_{1}(e_{1}, x_{1}), f_{2}(e_{2}, x_{2})).
  \]
  When the synchronisers $s_{1,2}$ and $s_{2,1}$ are understood, the
  synchronised product is denoted $M_{1} \otimes M_{2}$.
\end{definition}

Rermarks:
\begin{itemize}
  \item When forming a producy of more than two machines, the event labelings
        and synchronisers form a \emph{propagotor network} \todo{add ref}.
  \item To ensure synchronisation is tractable, we require the event labelings
        to satisfy the ascending chain condition.
\end{itemize}

\begin{definition}
  A \emph{lhCFG} consists in:
  \begin{itemize}
    \item A discrete machine $M = (E, S, a)$. Elements of $S$ are called
          \emph{locations}. For all locations $l \in S$, for all events $e_{1}$
          and $e_{2}$, if $e_{1}$ and $e_{2}$ are both enabled, then
          $e_{1} \lor e_{2}$ must be enabled.
    \item \emph{Continuous variables:} A set of variables $V_{C}$.
    \item \emph{Discrete variables:} A set of variables $V_{D}$.
    \item \emph{Dynamics:} For each location $l \in S$ for which at least one
          event $e$ is enabled, a dynamics for each continuous variable, that
          is, a term $\dyn(x)$ for each variable $x \in V_{C}$. The dynamics can
          also be \emph{unspecified} for some variables, when defining an
          \emph{open} lhCFG.
    \item \emph{Variants:} A function
    \[
      \vari \colon E \to \Open(V_{C}, V_{D})
    \]
          where $\Open(V_{C}, V_{D})$ is the lattice of open assertions on the
          variables $V_{C}, V_{D}$. The function $\vari$ should be contravariant join-preserving, that is:
          \[
            \vari(e_{1} \lor e_{2}) = \vari(e_{1}) \land \vari(e_{2}).
          \]
    \item \emph{Discrete update}: For each $e \in E$, and for each discrete
          variable $x \in V_{D}$, an update term $u_{x}$.
  \end{itemize}
\end{definition}

\todo[inline]{We don't have to do semantics of lhCFGs with free variables. This is because we do not do compositional reasoning. Parallel composition is there solely for describing complex models in a modular and readable manner. Compositional reasoning is future work.}

Rough description of semantics:
\begin{itemize}
  \item Define a valuation.
  \item At a location evolution is called \emph{flow}. Continuous variables
        evolve according to the dynamics. If a variable's dynamics is
        unspecified, then it evolves non-deterministically. Discrete variables
        remain constant during flow.
  \item Flow happens while all the variants of all the enabled events remain
        true. Because all variants are open, there is always an earliest time
        when one stops being true. Assume $T \subset E$ is the full set of variants
        that have been falsified at that time, that is, $\vari(e)$ is false for
        $e \in T$. By assumption, $\biglor_{e \in T} e$ is enabled, and its
        variant-assertion must be false since:
        $\vari(\biglor_{e \in T} e) = \bigland_{e \in T} \vari(e)$. So the
        transition $\biglor_{e \in T} e$ is taken, the location changes according
        to the action of $\biglor_{e \in T} e$, and the associated updates to the
        discrete variables.
    \end{itemize}

\begin{definition}[Parallel composition of open lhCFGs]
  Open lhCFGs may be parallel composed as follows. Let $\mathcal{M}_{1}$ and $\mathcal{M}_{2}$ be
  open lhCFGs. We assume that:
  \begin{itemize}
    \item The sets of discrete variables of the machines are disjoint. (Discrete
          variables are always ``private''.)
    \item The sets of continuous variables of the machines may overlap, but:
    \begin{itemize}
      \item A continuous variable's dynamics is either defined at all locations,
            or undefined at all locations, for each machine.
      \item If a shared continuous variable's dynamics is defined in both
            machines, then the definitions must be identical.
    \end{itemize}
    \item A pair of synchronisers is given between the event labelings:
          \[
            s_{1,2} \colon E_{1} \to E_{2}, \quad s_{2,1} E_{2} \to E_{1}.
          \]
    \end{itemize}
    Then the product machine $\mathcal{M}_{1} \otimes \mathcal{M}_{2}$
    \begin{itemize}
      \item Has underlying discrete machine $M_{1} \otimes M_{2}$ (the synchronised
            product of the underlying discrete machines).
      \item Continuous variables ${V_{C}}_{1} \cup {V_{C}}_{2}$.
      \item Discrete variables ${V_{D}}_{1} \cup {V_{D}}_{2}$.
      \item Dynamics for the variable $x \in {V_{C}}_{1} \cup {V_{C}}_{2}$ for the
            pair of locations $(l_{1}, l_{2})$ is any of the dynamics specified
            in $\mathcal{M}_{1}$ or $\mathcal{M}_{2}$, or undefined if the dynamics is undefined in
            both machines. The dynamics cannot be defined in both machines in
            case of a shared variable, so this is well-defined.
      \item The variant assertion $\vari((e_{1}, e_{2}))$ associated to the pair
            of events $(e_{1}, e_{2})$ is:
            \[
              \vari((e_{1}, e_{2})) = \vari(e_{1}) \land \vari(e_{2}).
            \]
      \item Since discrete variables are disjoint (private), the discrete
            updates do not conflict.
    \end{itemize}
\end{definition}

\begin{definition}[Hoare structure on lhCFG]
  A \emph{Hoare structure} for a lhCFG $\mathcal{M}$ is:
  \begin{itemize}
    \item \emph{Location invariants:} For each location $l$, an assertion
          $I_{l}$.
    \item For each location $l$, and each event $e$ that is enabled at $l$,
          \begin{itemize}
            \item \emph{Precondition:} an assertion $P_{l,e}$.
            \item \emph{Transition:} For each continuous variable $x \in V_{C}$,
                  an update function, that is, a function
                  \[
                    h_{x} : \mathrm{Term}^{V_{C} \cup V_{D}} \to \mathrm{Term}.
                  \]
          \end{itemize}
  \end{itemize}
  The Hoare structure is \emph{valid} if, for any execution starting at location
  $l$ and valuation $\rho$ of variables, if $\rho \vDash P_{l,e}$, then execution will
  eventually transition via the event $e$, the continuous variables will all be
  updated according to the update functions $h_{x}$, and $\rho' \vDash I_{l}$ will hold
  for all intermediate valuations $\rho'$. \todoil{Make this precise once the
    semantics machinery is all defined.}
\end{definition}

\begin{remark}
  Note that we can't easily translate this directly to a Hoare quadruple,
  because the post-condition of a Hoare quadruple doesn't have access to the
  ``variables as they were at the start of execution''. It's still possible but
  we'd have to duplicate all the variables and create a bunch of extraneous
  copying assignments (it's very ugly).
\end{remark}

\begin{theorem}
  Let $\mathcal{M}$ be a lhCFG with Hoare structure $(I_{l}, P_{l,e}, h_{x})$. Let
  $T = (s_{0} \xrightarrow{e_{0}} s_{1} \xrightarrow{e_{1}} \to \cdots \xrightarrow{e_{n-1}} s_{n})$
  be a discrete trace. Then the precondition
  \[
    \biglor_{0 \leq i < n} (h_{i-1} \circ \cdots \circ h_{0})^{-1}(P_{i})
  \]
  ensures that the system will execute the trace $T$, where
  \[
    P_{i} := P_{s_{i}, e_{i}}, \quad h_{i} := h_{s_i, e_{i}}.
  \]
\end{theorem}

\todoil{What follows is Ichiro's initial draft}

\subsection{Formal Definition}
\begin{definition}[lhCFG]
\todo{James, can you write this?}

(hCFG plus labels. For richer synchronization structures, we can use machineries that are similar to Winskel's \emph{event structures}~\cite{...} and Jackson's \emph{concepts}~\cite{daniel_jackson_essence_of_software}. This is future work)
 
\end{definition}

\subsection{Parallel Composition of lhCFG}

\subsection{Translation to dFHL, and Semantics}

\begin{itemize}
 \item An lhCFG $\mathcal{M}$ may have the dynamics of some variables undefined. In such case, the semantics of $\mathcal{M}$ is defined for each signal $\sigma'$ that valuates the dynamics-undefined variables
 \item We might even put lhCFGs in contexts, such as $x_{1},\dotsc, x_{m}\vdash \mathcal{M}$, in which case we can say that an lhCFG is \emph{closed} if $\vdash \mathcal{M}$
 \item dFHL translation and semantics of open lhCFGs is future work.
\begin{itemize}
 \item We'll have to do so when we do assume-guarantee reasoning.
 \item To translate open lhCFGs, we'd need to extend dFHL as well, so that their programs accommodate external signals. 
\end{itemize}

\end{itemize}

\section{Preconditions in lhCFGs}
Focus on closed lhCFGs...

\begin{definition}
 Let $\mathcal{M}=(L, \dotsc)$ be a closed lhCFG, and $l\in L$ be its location. 
Let 
\begin{displaymath}
 c(l)\;= \;
\bigl[\mathrm{dwhile} (\bigwedge_{i}f_{i}(x) >0 ) \{\dot{x}=g(x)\}
\bigr]\end{displaymath}
be the dwhile command associated with $l$; 
 let $e_{i}=(l\to l_{i})_{i}$ be the family of outgoing transition from $l$; and 
let $e_{i}$ be associated with the variant $\lnot(f_{i}>0)$ and the update $x:= g_{i}(x)$. 

For each $j$, 
the \emph{dFHL quadruple for taking the transition $e_{j}$} is 
\begin{equation}\label{eq:quadrupleForTakingTheTransition}
 \bigl\{ A \bigr\}
 \,\mathrm{dwhile} (f_{j}(x) >0 ) \{\dot{x}=g(x)\}\,
 \bigl\{ \true \bigr\}
 \;:\;
 \bigwedge_{i\neq j} f_{i}(x) >0 
\end{equation}
where $A$ is an arbitrary assertion. 
\end{definition}

\begin{remark}
The safety condition  $ \bigwedge_{i\neq j} f_{i}(x) >0 $ in the Hoare quadruple  in~(\ref{eq:quadrupleForTakingTheTransition}) is not strictly needed: since the program is a single dwhile loop, requiring $\lnot(f_{j}(x)>0)$ in the postcondition takes the same effect, assuming that any two conditions from $(f_{i}(x)>0)_{i}$ are not simultaneously violated. (Note that the loop terminates as soon as one of  $(f_{i}(x)>0)_{i}$ is violated.) 

Nevertheless, we prefer to use
the safety condition  $ \bigwedge_{i\neq j} f_{i}(x) >0 $ since it is easier to derive using dFHL rules. 
\end{remark}

\begin{definition}
 Let $\mathcal{M}=(L, \dotsc)$ be a closed lhCFG. A \emph{precondition assignment} for $\mathcal{M}$ is a function $\pi$ that assigns an assertion $\pi(l)$ to each location $l\in L$ that is subject to the following condition.
\begin{itemize}
 \item If $l$ is an error location, then $\pi(l)\cong \bot$. 
 \item (If $l$ is a successful final location, then we do not have to assume anything, since $\pi(l)$ just has to be below $\true$)
 \item For each location $l$, ... (we have to extend (\ref{eq:quadrupleForTakingTheTransition}), using the assignment for the successor location as a postcondition, and require that the Hoare quadruple is valid).

\end{itemize}
\end{definition}

\begin{theorem}
 Correctness
\end{theorem}

\section{Case Study}

\section{Experiments}\label{sec:expr}

\subsection{Research Questions}

We have several research questions:
\begin{itemize}
  \item[(RQ1)] Is the precondition calculated by the workflow safe? That is,
        when it is true, the safety condition is never violated.
  \item[(RQ2)] Is the precondition calculated by the workflow too restrictive?
        That is, is it false in instances that seem perfectly safe?
  \item[(RQ3)] How much work is it to specify a scenario?
\end{itemize}

To assess these we measured the following:
\begin{itemize}
  \item For instances where the precondition was true, but the safety condition was violated, we checked if it was still violated after increasing the accuracy of the simulation.
  \item For instances where the precondition was false, but the safety condition
        was not violated, we checked that a different behaviour of the POV and the SV
                (with the same starting conditions) does indeed lead to a
                simulation where the safety condition is violated.

\end{itemize}

\subsection{Experiment settings and results}

We ran simulations under different instances of the intersection scenario. The scenario instances were generated by the following parameter values for both the SV and the POV: we found them generate relevant scenario instances. Here the starting positions (distance to the center of the collision zone) are in meters (m) and velocities are in m/s :\\
$v_{init} \in \{3, 6, 9, 12, 15, 18\}$\\
$y_{init}\in \{5, 10, 15, 20, 25, 30, 35, 40, 45\}$\\
For constants, we used the following values :
$a_{max} = 2$ $m/s^{2}$, $b_{max} = 5$ $m/s^{2}$ and $\rho = 0.3$s \\
\indent	While simulating, for each instance we ran a simulation for each POV and SV behaviour/initial acceleration ($m/s^{2}$) within these values such that : \\
$a_{init}\in  \{-5, -4, -3, -2, -1, 0, 1, 2\}$\\
In total we conducted experiment on 2916 instances throughout 186,624 simulations.\\
\indent The statistics of the simulation results are given in \cref{table:one}. In the \emph{Unsafe} column we count the number of simulations where  both vehicle end up inside the collision zone at the same time and the number of instances where at least one simulation is unsafe. In the \emph{Precond violation} column we count the number of instances where the computed precondition states that this instance could end up with both vehicle in the collision zone, the calculated precondition is the same for each simulation of these instances. In the \emph{Precond inaccurate} column we count the number of simulations where both vehicles end up in the collision zone but the precondition states this couldn't have happened and the number of instances where at least one of those simulation happen. In the \emph{Partially false positive} column we count the number simulations that ended up with a safe outcome even though the precondition stated there could be a case where both vehicle end up in the collision zone and the number of instances where one of these simulations happened. Finally,  in the \emph{False positive} column we count the number of instances where every simulation ended up with a safe outcome even though the precondition stated there could be a case where both vehicle end up in the collision zone.

\begin{table*}
\begin{tabular}{ |c||c|c|c|c|c| } 
\hline
- & Unsafe & Precond violation & Precond innacurate & Partially false positive & False positive \\
\hline
Instances	& 590 & 620 & 0 & 511 & 30 \\ 
\hline
Simulations & 17,262 & 39,680 & 0 & 22,018 & 1,920 \\ 
\hline
\end{tabular}
\label{table:one}
\end{table*}

\subsection{Discussion}

\indent With those experimental results, we can now adress two of the research questions.
\begin{enumerate}
\item[(RQ1)] \emph{Is the precondition calculated by the workflow safe?} \\
		  	Through our experiments, simulations where the computed precondition was true never led to the safety condition being violated. Therefor the precondition is indeed safe for the initial parameters tested.
\item[(RQ2)] \emph{Is the precondition calculated by the workflow too restrictive?}\\
			For most instances, when the safety condition wasn't violated for every simulation of the instance, the precondition was true. However some instances didn't violate the safety precondition no matter the simulation and their precondition was computed as false (approximately 1 \%). For these instances, increasing the precision of the simulations didn't made one of them unsafe. We attribute those rare cases to implementation details {\color{red} here it is meant that $A$ is stronger than necessary}: All of those cases corresponded to situtations where both the SV and POV would stop. 
			When it comes to simulations, a lot more of them had the precondition false but the safety condition not being violated. In most cases (94 \%) it was because another simulation with similar starting parameters but a different POV behaviour led to both vehicles being in the collision zone at the same time. Some could argue that if this other POV behaviour was an extreme one (close to braking or accelerating for the maximum amount) and unique among this instance then it would mean that realisticly, our precondtion computation is too restrictive but that wasn't the case. In instances where only one of POV behaviour led to an unsafe outcome, this behaviour wasn't extreme in most cases therefore the precondition being false was still justified. {\color{red} here we mean that these singular unsafe simulations are not extreme but realistic}
\item[(RQ3)] \emph{How much work is it to specify a scenario?}
\end{enumerate}

\clearpage

\section{Oct 2022: Ichiro Understanding James' Work}
2022/10/28
\begin{itemize}
 \item Covered until the definition of type \texttt{Scenario}
 \item So far what has been done: to introduce an (abstract and finite) transition system, specified by first-order logic, much like in Event-B
 \item The abstract domains (e.g.\ \texttt{Time} as a three-element set) are defined so that they are the coarsest abstractions needed so far. It may be needed---in the future when we deal with more complicated scenarios---that these abstract domains be refined, but we can do so when it is needed. 
\end{itemize}

next: \texttt{globStep}

\section{Hybrid Control Flow Graphs}

\section{Generalities}\label{sec:generalities}
\todoil{Let's present our modeling formalism as 1) a variation of hybrid automata, but 2) specialized for the use of structuring dFHL reasoning. We should come up with a good name.}

\begin{definition}[state machine]
  A state machine $M$ consists of:
  \begin{itemize}
    \item A set $S_{M}$ of \emph{modes},
    \item A set $E_{M}$ of \emph{events},
    \item A function $f_{M} \colon E_{M} \times S_{M} \to S_{M} \cup \{\bot\}$,
    \item Subsets $\init, \ter \subseteq S_{M}$.
  \end{itemize}
  A \emph{transition} of $M$ is a triple $(s, e, s') \in S_{M} \times E_{M} \times S_{M}$
  such that $s' = f_{M}(e, s)$. When the machine $M$ is understood, we write
  $s \xrightarrow{e} s'$ instead of $s' = f_{M}(e, s)$.
\end{definition}

\begin{definition}[trace]
  A \emph{trace} of machine $M$ (of length $n$) is given by a sequence of $n+1$
  states $(s_{i})_{0 \leq i \leq n}$ and a sequence of events $(e_{i})_{0 \leq i < n}$ of
  $n$ events such that for all $0 \leq i < n$,
  $s_{i} \xrightarrow{e_{i}} s_{i+1}$ is a transition. The trace is
  \emph{complete} if $s_{0} \in \init$ and $s_{n} \in \ter$. We'll write traces as
  \[
    s_{0} \xrightarrow{e_{0}} s_{1} \xrightarrow{e_{1}} s_{2} \to \cdots \xrightarrow{e_{n-1}} s_{n}.
  \]
\end{definition}

\begin{definition}[hybrid state machine]
  A \emph{hybrid state machine} $H$ is given by:
  \begin{itemize}
    \item A state machine $M$,
    \item A vector space $V$ over $\bbR$,
    \item A set $D$,
    \item For each mode $m \in S_{M}$ that isn't terminal ($m \notin \ter$), and each
          $d \in D$, a vector field $\dyn(m,d)$ of $V$,
    \item For each $(e, d) \in E \times D$, an open subset $O_{e,d} \subseteq V$, a subset
          $G_{e,d} \subseteq V$ and a function
          $\jump(e, d) \colon V \to D$.
  \end{itemize}
\end{definition}
Remarks:
\begin{itemize}
  \item The machine $M$ is most often finite, and represents the ``control
        flow'' of the hybrid state machine.
  \item The vector space $V$ represents the continuous part of the state-space
        of the hybrid state machine, while the set $D$ represents the discrete
        part of the state space.
  \item The vector space $V$ could be replaced with any differentiable manifold.
  \item The open subset $O_{e,d}$, the ``event condition'', often takes the form
        $\vari(e, d) > 0$, for some continuous function
        $\vari(e, d) \colon V \to \bbR$ (the ``event variant''), or a conjunction
        of such conditions.
  \item The set $G_{e,d}$ is called the ``guard'' of the event $e$.
  \item The function $\jump(e,d)$ only needs to be defined over
        $\overline{O_{e,d}} \cap G_{e,d}$, where $\overline{O_{e,d}}$ is the border
        of $O_{e,d}$.
  \item The dynamics $\dyn(m, d)$ could be replaced by a differential inclusion.
  \item We should require, for each mode $m$ and $d \in D$,
  \[
    \bigcup_{m \xrightarrow{e} m'} G_{e,d} = V.
  \]
        But this alone does not guarantee the machine reaches a terminal state
        (e.g.~spending infinite time in some mode).
  \item TODO: require something else to ensure the next edge transition is
        always unique.
\end{itemize}

\begin{example}[obstacle stop]
  \label{ex:obstacle}
  Consider a vehicle at position $x$, which cruises for a reaction time of $\rho$
  seconds before starting to brake. The vehicle should come to a stop before an
  obstacle at position $o$. The following diagram represents a hybrid state
  machine for this situation.

  \begin{tikzpicture}[FAstyle]
    \node[state, initial, text width=1.2cm, align=center] (Cruising)
      {Cruising\\ $\dot{t} = 1$\\ $ \dot{x} = v$\\ $\dot{v} = 0$};
    \node[state, right=of Cruising, text width=1.2cm, align=center] (Braking)
      {Braking\\ $\dot{t} = 1$\\ $\dot{x} = v$\\ $\dot{v} = -1$};
    \node[state, accepting, below=of Cruising, text width=1.2cm, align=center, bottom color=red!20] (Crashed)
      {Crashed};
    \node[state, accepting, right=of Crashed, text width=1.2cm, align=center] (Stopped)
      {Stopped};
      \path[->]
        (Cruising)
          edge node[text width=2cm]{Brake:\\$t < \rho \mid x \neq o$} (Braking)
          edge node[swap]{Crash: $x < o$} (Crashed)
        (Braking)
          edge node[text width=2cm]{Stop:\\$v > 0 \mid x \neq o$} (Stopped)
          edge node[text width=2cm]{Crash:\\$x < o$} (Crashed)
        (Crashed);
      \end{tikzpicture}

      Each node of this diagram represents a mode. The ``Crashed'' mode is the only unsafe mode (in red). The nodes with a double border are terminal. Each edge is labelled with the corresponding event, condition and guard. E.g. `Brake: $t
      < \rho \mid x \neq o$' corresponds to the event $\text{Brake} \in
      E$, with condition $x < \rho$ and guard $x \neq o$. The continuous state is $V =
      \bbR^{3}$, with basis elements $t$, $x$ and $v$ representing time, position and speed respectively. In this case $D
      = \mathbf{1}$ so the jump functions are trivial (and not shown).
\end{example}

\begin{definition}[Operational semantics]
  For a hybrid machine $H$ with state-machine $M$, we define a timed transition
  system on $S_{M} \times D \times V$:
  \[
    \to \subseteq (S_{M} \times D \times V) \times \Rplus \times (S_{M} \times D \times V).
  \]
  We'll write $x \to_{t} y$ instead of $(x, t, y) \in \to$, to indicate that the
  system has evolved from state $x \in S_{M} \times D \times V$ to state $y \in S_{M} \times D \times V$
  during which $t \in \Rplus$ time has elapsed.

  Transitions fall into two categories:
  \begin{itemize}
    \item Continuous evolution: $(m, d, v) \to_{t_{0}} (m, d, v')$ when there
          exists a unique $\Psi \colon \Rplus \to V$ such that:
    \begin{itemize}
      \item $\Psi(0) = v$,
      \item $\Psi(t_{0}) = v'$,
      \item $\frac{d \Psi(t)}{dt} = \dyn(m,d)$,
      \item For all edges $m \xrightarrow{e} m'$,
      \[
        \forall t \in \Rplus. \ t < t_{0} \Rightarrow \Psi(t) \in O_{e,d}
      \]
    \end{itemize}
    \item A discrete jump: $(m, d, v) \to_{0} (m', d', v)$ along some edge
          $m \xrightarrow{e} m'$ when:
    \begin{itemize}
      \item $v \notin O_{e, d}$,
      \item $v \in G_{e, d}$,
      \item $d' = \jump(e, d)(v)$.
    \end{itemize}
  \end{itemize}
  The transition $\to^{*}$ is defined as the smallest transition such that:
  \begin{itemize}
    \item If $x \to_{t} y$ then $x \to_{t}^{*} y$,
    \item If $x \to_{t_{1}}^{*} y$ and $y \to_{t_{2}}^{*} z$, then $x \to_{t_{1} + t_{2}}^{*} z$.
  \end{itemize}

\end{definition}

\begin{definition}[Hoare structure]
  Given a hybrid machine $H$, a \emph{Hoare structure (?)} is,
  \begin{itemize}
    \item For each mode $m$, an \emph{invariant} $I_{m} \subseteq D \times V$,
    \item for each transition $m \xrightarrow{e} m'$, a \emph{precondition}
          $P_{m,e} \subseteq D \times V$ and a function $h_{m,e} \colon P_{m,e} \to V$ such
          that, when in mode $m$, for any $(d, x) \in P_{s,e}$, the hybrid machine
          is guaranteed to transition to mode $m'$ along edge $e$, and the
          resulting continuous state will be $h_{m,e}(d,x)$. Furthermore, all
          intermediate states will satisfy the invariant $I_{s}$.

          Formally: If $(d, x) \in P_{m,e}$, then there exists some
          $t_{0} \in \Rplus$ and $d \in D$ such that:
          \[
            (m, d, x) \to_{t_{0}} (m, d, h_{m,e}(d,x)) \to_{0} (m', d', h_{m,e}(d,x))
          \]
          and, for all $t < t_{0}$, $d' \in D$ and $x'\in V$, if
          $(m, d, x) \to_{t}^{*} (m', d', x')$ then $(d', x') \in I_{m}$.
  \end{itemize}
   \todoil{find a better name?}
\end{definition}

\begin{theorem}
  Given a hybrid machine $H$ with Hoare structure $(I_{m}, P_{m,e}, h_{m,e})$,
  for any trace
  $T := (s_{0} \xrightarrow{e_{0}} s_{1} \xrightarrow{e_{1}} s_{2} \to \cdots \xrightarrow{e_{n-1}} s_{n})$
  where $s_{n}$ is terminal, the precondition
  \[
    \bigcap_{0 \leq i < n} (h_{i-1} \circ \cdots \circ h_{0})^{-1}(P_{i}).
  \]
  ensures that the system will execute the trace $T$, where
  \[
    P_{i} := P_{s_{i}, e_{i}} \ \text{and} \ h_{i} := h_{s_{e}, e_{i}}.
  \]
\end{theorem}

\begin{lemma}
 (Some sanity-check lemma, such as mutual translation between our formalism and dFHL)
\end{lemma}

If the two formalisms (ours and dFHL) are mutually translatable, why we introduce ours?
\begin{itemize}
 \item Parallel composition is easier
 \item Discrete interaction patterns at intersections are complex. Our state-based formalism allows systematic \emph{rule-based  generation} of them. For example, exhaustive enumeration is easier in our formalism than in dFHL. 
\end{itemize}

\clearpage

\section{Workflow}\label{sec:workflow}

We apply the whole workflow to the simple example.

\begin{enumerate}
  \item \emph{Define hybrid machine.} Done above.
  \item \emph{Define global safety property.} $x < o$.
  \item \emph{Compute safe traces.} In this case there is only one safe
        trace:
  \[
    \text{Cruising} \xrightarrow{\text{Brake}} \text{Braking} \xrightarrow{\text{Stop}} \text{Stopped}.
  \]
  \item \emph{Worst case analysis.} Doesn't apply to this example.
  \item \emph{Define/compute/check the Hoare structure.} The invariants are:
  \begin{align*}
    \text{Cruising} &: x < o \land v \geq 0 \land t \leq \rho\\
    \text{Braking} &: x < o \land v \geq 0 \land t \geq \rho\\
    \text{Stopped} &: x < o \land v = 0 \land t \geq \rho
  \end{align*}
  The Hoare transitions are:
  \begin{itemize}
    \item $\text{Cruising} \xrightarrow{\text{Brake}} \text{Braking}$:
    \begin{itemize}
      \item precondition: $t = 0 \land v \geq 0 \land x + \rho v < o$
      \item mapping: $(t, x, v) \mapsto (\rho, x + \rho v, v)$
      \item check: \texttt{delay-stop-1.dfhl}
    \end{itemize}
    \item $\text{Braking} \xrightarrow{\text{Stop}} \text{Stopped}$:
    \begin{itemize}
      \item precondition: $v \geq 0 \land v^{2}/2 < o - x$
      \item mapping: $(t, x, v) \mapsto (t + v, x + v^{2}/2, 0)$
      \item check: \texttt{delay-stop-2.dfhl}
    \end{itemize}
  \end{itemize}
  \item \emph{Check global safety.} Trivial:
  \begin{align*}
    \text{Cruising} &: x < o \land v \geq 0 \land t \leq \rho \Rightarrow x < o\\
    \text{Braking} &: x < o \land v \geq 0 \land t \geq \rho \Rightarrow x < o\\
    \text{Stopped} &: x < o \land v = 0 \land t \geq \rho \Rightarrow x < o
  \end{align*}
  \item \emph{Compute trace preconditions.}
  Using the formula:
  \begin{align*}
    & P_{\text{Cruising} \xrightarrow{\text{Brake}} \text{Braking}} \land h_{\text{Cruising} \xrightarrow{\text{Brake}} \text{Braking}}^{-1}(P_{\text{Braking} \xrightarrow{\text{Stop}} \text{Stopped}})\\
    & \Leftrightarrow \ \left( t = 0 \land v \geq 0 \land x + \rho v < o \right) \land \left( v \geq 0 \land v^{2}/2 < o - (x + \rho v) \right)\\
    & \Leftrightarrow \ t = 0 \land v \geq 0 \land v^{2}/2 < o - x - \rho v
  \end{align*}
\end{enumerate}

\clearpage

\subsection{The Interaction State Machine (Tentative)}
A (discrete) \emph{machine} $M$ is given by:
\begin{itemize}
  \item A finite set $S_{M}$ of states,
  \item A finite set $E_{M}$ of events,
  \item A transition function $f_{M} \colon E_{M} \times S_{M} \to S_{M} \cup \{\bot\}$,
  \item Subsets $\init, \safe, \ter \subseteq S$.
\end{itemize}

A transition is a triple $(s, e, s') \in S_{M} \times E_{M} \times S_{M}$ such that
$s' = f_{M}(e, s)$. When the machine $M$ is understood, they are also written
$s \xrightarrow{e} s'$.

A \emph{trace} of machine $M$ (of length $n$) is given by some starting state
$s_{0} \in \init$ and a sequence $(e_{i} \mid 0 \leq i < n)$ of $n$ events such that the
$s_{i+1} := f_{M}(e_{i}, s_{i})$ are all defined (i.e.~$\neq \bot$), and
$s_{n} \in \ter$. Traces can also be written
\[
  s_{0} \xrightarrow{e_{0}} s_{1} \xrightarrow{e_{1}} s_{2} \to \cdots \xrightarrow{e_{n-1}} s_{n}.
\]
Such a trace is \emph{safe} if $\forall i, \ s_{i} \in \safe$.

Given some conditions (e.g.~no cycles), all traces are finite, and there are
only finitely many of them.

A \emph{continuous semantics} for $M$ is:
\begin{itemize}
  \item An affine space $\bbR^{m}$ for some $m \geq 0$.
  \item For each $s \in S$, a condition $C_{s} \subseteq \bbR^{m}$, the ``invariant'', and
        a dynamics $d_{s} \colon \bbR^{m} \to T \bbR^{m}$.
  \item For each event $e$, an open subset $O_{e} \subseteq \bbR^{m}$, whose boundary is
        denoted $B_{e} \subseteq \bbR^{m}$,
  \item For each transition $s \xrightarrow{e} s'$, a \emph{guard}
        $G_{s \xrightarrow{e} s'} \subseteq M$
\end{itemize}

The continuous semantics is \emph{valid} if: For each transition
$s \xrightarrow{e} s'$, for each $x \in G_{s \xrightarrow{e} s'}$, there exists a
single solution $\Phi_{s \xrightarrow{e} s'}$ to the dynamics $d_{s}$ with initial condition $\Phi(0,x)=x$,
and there exists some $\tend \in \Rplus$ such that:
\begin{itemize}
\item $0 \leq t < \tend \ \Rightarrow \ \Phi_{s \xrightarrow{e} s'}(t, x) \in C_{s} \cap O_{e}$.
\item $\Phi_{s \xrightarrow{e} s'}(\tend, x) \in B_{s} \cap C_{s'}$
\end{itemize}
The resulting function
\begin{align*}
  G_{s \xrightarrow{e} s'} &\to \bbR^{m}\\
  x &\mapsto \Phi_{s \xrightarrow{e} s'}(x,\tend)
\end{align*}
is denoted $f_{s \xrightarrow{e} s'}$.

Given a trace
$s_{0} \xrightarrow{e_{0}} s_{1} \xrightarrow{e_{1}} s_{2} \to \cdots \xrightarrow{e_{n-1}} s_{n}$,
and a valid continuous semantics, a \emph{continuous trace} is given by a
function $\Phi \colon [0,\tend] \to \bbR^{m}$ such that there exists a sequence
$(t_{i} \in [0, \tend] \mid 0 \leq i \leq n)$ of times, such that
\[
t_{0} = 0, \ t_{n} = \tend
\]
\[
\forall i. \ \forall t. \ t_{i} \leq t \leq t_{i+1} \ \Rightarrow \ \Phi(t) = \Phi_{s_{i} \xrightarrow{e_{i}} s_{i+1}}(t - t_{i}, \Phi(t_{i})).
\]
Note that the existence of this continuous trace depends on the existence of the
solutions $\Phi_{s_{i} \xrightarrow{e_{i}} s_{i+1}}$.

Lemma: If the continuous trace over a discrete trace exists, then it is uniquely
determined by the starting point $\Phi(0)$. We then refer to it as the continuous
trace \emph{starting} at $\Phi(0)$.

Assume given a trace
\[
  T := (s_{0} \xrightarrow{e_{0}} s_{1} \xrightarrow{e_{1}} s_{2} \to \cdots \xrightarrow{e_{n-1}} s_{n}).
\]
We wish to determine $T$'s \emph{global precondition}: the set of points
$x_{0} \in \bbR^{n}$ such that the continuous trace $\Phi$ over $T$ starting at
$x_{0}$ exists, and satisfies the all the invariants along its whole domain.

For the trace $T$, we will use the notation:
\begin{align*}
  f_{i} &:= f_{(s_{i}, e_{i}, s_{i+1})}\\
  G_{i} &:= G_{(s_{i}, e_{i}, s_{i+1})}
\end{align*}


Lemma: For traces of length $n$, the global precondition is:
\[
\bigcap_{0 \leq i < n} (f_{i-1} \circ \cdots \circ f_{0})^{-1}(G_{i}).
\]

Proof by induction:

Base case is trivial. Assume that the result is true for all traces of length
$n$. We'll prove the statement for traces of length $n+1$. Let
$T := (s_{0} \xrightarrow{e_{0}} s_{1} \xrightarrow{e_{1}} s_{2} \to \cdots \xrightarrow{e_{n}} s_{n+1})$
be a trace of length $n+1$.
Since
\[
  T := (s_{1} \xrightarrow{e_{1}} s_{2} \to \cdots \xrightarrow{e_{n}} s_{n+1})
\]
is a trace of length $n$, we know that
\[
  \bigcap_{1 \leq i < n+1} (f_{i-1} \circ \cdots \circ f_{1})^{-1}(G_{i})
\]
is its global precondition. Therefore the global precondition of $T$ is:
\begin{align*}
  &G_{0} \cap f_{0}^{-1} \left( \bigcap_{1 \leq i < n+1} (f_{i-1} \circ \cdots \circ f_{1})^{-1}(G_{i}) \right)\\
  &=  G_{0} \cap \bigcap_{1 \leq i < n+1} f_{0}^{-1} \left( (f_{i} \circ \cdots \circ f_{1})^{-1}(G_{i}) \right)\\
  &=  G_{0} \cap \bigcap_{1 \leq i < n+1} (f_{i} \circ \cdots \circ f_{1} \circ f_{0})^{-1}(G_{i})\\
  &=  \bigcap_{0 \leq i < n+1} (f_{i} \circ \cdots \circ f_{0})^{-1}(G_{i}).
\end{align*}

\section{Workflow}

A number of steps are necesary to ultimately generate the proof obligations for one scenario :

\begin{enumerate}
	\item First the hybrid state machine $H = (M,V,D,S_{F},T)$ needs to be modeled for the scenario, with:
	\begin{itemize}
		\item It's state machine $M = (S,E,f,S_{Init},S_{Terminal})$
		\item It's vector space $V$ over a field $F$
		\item It's subset $D$
		\item $S_{F}$ the set composed of vector fields  of $\dyn(s,d)$ for each non terminal state s and element $ d \in D $
		\item $T$ the set of guards $G_{e,d}$, conditions  $O_{e,d}$ and function $\jump(e, d)$ for each $(e, d) \in E \times D$
	\end{itemize}
	\item Then specify a predicate $p_{Safety} \colon D^{n} \to \mathbb{B}$ . It is the global safety property we ultimately want to ensure at all time of the scenario. 
	\item Define the Hoare structure for H. This is done by :
	\begin{itemize}
		\item For each mode, specify its invariant in the shape of a predicate on  $D \times V \to \mathbb{B}$ describing the specific aspects of that mode.
		\item Build the subset $S_{Safe} \in S$ of modes which satisfy the safety predicate
		\item Specify the precondition $P_{s \xrightarrow{e} s'} \subseteq D \times V \to \mathbb{B}$ for each transition between safe modes $s \xrightarrow{e} s'$
	\end{itemize}
	\item For each safe mode $s \in S_{Safe}$ , prove that the global safety porperty is satisfied (done by construction) 
	\item Enumerate all possible traces being both safe and complete.
	\item Worst Case analysis For each trace having the the worst starting conditions 	  
	\item Find the global precondition for each safe traces being \[ \bigcap_{0 \leq i < n} (f_{i-1} \circ \cdots \circ f_{0})^{-1}(G_{i}). \] for a trace of lenght $n$.
    
\end{enumerate}

\clearpage
\begin{example}[Traffic light]
  \label{ex:Traffic light}
  Consider a vehicle at position $x$ approaching a traffic light at position $o$, which can display 3 colors $Light \in \{Green, Orange, Red\}$ .
  Starting green, it turns orange in $t_{O}$ amount of time and finally red in $t_{R}$ (we call $t_{OR}$ the time it takes to switch from orange to red). 
  The vehicle cruises as long as the light is green. 
  When it turns orange, the vehicle chooses to brake or accelerate depending on a certain position $c$ being ahead or behind of it. 
  When the light turns red, it brakes and try to stop before it.
  It should either pass the traffic light before it turns red or come to a stop before its position. 
  The following diagram represents a hybrid state machine for this situation.

	\begin{tikzpicture}[FAstyle]
	    \node[state, initial, text width=1.2cm, align=center] (Cruising)
	      {Cruising\\ $ \dot{t} = 1$\\  $ \dot{x} = v$\\ $\dot{v} = 0$};
	    \node[state, above=of Cruising, text width=1.2cm, align=center] (Braking)
	      {Braking\\ $ \dot{t} = 1$\\ $\dot{x} = v$\\ $\dot{v} = -1$}; 
		\node[state,below=of Cruising, text width=1.2cm, align=center] (Accelerating)
	      {Accelerating\\ $ \dot{t} = 1$\\ $\dot{x} = v$\\ $\dot{v} =1$};
	    \node[state,left=of Cruising, text width=1.2cm, align=center] (Red)
	      {Red\\ $\dot{t} = 1$\\ $\dot{x} = v$\\ $\dot{v} = -1$};
	    \node[state, accepting,left=of Red, text width=1.2cm, align=center, bottom color=red!20] (Through Red)
	      {Through Red};
	    \node[state, accepting,right=of Cruising, text width=1.2cm, align=center] (Passed)
	      {Passed};
	    \node[state, accepting, left=of Braking, text width=1.2cm, align=center] (Stopped)
	      {Stopped};
	      \path[->]
	        (Cruising)
	          edge node[text width=2cm, align=right]{Brake:\\$t_{c\to b}$} (Braking)
	          edge node[text width=2cm]{Accel:\\$t_{c\to a}$} (Accelerating)
		  edge node[text width=2cm]{Pass:\\$x \leq o$} (Passed)
	        (Braking)
	          edge node[text width=2cm]{Stop:\\$v > 0 \mid x \neq o$} (Stopped)
	          edge node[text width=2cm]{Light:\\$t_{b\to r}$} (Red)
		  edge node[text width=2cm]{Pass:\\$x \leq o$} (Passed)
	        (Accelerating)
	          edge node[text width=2cm,, align=right]{Pass:\\$x \leq o$} (Passed)
	          edge node[text width=2cm, align=right]{Light:\\$Light \neq Red $} (Red)
		(Red)
		  edge node[text width=2cm]{Pass through:\\$x \leq o$} (Through Red)
		  edge node[text width=2cm, align=right]{Stop:\\$v > 0 \mid x \neq o$} (Stopped);
	      \end{tikzpicture}

  Where : \\
	$t_{c\to a}$ : $Light \neq Orange \mid x > c$ \\
	$t_{c\to b}$ : $Light \neq Orange \mid x \leq c$ \\
	$t_{b\to r}$ : $Light \neq Red \mid v > 0$  \\

  The global safety invariant in this situation is : \\
  $Light \neq Red \lor x <o$

	Let us specify the Hoare structure for this Hybrid state machine and situation. The modes invariants are:
	  \begin{align*}
	    \text{Cruising} &: x < o \land v \geq 0 \land t \leq t_{O}\\
	    \text{Braking} &: x < o \land v \geq 0 \land t_{O} \geq t \geq t_{R}\\
	    \text{Accelerating} &: x < o \land v \geq 0 \land t_{O} \geq t \geq t_{R}\\
	    \text{Red} &: x < o \land v \geq 0 \land t \geq  t_{R}\\
	    \text{Stopped} &: x < o \land v = 0 \\
	    \text{Passed} &: x \geq o \land v \geq 0 \\
	  \end{align*}
	 Therefore every mode is Safe except for $Through Red$

	 \vspace{165mm}
	  The Hoare transitions are:
	  \begin{itemize}
	    \item $\text{Cruising} \xrightarrow{\text{Brake}} \text{Braking}$:
	      \\precondition: $t = 0 \land v \geq 0 \land x < c \land x +  t_{O} v < o$
	     \item $\text{Cruising} \xrightarrow{\text{Accel}} \text{Accelerating}$:
	      \\precondition:  $t = 0 \land v \geq 0 \land x \geq c \land x +  t_{O} v < o$
	    \item $\text{Cruising} \xrightarrow{\text{Pass}} \text{Passed}$:
	      \\precondition:  $t = 0 \land v \geq 0 \land x +  t_{O} v \geq o$
	    
	    \item $\text{Braking} \xrightarrow{\text{Stop}} \text{Stopped}$:
	     \\precondition:  $v \geq 0 \land x + \frac{v^{2}}{2}  < o$
	    \item $\text{Braking} \xrightarrow{\text{Light}} \text{Red}$:
	     \\precondition: $v \geq 0 \land x +  t_{OR} (v - \frac{ t_{OR}}{2}) < o$
	    \item $\text{Braking} \xrightarrow{\text{Pass}} \text{Passed}$:
	     \\precondition:  $v \geq 0 \land x +  t_{OR} (v - \frac{ t_{OR}}{2}) \geq o$

	    \item $\text{Accelerating} \xrightarrow{\text{Light}} \text{Red}$:
	     \\precondition: $v \geq 0 \land x +  t_{OR} (v + \frac{ t_{OR}}{2}) < o$
	    \item $\text{Accelerating} \xrightarrow{\text{Pass}} \text{Passed}$:
	     \\precondition: $v \geq 0 \land x +  t_{OR} (v + \frac{ t_{OR}}{2}) \geq o$

	    \item $\text{Red} \xrightarrow{\text{Stop}} \text{Stopped}$:
	     \\precondition: $v \geq 0 \land x + \frac{v^{2}}{2}  < o$ \\
	     
	    \end{itemize}	

	Safe traces are :
	 \begin{enumerate}
		\item  \[ \text{Cruising} \xrightarrow{\text{Pass}} \text{Passed} . \]
		\item  \[ \text{Cruising} \xrightarrow{\text{Accel}} \text{Accelerating} \xrightarrow{\text{Pass}} \text{Passed}. \]
		\item  \[ \text{Cruising} \xrightarrow{\text{Accel}} \text{Accelerating} \xrightarrow{\text{Light}} \text{Red} \xrightarrow{\text{Stop}} \text{Stopped}. \]
		\item  \[ \text{Cruising} \xrightarrow{\text{Brake}} \text{Braking} \xrightarrow{\text{Stop}} \text{Stopped}. \]
		\item  \[ \text{Cruising} \xrightarrow{\text{Brake}} \text{Braking} \xrightarrow{\text{Pass}} \text{Passed}. \]
		\item  \[ \text{Cruising} \xrightarrow{\text{Brake}} \text{Braking} \xrightarrow{\text{Light}} \text{Red} \xrightarrow{\text{Stop}} \text{Stopped}. \]
	\end{enumerate}

	For which the global preconditions are, using the formula:
	  \begin{enumerate}
	     \item $P_{\text{Cruising} \xrightarrow{\text{Pass}} \text{Passed}}$\\
	     \item $P_{\text{Cruising} \xrightarrow{\text{Accel}} \text{Accelerating}} \land h_{\text{Cruising} \xrightarrow{\text{Accel}} \text{Accelerating}}^{-1}(P_{\text{Accelerating} \xrightarrow{\text{Pass}} \text{Passed}})$\\
	     \item $P_{\text{Cruising} \xrightarrow{\text{Accel}} \text{Accelerating}} \land h_{\text{Cruising} \xrightarrow{\text{Accel}} \text{Accelerating}}^{-1}(P_{\text{Accelerating} \xrightarrow{\text{Light}} \text{Red}} \land  h_{\text{Accelerating} \xrightarrow{\text{Light}} \text{Red}}^{-1}(P_{\text{Red} \xrightarrow{\text{Stop}} \text{Stopped}}))$\\
	     \item $P_{\text{Cruising} \xrightarrow{\text{Brake}} \text{Braking}} \land h_{\text{Cruising} \xrightarrow{\text{Brake}} \text{Braking}}^{-1}(P_{\text{Braking} \xrightarrow{\text{Stop}} \text{Stopped}})$\\
	     \item $P_{\text{Cruising} \xrightarrow{\text{Brake}} \text{Braking}} \land h_{\text{Cruising} \xrightarrow{\text{Brake}} \text{Braking}}^{-1}(P_{\text{Braking} \xrightarrow{\text{Pass}} \text{Passed}})$\\
	     \item $P_{\text{Cruising} \xrightarrow{\text{Brake}} \text{Braking}} \land h_{\text{Cruising} \xrightarrow{\text{Brake}} \text{Braking}}^{-1}(P_{\text{Braking} \xrightarrow{\text{Light}} \text{Red}} \land  h_{\text{Braking} \xrightarrow{\text{Light}} \text{Red}}^{-1}(P_{\text{Red} \xrightarrow{\text{Stop}} \text{Stopped}}))$\\
	  \end{enumerate}
	\end{example}

\section{Case study}

\subsection{Scenario}

In this section we use the above to generate RSS rules for an intersection. The
situation is shown in figure \ref{fig:intersection}.
\begin{figure}[tbp]
 \centering
  \includegraphics[width=\columnwidth]{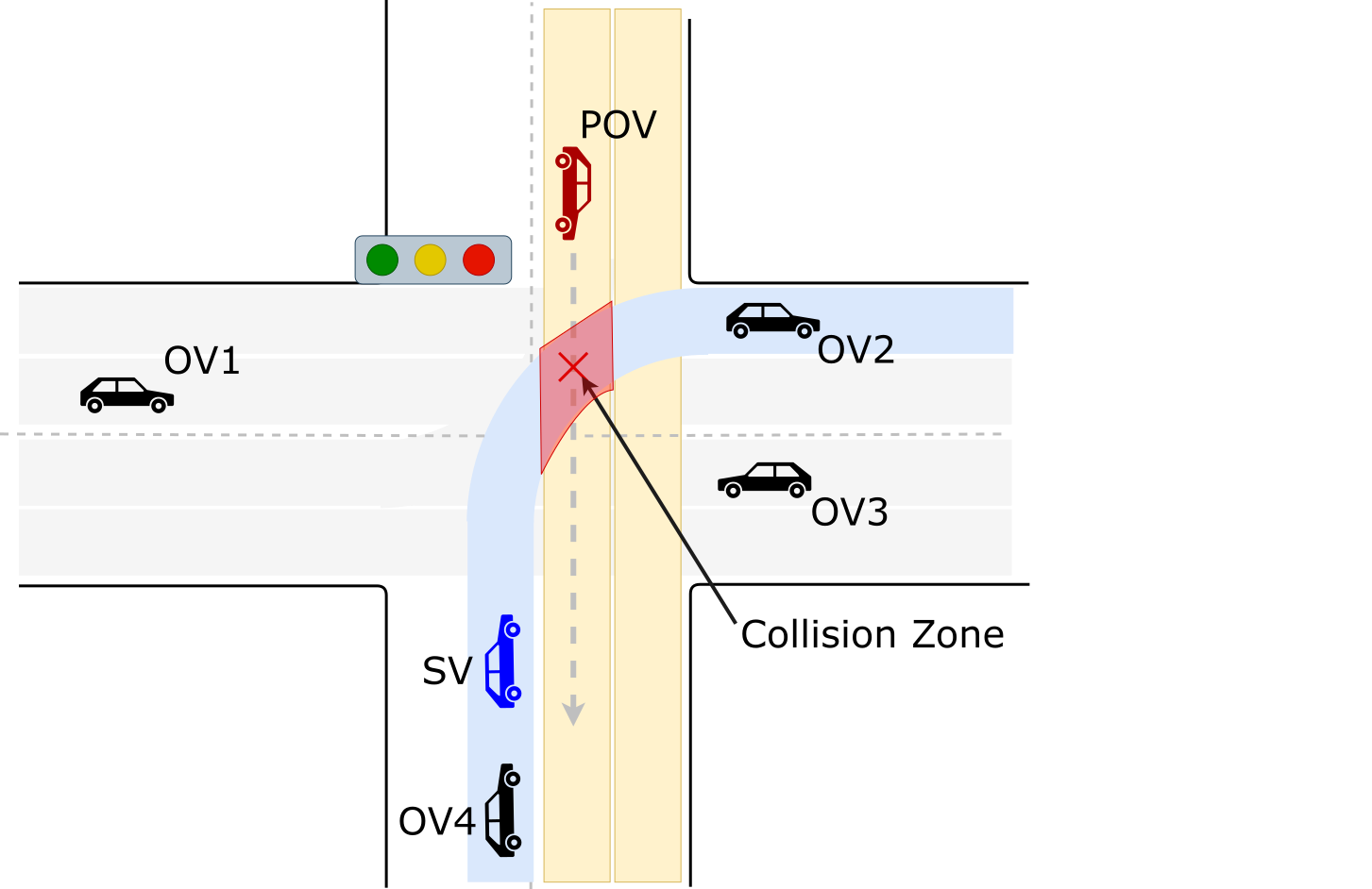}
 \caption{Intersection}
\end{figure}
We only consider {\POV{}} opposite {\SV} which has priority crossing the
intersection. {\SV} is attempting to make a right turn.

We consider two levels\todoil{Insert a reference to hierarchical RSS?} of RSS
rules for this scenario:
\begin{itemize}
  \item Level 0: Collision avoidance. At this level we assume only basic
        constraints on {\POV{}}'s behaviour and emergency measures. The only
        objective is to avoid collisions. The intersection of {\SV}'s and {\POV{}}'s
        lanes is called the ``collision zone''. A situation is deemed unsafe if
        (any part of) both vehicles simultaneously occupy the collision zone.
  \item Level 1: Courteous driving. At this level more is assumed about {\POV{}}'s
        behaviour, and {\SV} is only meant to enter the intersection (beyond the
        stop line) if it should be able to drive smoothly through the whole
        intersection without having to stop midway.
\end{itemize}

In this paper we will only look at the derivation of the level 0 ruleset. Both
{\SV} and {\POV{}} are assumed to have bounded speed and acceleration. The proper
response for the {\SV} is always to brake at the maximum braking rate, after a
reaction time of $\rho$ seconds. The {\POV{}}'s behaviour is constrained too: if the {\SV}
enters the collision zone ahead of it, then after a reaction time of at most $\rho$
seconds, it must begin braking till it has come to a stop.

The design of these reactions should be understood in the context of a wider
ruleset for the whole intersection scenario. The {\SV} should be able to perform an
emergency stop at any time, for example to account for a pedestrian suddenly
advancing towards the road. Thus the {\SV} can never guarantee to exit the
collision zone, since it may be forced to perform an emergency stop while it is
crossing it. For the {\SV} to be able to enter the zone at all then, it must be
able to assume the {\POV{}} will react to it being in the collision zone, even
though the {\POV{}} has priority.

\subsection{Modelling}

To model the scenario we define a network of six hCFGs: three model the {\SV}, and
three model the POV. Together they define an hCFG $\mathcal{I}$.

\subsubsection{Modelling {\SV}}

{\SV} is modelled by three hCFGs:
\begin{itemize}
  \item {\SV}'s position is modelled by the hCFG
        \[
        \mathsf{SVPos} := (L, \Sigma, E, \Flow, \Guard, \Assign)
        \]
        where:
  \begin{align*}
    L &= \{\mathsf{SVBeforeCZ}, \mathsf{SVInCZ}, \mathsf{SVAfterCZ}\},\\
    \Sigma &= \{ \mathsf{SVEnterCZ}, \mathsf{SVExitCZ}\},\\
    E &= \left\{
            \begin{array}{l}
              (\mathsf{SVBeforeCZ, \mathsf{SVEnterCZ}, \mathsf{SVInCZ}}),\\ (\mathsf{SVInCZ}, \mathsf{SVExitCZ}, \mathsf{SVAfterCZ})
            \end{array}
         \right\},\\
    \Flow &= (l \mapsto (\dot{x}_{\text{SV}} = v_{\text{SV}}))
    \\
    \Guard &=
    \left(
    \begin{array}{l}
      \mathsf{SVEnterCZ} \mapsto x_{\text{SV}} \geq c_{\text{start}, \text{SV}}\\
      \mathsf{SVExitCZ} \mapsto x_{\text{SV}} \geq c_{\text{end}, \text{SV}}
    \end{array}
    \right)
    \\
    \Assign &= (e \mapsto ()).
  \end{align*}
        Diagrammatically, the hCFG $\mathsf{SVPos}$ is represented in figure
        \ref{fig:sv-position-hcfg}.
        \begin{figure}[tbp]
  \centering
\resizebox{\columnwidth}{!}{%
    \begin{tikzpicture}[FAstyle]
          \node[state, initial, text width=1.8cm, align=center] (BeforeCZ)
            {$\mathsf{SVBeforeCZ}$\\$\dot x_{\text{SV}} = v_{\text{SV}}$};
          \node[state, right=of BeforeCZ, text width=1.6cm, align=center] (InCZ)
            {$\mathsf{SVInCZ}$\\$\dot x_{\text{SV}} = v_{\text{SV}}$};
          \node[state, right=of InCZ, text width=1.6cm, align=center] (AfterCZ)
            {$\mathsf{SVAfterCZ}$\\$\dot x_{\text{SV}} = v_{\text{SV}}$};
            \path[->]
              (BeforeCZ)
                edge node[text width=2cm]{$\mathsf{SVEnterCZ}$:\\$x_{\text{SV}} \geq c_{\text{start},\text{SV}}$} (InCZ)
              (InCZ)
                edge node[text width=2cm]{$\mathsf{SVExitCZ}$:\\$x_{\text{SV}} \geq c_{\text{end},\text{SV}}$} (AfterCZ);
        \end{tikzpicture}%
        }
  \caption{hCFG for SV's position.}
  \label{fig:sv-position-hcfg}
\end{figure}
        This hCFG preoccupies itself only with SV's position with respect to the collision zone: $c_{\text{start},\text{SV}}$ and $c_{\text{end},\text{SV}}$ mark the start and end of the collision zone respectively.\todoil{James: Technically SV and POV have 2 distinct collision zones. Each is calculated as the set of positions of the centerpoint of the vehicle for any other part of the vehicle to occupy the actual collision zone. This seemed liked to much detail to include an explanation of.} $\mathsf{SVPos}$ is defined over two variables $x_{\text{SV}}$, representing the SV's position (along it's lane), and $v_{\text{SV}}$ representing the SV's speed. Note that the dynamics of $v_{\text{SV}}$ is not defined in this hCFG (which is therefore open).

  \item \todoil{James: update these hCFGs to be consistent with the first one.} There are two possibilities for SV's speed, either it brakes at maximum before the response time $\rho$, or it is accelerating at maximum before the response time $\rho$. Maximum braking case:\newline \resizebox{\columnwidth}{!}{
          \begin{tikzpicture}[FAstyle]
          \node[state, initial, text width=1.6cm, align=center] (Any)
          {SV Max Brake\\$v_{\text{SV}}' = b$};
          \node[state, right=of BeforeCZ, text width=1.6cm, align=center] (Braking)
            {SV Braking\\$v_{\text{SV}}' = b$};
          \node[state, right=of InCZ, text width=1.6cm, align=center] (Stopped)
            {SV Stopped\\$v_{\text{SV}}' = 0$};
            \path[->]
              (Any)
                edge node[text width=2cm]{SV Start Braking} (Braking)
              (Braking)
                edge node[text width=2cm]{Stop:\\$v_{\text{SV}} > 0$} (Stopped);
              \end{tikzpicture}
        }
        Maximum acceleration case:\newline
        \resizebox{\columnwidth}{!}{
          \begin{tikzpicture}[FAstyle]
          \node[state, initial, text width=1.6cm, align=center] (Any)
          {SV Max Accel\\$v_{\text{SV}}' = a$};
          \node[state, right=of BeforeCZ, text width=1.6cm, align=center] (Braking)
            {SV Braking\\$v_{\text{SV}}' = b$};
          \node[state, right=of InCZ, text width=1.6cm, align=center] (Stopped)
            {SV Stopped\\$v_{\text{SV}}' = 0$};
            \path[->]
              (Any)
                edge node[text width=2cm]{SV Start Braking} (Braking)
              (Braking)
                edge node[text width=2cm]{Stop:\\$v_{\text{SV}} > 0$} (Stopped);
              \end{tikzpicture}
            }
        The whole workflow forks at this point, and at the end the preconditions must be combined (conjunction) to obtain the true precondition which accounts for both cases.
  \item Response time:\newline
  \resizebox{\columnwidth}{!}{
    \begin{tikzpicture}[FAstyle]
          \node[state, initial, text width=1.6cm, align=center] (Running)
            {SV Timer Running\\$t' = 1$};
          \node[state, right=of BeforeCZ, text width=1.6cm, align=center] (Rang)
          {SV Timer Rang\\$t' = 1$};
            \path[->]
              (Running)
              edge node[text width=2cm]{SV Start Braking: $t < \rho$} (Rang);
            \end{tikzpicture}
  }
    SV responds after a reaction time of $\rho$ seconds.
  \end{itemize}
  Note that there is only a single synchronised event in the above three machines: `SV Start Braking'. The second and third machine could have been combined, but we choose to present them as separate but synchronised machines to show the similarities in modelling with the POV, which is treated next.

\subsubsection{Modelling POV}

POV is also modelled by the parallel composition of 3 machines:
\begin{itemize}
  \item Position:\newline
  \resizebox{\columnwidth}{!}{%
    \begin{tikzpicture}[FAstyle]
          \node[state, initial, text width=1.9cm, align=center] (BeforeCZ)
            {POV Before CZ\\$x_{\text{POV}}' = v_{\text{POV}}$};
          \node[state, right=of BeforeCZ, text width=1.9cm, align=center] (InCZ)
            {POV In CZ\\$x_{\text{POV}}' = v_{\text{POV}}$};
          \node[state, right=of InCZ, text width=1.9cm, align=center] (AfterCZ)
            {POV After CZ\\$x_{\text{POV}}' = v_{\text{POV}}$};
            \path[->]
              (BeforeCZ)
                edge node[text width=2cm]{POV Enter:\\$x_{\text{POV}} < c_{\text{start},\text{POV}}$} (InCZ)
              (InCZ)
                edge node[text width=2cm]{POV Exit:\\$x_{\text{POV}} < c_{\text{end},\text{POV}}$} (AfterCZ);
        \end{tikzpicture}%
        }
        \item There are two possibilities for POV's speed, either it brakes at maximum or accelerates at maximum before responding to SV entering the collision zone.
        Maximum braking case:\newline \resizebox{\columnwidth}{!}{
          \begin{tikzpicture}[FAstyle]
          \node[state, initial, text width=1.6cm, align=center] (Any)
          {POV Max Brake\\$v_{\text{POV}}' = b$};
          \node[state, right=of BeforeCZ, text width=1.6cm, align=center] (Braking)
            {POV Braking\\$v_{\text{POV}}' = b$};
          \node[state, right=of InCZ, text width=1.6cm, align=center] (Stopped)
            {POV Stopped\\$v_{\text{POV}}' = 0$};
            \path[->]
              (Any)
                edge node[text width=2cm]{POV Start Braking} (Braking)
              (Braking)
                edge node[text width=2cm]{Stop:\\$v_{\text{POV}} > 0$} (Stopped);
              \end{tikzpicture}
        }
        Maximum acceleration case:\newline
        \resizebox{\columnwidth}{!}{
          \begin{tikzpicture}[FAstyle]
          \node[state, initial, text width=1.6cm, align=center] (Any)
          {POV Max Accel\\$v_{\text{POV}}' = a$};
          \node[state, right=of BeforeCZ, text width=1.6cm, align=center] (Braking)
            {POV Braking\\$v_{\text{POV}}' = b$};
          \node[state, right=of InCZ, text width=1.6cm, align=center] (Stopped)
            {POV Stopped\\$v_{\text{POV}}' = 0$};
            \path[->]
              (Any)
                edge node[text width=2cm]{POV Start Braking} (Braking)
              (Braking)
                edge node[text width=2cm]{POV Stop:\\$v_{\text{POV}} > 0$} (Stopped);
              \end{tikzpicture}
            }
        Again the workflow forks at this point.
  \item Response time:\newline
  \resizebox{\columnwidth}{!}{
        \begin{tikzpicture}[FAstyle]
         \node[state, initial, text width=1.6cm, align=center] (NotResponding)
            {POV Not Responding\\$t' = 1$};
          \node[state, right=of NotResponding, text width=1.6cm, align=center] (Running)
            {POV Timer Running\\$t' = 1$};
          \node[state, right=of Running, text width=1.6cm, align=center] (Rang)
          {POV Timer Rang\\$t' = 1$};
          \path[->]
            (NotResponding) edge node[text width=2cm]{SV Enter CZ: $r := t$} (Running)
              (Running)
              edge node[text width=2cm]{POV Start Braking: $t < r + \rho$} (Rang);
            \end{tikzpicture}
  }

        POV responds to SV entering the collision zone after a reaction time of
        $\rho$ seconds. The variable $r$ is a discrete variable.
  \end{itemize}
  There is a synchronisation between these three machines that model the POV:
  the shared event `POV Start Braking'.

  The two machines which model the SV and the POV are also parallel composed,
  and they share the event `SV Enter CZ', which therefore has the effect of
  starting POV's response timer.

  Recall that the positions of the hCFG $\mathcal{I}$ defined by this network of hCFGs are tuples
  \[
    l = (l_{\mathsf{SVPos}}, l_{\mathsf{SVSpeed}}, l_{\mathsf{SVTimer}}, l_{\mathsf{POVPos}}, l_{\mathsf{POVSpeed}}, l_{\mathsf{POVTimer}})
  \]
  where $l_{i}$ is a location of hCFG $i$. A position $l$ of $\mathcal{I}$ is defined to
  be final if one of the following is true:
  \begin{itemize}
    \item $l_{\mathsf{SVPos}} = \mathsf{SVAfterCZ}$,
    \item $l_{\mathsf{POVPos}} = \mathsf{POVAfterCZ}$,
    \item $l_{\mathsf{SVPos}} = \mathsf{SVBeforeCZ} \land l_{\mathsf{SVSpeed}} = \mathsf{SVStopped}$,
    \item $l_{\mathsf{SVSpeed}} = \mathsf{SVStopped} \land l_{\mathsf{POVSpeed}} = \mathsf{POVStopped}$.
  \end{itemize}
  In all these cases there is no possibility of further collision, or the
  situation will no longer evolve.

\clearpage
\section{Hybrid Automata differences}

An Hybrid Automata $M(S,\epsilon,E,X,Init,Inv,Flow,Jump)$ is given by :
\begin{itemize}
	\item $S$ a finite set of modes
	\item $\Sigma$ a finite set of event/transition names/labels
	\item $E \subseteq S \times \Sigma \times S$ its finite set of events/edges/control switches
	\item $X \subseteq \bbR$ a finite set of real valued variables. If $x$ is a variable then $\dot{x}$ is its 1st derivative
	\item Init is a function that indicates possible valuation of X variables when the system starts in specific modes
	\item Inv is a function that associate a predicate to each mode restraining the valuation of variables in this mode
	\item Flow is a function that (associate a dynamic to each mode) dictates the possible continuous evolutions of variables in each mode
	\item Jump is a function that, for each edge constrains the possibility of taking that edge (guard) and indicates possible updates of the variables when taking that edge
\end{itemize}

Hybrid Automaton can be subject to defined operators such as product or union/intersection.
\\
\\
\\
There are few differences for our application with Hybrid state systems defined above which are :
\begin{enumerate}
	\item There is no notion of Terminal modes or Safe modes. In practice termination is based on a specified property on one/multiple variables (e.g exit when $x > 100$)
	\item There is no Condition nor determinism in Hybrid automaton. In Hybrid state machine, the condition is used to force the transition to another mode. In Hybrid automata, you can stay in one mode as long as you satisfy its invariant. The conditions can also be used to make sure the hybrid state machine is deterministic.
	\item Since there is no Terminal modes, the complete traces don't exist either in Hybrid automaton. Transitions also differ in HA as there are 2 types of transitions: discrete ones (changing mode by taking an edge) and continuous ones (noticeable change in the variables of X)
	\item In Hybrid Automaton, taking an edge is an instantanous action, for each edge the invariant of it's origin mode\\
\end{enumerate}
The 1st and 3rd point can be adjusted by adding subsets for Terminal and safe nodes and adding a new definition of trace for Hybrid automaton.
For the 2nd point it is possible to translate the condition into a guard predicate and adding it to the existing guard. However by doing that, even if we add a function stating that edges have to be taken as sson as the guard is satisfied we would still lose determinism. In the following diagram we are looking at a system which at some point transition to a B mode without knowing much of its variable when it does so. It can then go to two different outcomes :  the C and D modes depending on the ever incrementing time value.
\\
\begin{example}[determinism]
\begin{tikzpicture}[FAstyle]
	    \node[state, text width=1.2cm, align=center] (A)
	      {A\\ $ \dot{x} = 1$\\ $\dot{t} = 1$};
	    \node[state, right=of A, text width=1.2cm, align=center] (B)
	      {B\\ $ \dot{x}=sin(t)$\\ $\dot{t} = 1$}; 
	    \node[state, below right=of B, text width=1.2cm, align=center] (C)
	      {C\\ };
	     \node[state,right=of B, text width=1.2cm, align=center] (D)
	      {D\\};
	    \node[state, below=of A, text width=1.2cm, align=center] (Ah)
	      {Ah\\ $ \dot{x} = 1$\\ $\dot{t} = 1$};
	    \node[state, right=of Ah, text width=1.2cm, align=center] (Bh)
	      {Bh\\ $ \dot{x}=sin(t)$\\ $\dot{t} = 1$}; 
	    \node[state, below right=of Bh, text width=1.2cm, align=center] (Ch)
	      {Ch\\};
	   \node[state, right=of Bh, text width=1.2cm, align=center] (Dh)
	      {Dh\\ };
	      \path[->]
	        (A)
	          edge node[text width=2cm, align=right]{toB:\\$t < cons$} (B)
	        (B)
	          edge node[text width=2cm]{toC:\\$x \neq -0.5 \mid ...$} (C)
		(B)
	          edge node[text width=2cm]{toD:\\$x \neq 0.5 \mid ...$} (D)
	        (Ah)
	          edge node[text width=2cm, align=right]{toBh:\\$x \geq cons $} (Bh)
	        (Bh)
	          edge node[text width=2cm]{toCh:\\$x = -0.5 \land ...$} (Ch)
		(Bh)
	          edge node[text width=2cm]{toDh:\\$x = 0.5 \land ...$} (Dh);
\end{tikzpicture}
In this example, we compare the B mode of an hybrid state machine with its equivalent Bh mode of an hybrid automata.
In that mode dynamics  $x \in [-1,1]$ since it is the $cos(t)$. We don't know the value of $t$ when entering the B mode.
Ultimately one of the two edges going out of B will be taken since one of the condition will be satisfied before the other.
However on the hybrid automata, we can't gaurantee that the first edge for which the guard predicate is satisfied will taken right away and therefore leading to non determinism in that example.\\
A work around would be to add the edge of the condition as a mode invariant to make sure we don't stay on that mode when an edge becomes available to take.
In the above example it would mean adding $x\neq0.5 \land x\neq-0.5$ as a Bh invariant which doesn't work. This, moreover, might be more restricting that the initial intended mode invariant and in some cases creates non determinism where they were none when the condition depends on 2 variables or more:
\begin{tikzpicture}[FAstyle]
	    \node[state, text width=1.2cm, align=center] (A)
	      {A\\ $ \dot{x} = 1$\\ $\dot{t} = 1$};
	    \node[state, right=of A, text width=1.2cm, align=center] (B)
	      {B\\ $ \dot{x}=sin(t)$\\ $ \dot{z}=sin(t^{3})$\\}; 
	    \node[state, below right=of B, text width=1.2cm, align=center] (C)
	      {C\\ };
	     \node[state, above right=of B, text width=1.2cm, align=center] (D)
	      {D\\};
	   \node[state, right=of B, text width=1.2cm, align=center] (E)
	      {E\\ };
	      \path[->]
	        (A)
	          edge node[text width=2cm, align=right]{toB:\\$x < cons$} (B)
	        (B)
	          edge node[text width=2cm]{toC:\\$x \neq -0.5 \land y < -0.2$} (C)
		(B)
	          edge node[text width=2cm]{toD:\\$x \neq 0.5 \land y > 0.2$} (D)
	        (B)
	          edge node[text width=2cm, align=right]{toE:\\$x \neq 0 \land y \neq 1 $} (E);
	       
\end{tikzpicture}
\end{example}
\clearpage

\section{Experiments}

We ran the workflow on 1 scenarios:
\begin{itemize}
  \item Intersection
\end{itemize}

\subsection{Research Questions}

We have two research questions:
\begin{itemize}
  \item[(RQ1)] Is the precondition calculated by the workflow safe? That is,
        when it is true, the safety condition is never violated.
  \item[(RQ2)] Is the precondition calculated by the workflow too restrictive?
        That is, is it false in instances that seem perfectly safe?
  \item[(RQ3)] Modelling effort: how much work is it to specify a scenario?
\end{itemize}

To assess these we measured the following:
\begin{itemize}
  \item For instances where the precondition was true, but the safety condition
  was violated:
  \begin{itemize}
    \item (bonus) We used robust semantics of the safety condition to
     measure if the violation was serious.
     \item (don't need) We measured the minimum distance to another instance where the
          precondition was true but the safety condition was not violated, to
          measure how sensitive the simulation was to starting conditions.
    \item (super bonus) We measured the effect of increased simulation accuracy on these
          instances, noting that the distance became smaller.
  \end{itemize}
  \item For instances where the precondition was false, but the safety condition
        was not violated:
        \begin{itemize}
    \item Look at the robustness of the precondition.
    \item (accessory to last point) We used robust semantics of the safety condition to measure if the
    safety condition was close to being violated.
    \item (bonus) We measured the minimum distance to another instance \emph{with the same behaviour of POV} where the safety condition was violated, to check sensitivity to starting conditions.
    \item (bonus) Effect of increased simulation accuracy?
          \item (most important) We check that a different behaviour of the POV
                (with the same starting conditions) does indeed lead to a
                simulation where the safety condition is violated.

                Metric: how many instances we get where the RSS precondition is violated, and all of POVs behaviours do not lead to a collision.

  \end{itemize}
\end{itemize}

RQ3: compare to a hybrid program.

\subsection{Delay-Stop machine}

For the Delay-Stop machine, we ran the following experiment:
\begin{itemize}
  \item Car travelling initially at 16m/s.
  \item The position of the wall was variable (from 45 to 55 metres).
  \item The system was simulated at a frequency of 1000 steps per second.
\end{itemize}
The precondition was evaluated at the start. It was accurate at predicting the
safety of the resulting trace up to 0.008m (8mm). Increasing the simulation
precision to 10,000 resulted in an error of 2mm, thus the error is likely just
due to the inaccuracy of the simulation.

\todoil{Rerun this with all the analysis explained above.}

\subsection{Intersection machine}

\todoil{Explain how behaviour of POV was simulated}
During simulation, with similar starting parameters, the POV's behaviour can greatly influence the safety of the resulting trace.
By cruising, accelerating or braking, it can change it's collision zone entrance timing and the time spent in it thus affecting the safety.
From a simulation perspective we must consider the worst possible behaviour of the POV which is when it actively tries to be in the CZ at the same time as the SV.
\begin{itemize}
  \item To do that, the POV brakes for the max amount when inside the CZ, making the time it spends in it as long as possible.
  \item To find the worst timing(s) at which it could enter the CZ, we also simulate different constant acceleration values before reaching it.
\end{itemize}

\section{CONCLUSIONS}

Conclusion goes here.

\addtolength{\textheight}{-12cm}   



\section*{APPENDIX}

Appendixes should appear before the acknowledgment.

\section*{ACKNOWLEDGMENT}

Acknowledgment goes here.

\end{auxproof}


\end{document}

